\documentclass[10pt,twocolumn,letterpaper]{article}

\usepackage[pagenumbers]{cvpr} 

\usepackage[accsupp]{axessibility} 
\usepackage[dvipsnames]{xcolor}
\definecolor{aliceblue}{rgb}{0.94, 0.97, 1.0}

\usepackage{circledsteps}

\usepackage{graphicx}
\usepackage{float}
\usepackage{stfloats}

\usepackage{subcaption}

\usepackage{multirow}
\usepackage{booktabs}
\usepackage{amssymb}
\usepackage{bbding}
\usepackage{pifont}
\usepackage{wasysym}
\usepackage{utfsym}
\usepackage{fontawesome}
\usepackage{soul} 
\usepackage{color, xcolor} 
\usepackage{colortbl} 
\usepackage{xcolor}
\usepackage{array}  

\definecolor{red}{RGB}{255, 0, 0}
\definecolor{blue}{RGB}{0, 0, 255} 
\definecolor{SeaGreen}{RGB}{32, 158, 133} 
\definecolor{LightSeaGreen}{RGB}{32,178,170}
\definecolor{apricot}{rgb}{0.98, 0.81, 0.69}
\definecolor{babyblue}{rgb}{0.54, 0.81, 0.94}
\definecolor{celadon}{rgb}{0.67, 0.88, 0.69}
\definecolor{dollarbill}{rgb}{0.52, 0.73, 0.4}
\definecolor{inchworm}{rgb}{0.7, 0.93, 0.36}
\definecolor{lavenderblue}{rgb}{0.8, 0.8, 1.0}
\definecolor{lightblue}{rgb}{0.68, 0.85, 0.9}
\definecolor{lightgray}{rgb}{0.83, 0.83, 0.83}
\definecolor{grannysmithapple}{rgb}{0.66, 0.89, 0.63}

\definecolor{cvprblue}{rgb}{0.21,0.49,0.74}
\usepackage[pagebackref,breaklinks,colorlinks,citecolor=cvprblue]{hyperref}

\def\paperID{} 
\def\confName{CVPR}
\def\confYear{2024}

\title{Delving into the Trajectory Long-tail Distribution for Muti-object Tracking}

\author{Sijia Chen \thanks{Equal contribution}~ \hspace{16pt} En Yu $^{*}$ \hspace{16pt} Jinyang Li \hspace{16pt} Wenbing Tao \thanks{Corresponding author}\\
Huazhong University of Science and Technology, China\\
{\tt\small \{sijiachen, yuen, jinyangli, wenbingtao\}@hust.edu.cn}
}

\begin{document}
\maketitle

\begin{abstract}

Multiple Object Tracking (MOT) is a critical area within computer vision, with a broad spectrum of practical implementations. Current research has primarily focused on the development of tracking algorithms and enhancement of post-processing techniques. Yet, there has been a lack of thorough examination concerning the nature of tracking data it self. In this study, we pioneer an exploration into the distribution patterns of tracking data and identify a pronounced long-tail distribution issue within existing MOT datasets. We note a significant imbalance in the distribution of trajectory lengths across different pedestrians, a phenomenon we refer to as ``pedestrians trajectory long-tail distribution''. Addressing this challenge, we introduce a bespoke strategy designed to mitigate the effects of this skewed distribution. Specifically, we propose two data augmentation strategies, including Stationary Camera View Data Augmentation (SVA) and Dynamic Camera View Data Augmentation (DVA) , designed for viewpoint states and the Group Softmax (GS) module for Re-ID. SVA is to backtrack and predict the pedestrian trajectory of tail classes, and DVA is to use diffusion model to change the background of the scene. GS divides the pedestrians into unrelated groups and performs softmax operation on each group individually. Our proposed strategies can be integrated into numerous existing tracking systems, and extensive experimentation validates the efficacy of our method in reducing the influence of long-tail distribution on multi-object tracking performance. The code is available at \href{https://github.com/chen-si-jia/Trajectory-Long-tail-Distribution-for-MOT}{https://github.com/chen-si-jia/Trajectory-Long-tail-Distribution-for-MOT}.

\end{abstract}

\section{Introduction}
\label{sec:intro}
\quad

\begin{figure}[ht]
\begin{subfigure}{.48\linewidth}
  \centering
  \includegraphics[width=1.0\linewidth]{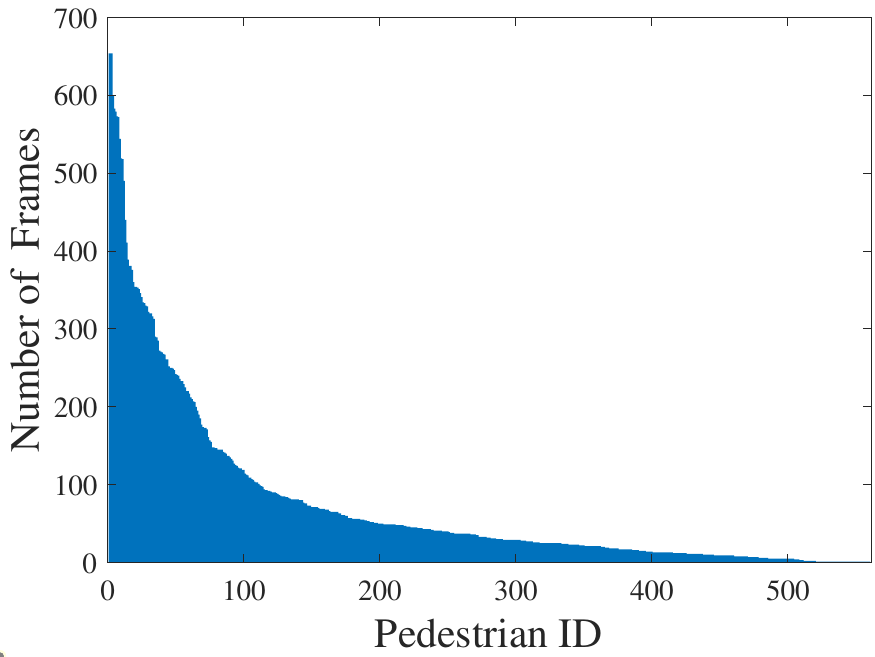}
  \caption{MOT15}
  \label{fig:a}
\end{subfigure}
\hfil
\begin{subfigure}{.48\linewidth}
  \centering
  \includegraphics[width=1.0\linewidth]{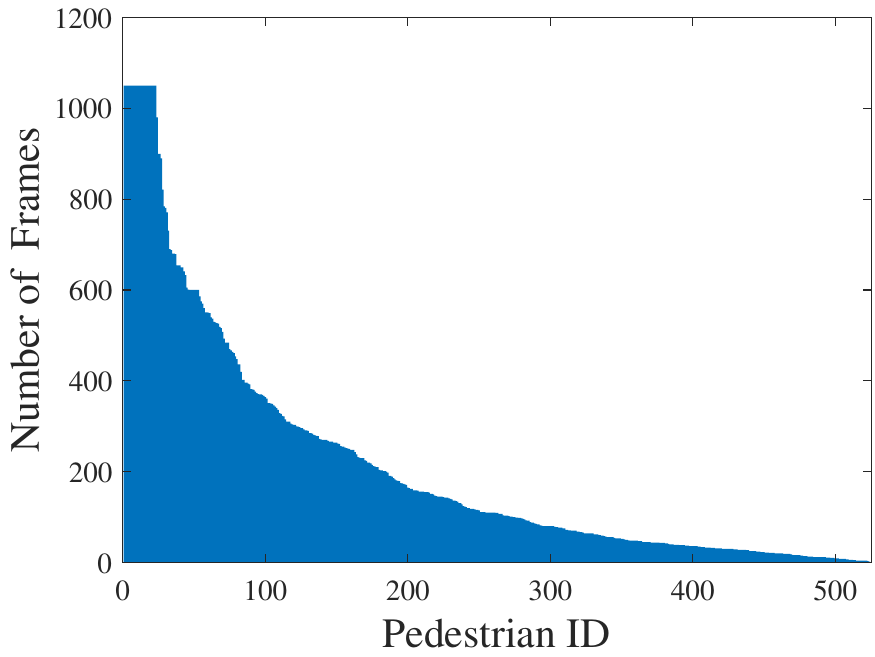}
  \caption{MOT16}
  \label{fig:b}
\end{subfigure}
\newline
\begin{subfigure}{.48\linewidth}
  \centering
  \includegraphics[width=1.0\linewidth]{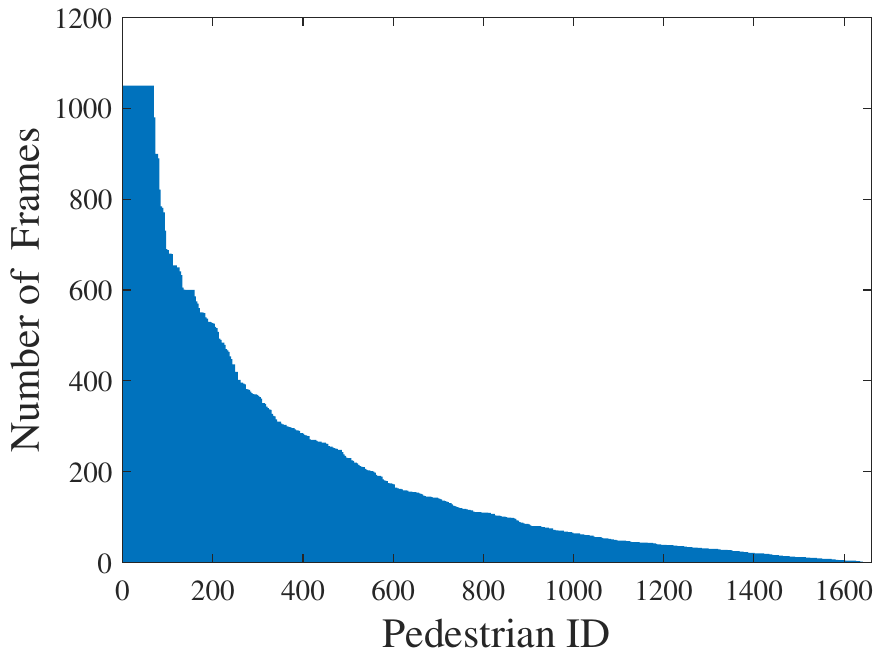}
  \caption{MOT17}
  \label{fig:c}
\end{subfigure}
\hfil
\begin{subfigure}{.48\linewidth}
  \centering
  \includegraphics[width=1.0\linewidth]{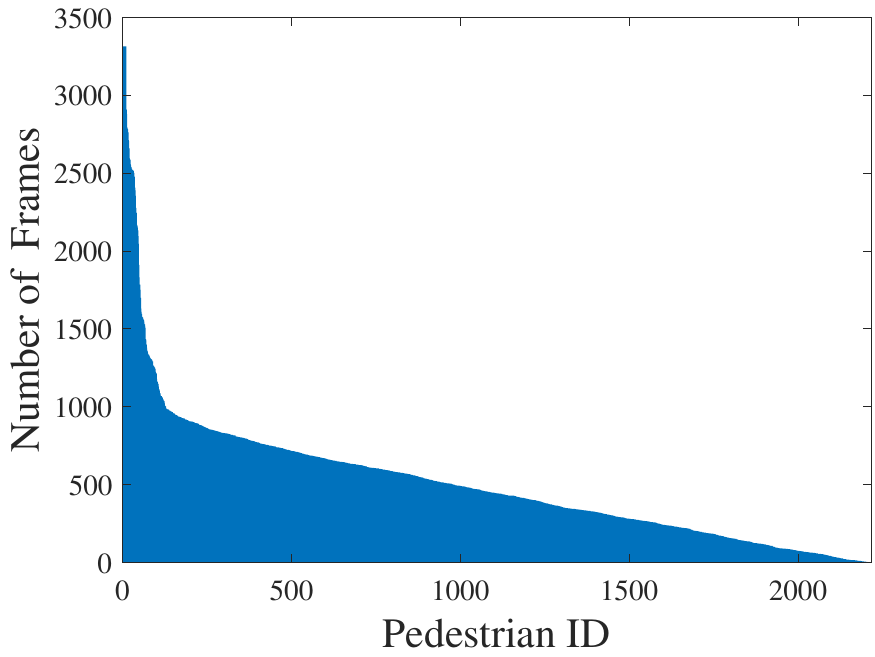}
  \caption{MOT20}
  \label{fig:d}
\end{subfigure}
\caption{The number of frames of pedestrians with different identities in the MOTChallenge datasets. We stipulate that different pedestrian identities are regarded as different pedestrian classes.}
\label{fig:The number of frames of pedestrians with different identities in the MOTChallenge datasets.}
\end{figure}

Multi-object tracking stands as one of the fundamental and most challenging tasks in computer vision. This technology involves the tracking of multiple objects of interest in video sequences, providing essential parameters such as location, trajectory and velocity. Multi-object tracking has found widespread applications in domains like autonomous driving, video analysis and smart transportation~\cite{bashar2022multiple}.

The prior research efforts in the field of multi-object tracking have primarily focused on the designs of tracking networks and post-processing strategies. 
However, the current MOT methods do not pay attention to the pedestrian long-tail characteristics of tracking data. 
Thus, we conducted an analysis experiment to count the number of frames of pedestrians with different identities in the MOTChallenge datasets.
As shown in \cref{fig:The number of frames of pedestrians with different identities in the MOTChallenge datasets.}, we observe a large difference in the number of frames for different pedestrian identities.
Because the characteristic of the long-tail distribution is that the head classes possess a substantial number of instances and the tail classes has only a few instances, we conclude that the number of pedestrian identities obeys the long-tail distribution.

There is a common problem with long-tail distribution datasets: The network is trained on long-tail distribution data often results in a bias towards learning features associated with the prevalent head classes, while neglecting those in the less represented tail classes. At present, the problem of improving long-tail distribution can be divided into three aspects: class re-balancing, information augmentation and module improvement. In the image data collected by the camera, some people stay in the image for a long time, and some people move across the image in a hurry. Due to the reasons of the data itself, the network will learn less features of people who have hurriedly passed by. For the current Re-ID branch of multi-object trackers, most of them regard Re-ID as a classification problem and use the softmax module to calculate the classification probability. However, the softmax module has a huge flaw: the weights of classes with large weights become larger, and the weights of classes with small weights become smaller, which will intensify the long-tail distribution effect on the long-tail distribution data.  
Hence, to improve the issue, we propose our solution from two key perspectives: information augmentation and module improvement. 

In the perspective of information augmentation, we classify camera data into two categories: stationary camera view data and dynamic camera view data, based on the motion status of the camera.
For stationary camera view data, we propose the Stationary Camera View Data Augmentation (SVA) strategy that encompasses two techniques: backtracking continuation and prediction continuation. The backtracking continuation is applied to the pedestrians of tail classes in the middle frame of the training sequence data, while the prediction continuation is employed for the pedestrians of tail classes in the final frame of the training sequence data. This strategy can promote the network's learning of pedestrian trajectories in the tail classes. For dynamic camera view data, the Dynamic Camera View Data Augmentation (DVA) strategy is proposed. This strategy uses the diffusion model to perform style transformation on the scene background, improving the network's attention to the features of pedestrians areas.

In terms of module enhancement, we devise the Group Softmax (GS) module. The GS groups pedestrians with a similar number of training samples together, and then computes the softmax and cross-entropy loss for each group individually, preventing a significant suppression of tail classes by the weights of head classes, improving the network’s ability to extract appearance features of tail classes.

We apply our tailored solution to the SOTA FairMOT~\cite{zhang2021fairmot} and CSTrack~\cite{liang2022rethinking} of multi-object tracking networks and evaluate them on four public MOT benchmarks, ie., MOT15~\cite{leal2015motchallenge}, MOT16~\cite{milan2016mot16}, MOT17~\cite{milan2016mot16} and MOT20~\cite{dendorfer2020mot20}. The experimental results clearly demonstrate that our approach appear to significant improvements. The performance of our network trained using only MOT20 data far exceeds the performance of the baseline trained using mixed data on the MOT20 test set. For example, FairMOT using our strategy is only trained on the MOT20 data, which is better than FairMOT trained using the mixed data. On the MOT20 test set, it increased 4.1\% MOTA and 3.0\% IDF1.

The main contributions of this work are as follows:

\begin{itemize}  [leftmargin=8mm]
  \item [1.]
  We serve as  the first to discover the long-tail distribution problem in multi-object tracking and point out that this problem is caused by the imbalance of the number of frames for different pedestrians.
  \item [2.]
  We propose the tailored data augmentation strategies, including SVA and DVA, from information augmentation perspective. SVA is used to backtrack and predict the pedestrians trajectory of tail classes, and DVA is used to change the background of the scene. Additionally, we design the GS module from module improvement perspective. The GS divides pedestrians with different identities into unrelated groups and performs a separate softmax operation on each group.
  \item [3.]
  We apply our method to two SOTA Joint Detection and Tracking algorithms and evaluate on MOTChallenge datasets. The evaluation results showcase robust performance  improvements, serving as a compelling validation of the efficacy of our method.
\end{itemize}

\section{Related Work}
\label{sec:Related_Work}

\begin{figure*}
 \setlength{\belowcaptionskip}{-0.2cm} 
  \centering
    \includegraphics[width=1.0\linewidth]
    {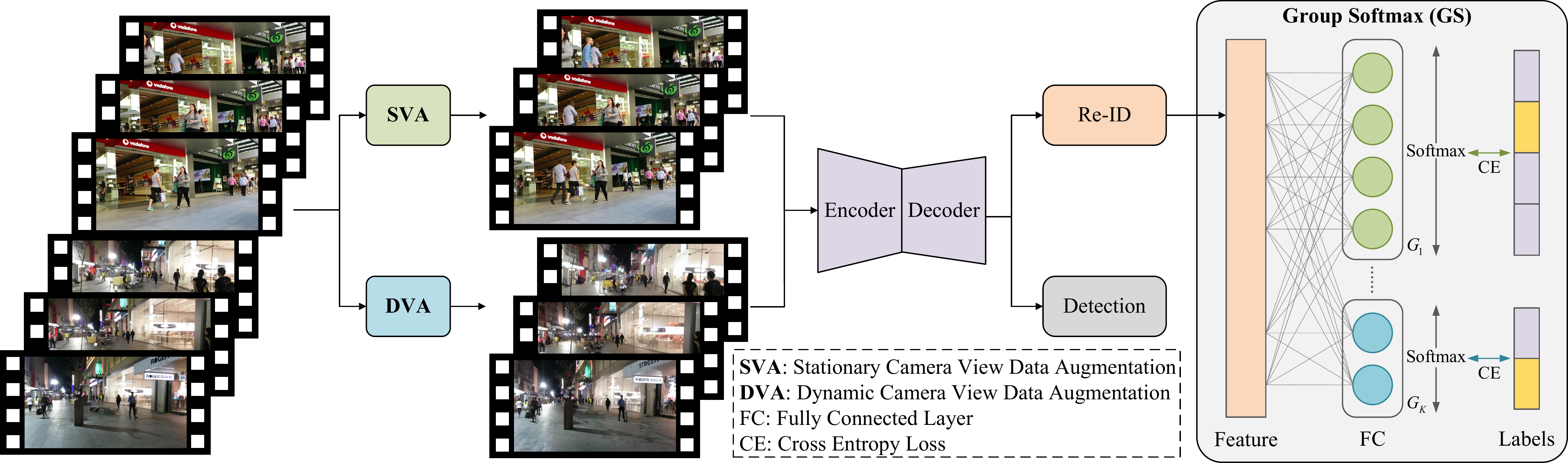}
  \caption{Overall pipeline of our strategies. Our strategies comprise 3 modules: (1) SVA: To backtrack and predict the pedestrians trajectory of tail classes. (2) DVA: To use diffusion model to change the background of the scene. (3) GS: To divide the pedestrians with different identities into unrelated groups and perform softmax operation on each group individually.}
  \label{fig:Overall pipeline of our strategies}
\end{figure*}

\textbf{Multiple Object Tracking.} We review three main multiple object tracking frameworks: Tracking-by-Detection, Joint Detection and Tracking, and Transformer-based Tracking. Tracking-by-Detection \cite{bewley2016simple, wojke2017simple,han2022mat, zhang2022bytetrack, du2023strongsort, fischer2023qdtrack} primarily consists of two main components. In the detection phase, a detector is established to locate objects of interest. In the association phase, early methods used motion predictors to forecast the positions of objects in the next frame and relied on positional information to associate objects across consecutive frames. 
Joint Detection and Tracking\cite{wang2019towards, bergmann2019tracking, zhou2020tracking, tokmakov2021learning, zhang2021fairmot, liang2022rethinking, yu2022relationtrack}, a unified network simultaneously produces detection results and the corresponding appearance features of the detected objects. Subsequently, association methods are employed to link objects between consecutive frames. 
Transformer-based Tracking \cite{sun2020transtrack, meinhardt2022trackformer, zhou2022global, zeng2022motr, zhang2023motrv2, yu2023motrv3, yu2023generalizing} uses a transformer to combine detection queries with queries derived from the previous frame predictions for detecting and tracking objects in the current frame. This approach eliminates the need for post-processing steps, enabling end-to-end multi-object tracking.

\noindent \textbf{Long-tail Distribution.}
Improving long-tail distribution issues can be categorized into three aspects: class re-balancing, information augmentation, and module improvement. 
Class re-balancing including resampling, cost-sensitive learning, and Logit adjustment. Resampling \cite{kang2019decoupling, ren2020balanced, zhang2021videolt, feng2021exploring}
is one of the most widely used methods for addressing class imbalance. 
Cost-sensitive learning \cite{cao2020domain, cao2019learning, ye2020identifying, cui2019class} involves adjusting the loss weights of different classes. 
Logit adjustment \cite{zhao2022adaptive, tao2023local, wang2023balancing} is a post-hoc technique that shifts the model's logits based on label frequencies.
Information augmentation including transfer learning and data augmentation. transfer learning aims to transfer knowledge from a source domain to enhance model training in a target domain. Transfer learning \cite{yin2019feature, liu2020deep, he2019rethinking, zoph2020rethinking, xiang2020learning} primarily includes four approaches: head-tail knowledge transfer, model pre-training, knowledge distillation, and self-training. Data augmentation \cite{chou2020remix, zang2021fasa, li2021metasaug} utilizes an enhancement technique to increase the size and quality of the model training dataset, playing a significant role in optimizing especially limited training sets.
Module improvement including representation learning, classifier design, decoupled training, and ensemble learning. 
Representation learning \cite{cui2021parametric, samuel2021distributional, zhong2019unequal, zhu2020inflated, zhong2021improving} entails refining network structures to facilitate a more effective acquisition of informative representations. Classifier design \cite{he2020momentum, liu2021gistnet} is designed by setting up the appropriate classifier so that it focuses more on the tail class. Decoupled training \cite{kang2020exploring, zhang2021distribution, desai2021learning} untangles the learning process by segregating it into representation learning and classifier training, ensuring that they do not affect each other. Ensemble learning \cite{zhou2020bbn, guo2021long, li2020overcoming} to solve the issues of long-tail learning by deliberately generate and amalgamate multiple network modules.

\section{Methodology}
\label{sec:Methodology}

\subsection{Overview}
\quad

In this work, we choose the multi-object tracking algorithm of the Joint Detection and Tracking framework~\cite{wang2019towards} to carry out our strategies, which is illustrated in \cref{fig:Overall pipeline of our strategies}. From the data perspective, we propose two bespoke data augmentation methods, including Stationary Camera View Data Augmentation (SVA) and Dynamic Camera View Data Augmentation (DVA), designed for viewpoint states, to simulate the pedestrians of tail classes motion trajectory and change the background style. In addition, we commence by addressing the similarity metric Re-ID used in the association and propose the Group Softmax (GS) module to improve the appearance recognition performance for the pedestrians of tail classes.

\subsection{Camera View Data Augmentation}
\quad

We observed the distinct characteristic in the data from the multi-object tracking, which the data was collected under varying camera motion conditions. Depending on the camera motion conditions during data collection, it can be categorized into data collected from static cameras and data collected from moving cameras. Consequently, we developed the customized data augmentation methods for data collected from stationary cameras and dynamic cameras. 

In particular, we define the calculation formula for category division as shown in \cref{eq:classes}. Then, we use \cref{eq:classes} to divide the pedestrian categories in each sequence in the dataset into head classes and tail classes.
\begin{equation}
C_{i} = \begin{cases}C_{i}^{tail} & \frac{1}{R_i} \geq T_j \\ C_{i}^{head} & \frac{1}{R_i}<T_j\end{cases}
\label{eq:classes}
\end{equation}
where ${C}_i$ represents the classes to which category $i$ belongs, ${R}_i$ represents the ratio of the number of category $i$ to the number of all categories in $j$ sequence, $T_j$ represents the class threshold in $j$ sequence for determining whether the classes of is tail or not.

\subsubsection{Stationary Camera View Data Augmentation}
\label{sec:SVA}
\quad

\begin{figure}
 \setlength{\belowcaptionskip}{-0.2cm} 
  \centering
   \includegraphics[width=1.0\linewidth]{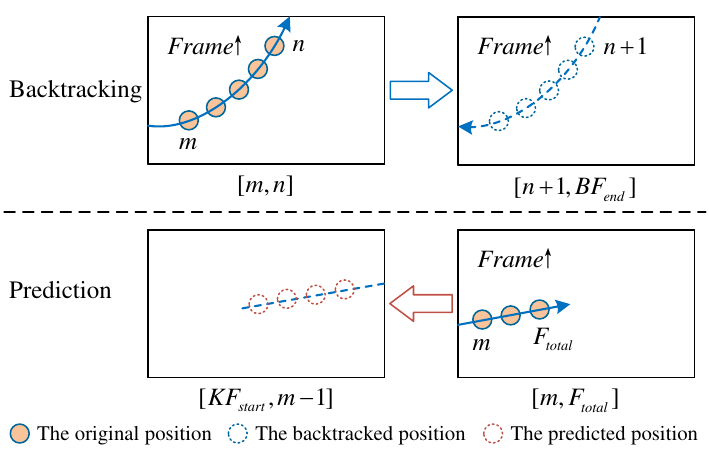}
   \caption{Illustration of the Stationary Camera View Data Augmentation (SVA).}
   \label{fig:SVA}
\end{figure}

For multi-object tracking data captured by stationary cameras, common data augmentation methods, such as image color transformation, image blending and image cropping, are currently available. Although these methods can be applied, they are not specifically designed for multi-object tracking tasks and lack customized designs for tracking targets. Therefore, we propose the Stationary Camera View Data Augmentation (SVA) strategy tailored for the multi-object tracking of data captured by stationary cameras, focusing on quantity augmentation for the pedestrians of tail classes. The SVA strategy includes backtracking continuation and prediction continuation, which is shown in \cref{fig:SVA}. 
The backtracking continuation is to add the reversed original track in the subsequent frames after the end of the original track, applied to the pedestrians of tail classes in the middle frame of the training sequence data. The prediction continuation is to add the future trajectory predicted using the position information of the original trajectory to the previous frames at the beginning of the original trajectory, employed for the pedestrians of tail classes in the final frame of the training sequence data.

\noindent \textbf{Backtracking continuation.}
For a training video with a total of $ F_{\emph{total}}$ frames, if the pedestrian trajectory of tail classes appears in the $ m$-th frame and disappears in the $ n$-th frame, satisfying the condition $ n<F_{total}$, we employ the Segment Anything Model (SAM)~\cite{kirillov2023segment} algorithm to segment the image area of the pedestrian that appear in frames from the $ m$-th to the $ n$-th and then overlay these image areas in reverse order onto frames following the $ n$-th frame. The backtracking continuation can be formulated as:
\begin{equation}
\begin{gathered} 
{BP}_j^k=P_i^k (m \leq i \leq n, n+1 \leq j \leq {BF}_{\emph{end}})
\end{gathered}
\end{equation}
where ${BP}_j^k$ represents the backtracked image position of the $k$-th pedestrian in the training data at the $j$-th frame, $P_i^k$ represents the image position of the $k$-th pedestrian in the training data at the $i$-th frame, and ${BF}_{\emph{end}}$ is the backtracking continuation cutoff frame in the training dataset, and the value is the minimum value of $F_{\emph{total}}$ and $(2n-m)$.

\noindent \textbf{Prediction continuation.}
For a training video with a total of $ F_{\emph{total}}$ frames, if the pedestrian trajectory of tail classes appears in the final frame, we input the $x$ and $y$ image coordinates of the pedestrian appearing from the $m$-th frame to the $ F_{\emph{total}}$-th frame into a Kalman filter to predict the subsequent $x$ and $y$ image coordinates of the pedestrian, while ensuring that the predicted image coordinates fall in the image size range. In this pedestrian trajectory, we randomly select the pedestrian with visibility no less than the visibility threshold $T_v$, where $0 \leq T_v \leq 1$, using the SAM~\cite{kirillov2023segment} algorithm to segment the pedestrian. 
The segmented image area is superimposed on the frame before the pedestrian trajectory appears based on the predicted $x$ and $y$ image coordinate randomly selected from the predicted image coordinates. The prediction continuation can be formulated as:
\begin{equation}
\begin{gathered}
{KP}_j^k=R(KF\left(P_i^k\right)) (m \leq i \leq F_{\emph{total }}, {KF}_{\emph{start }} \leq j < m)
\end{gathered}
\end{equation}
where ${KP}_j^k$ represents the Kalman filter predicted image position of the $k$-th pedestrian in the training data at the $j$-th frame, 
$R()$ represents the function that randomly selects an image position, 
$KF\left(P_i^k\right)$ represents the image positions predicted by the Kalman filter using the images coordinates of the $m$-th frame to $F_{\emph{total }}$ for the $i$-th pedestrian in the training data, 
and $P_i^k$ represents the image position of the $k$-th pedestrian in the training data at the $i$-th frame. ${KF}_{start}$ is the starting frame for applying the prediction continuation in the training dataset, and the value is the maximum value of 1 and $(2m-F_{\emph{total}})$.

\subsubsection{Dynamic Camera View Data Augmentation}
\label{sec:DVA}
\quad

\begin{figure}
 \setlength{\belowcaptionskip}{-0.25cm}
  \centering
    \includegraphics[width=1.0\linewidth]{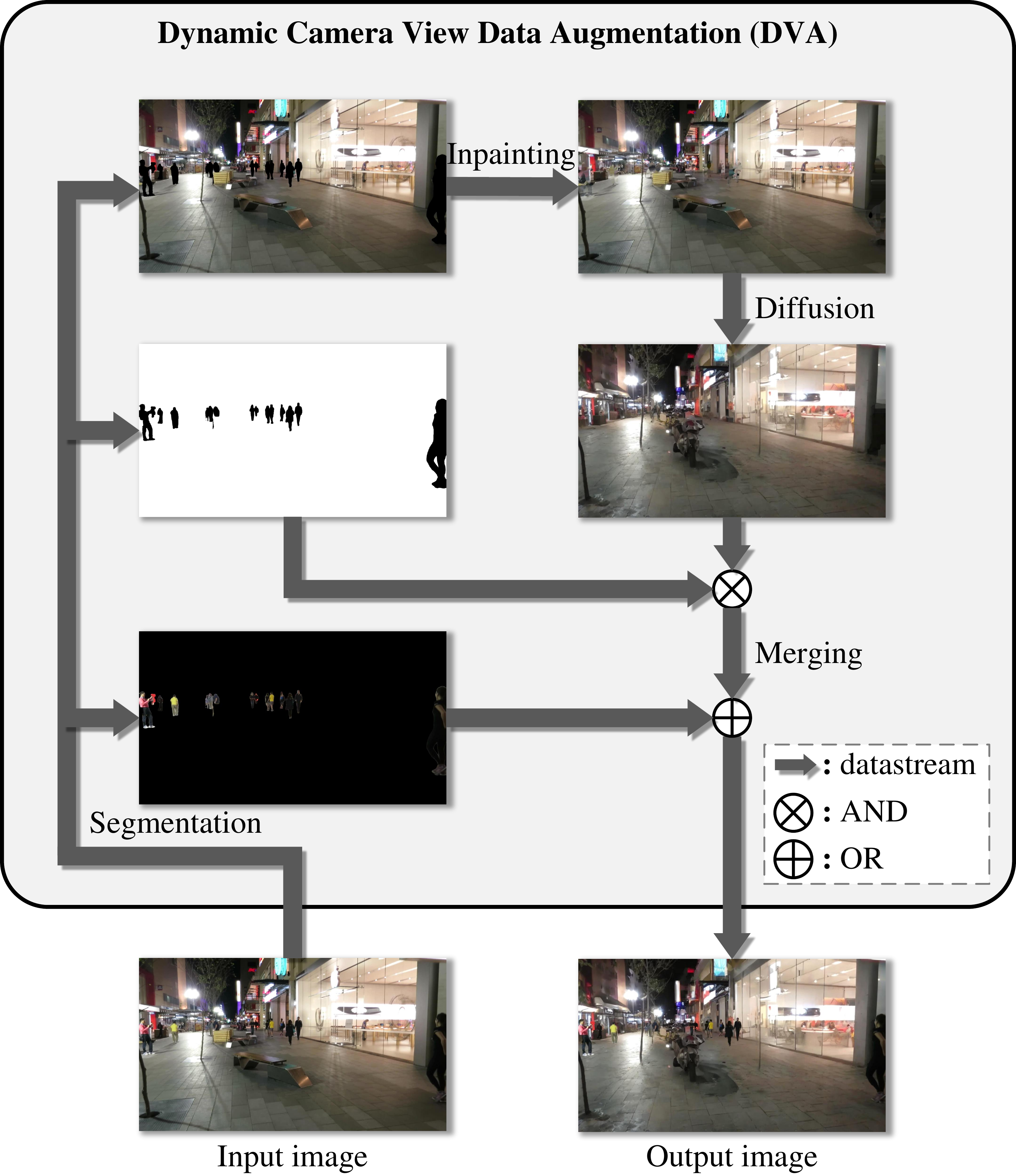}
  \caption{Illustration of the Dynamic Camera View Data Augmentation (DVA).}
  \label{fig:DVA}
\end{figure}

Due to the characteristics of significant scene and subject size variations in data captured by dynamic cameras, the traditional data augmentation methods struggle to adapt to these changes. To address this issue, we propose the Dynamic Camera View Data Augmentation (DVA) strategy, as depicted in \cref{fig:DVA}. The strategy comprises four main steps: image segmentation, image inpainting, image diffusion and image merging. This strategy, designed for input from dynamic camera perspectives, begins by using the image segmentation algorithm SAM~\cite{kirillov2023segment} to separate pedestrians in the input image derived from the sequence, resulting in image with pedestrians removed, image with pedestrian mask, and image containing only the pedestrians area. Next, the image inpainting algorithm Navier-Stokes~\cite{bertalmio2001navier} is applied to repair the image with pedestrians removed, producing the repaired image. Following this, the Stable Diffusion~\cite{rombach2022high} is used to process the repaired image, resulting in the diffused image. Finally, the image  with pedestrians mask and the image containing only the pedestrians area obtained from the earlier segmentation step are merged with the diffused image to generate the output image.

\noindent \textbf{Image Segmentation.} SAM, which stands for Segment Anything Model, is the largest segmentation model ever released by Meta. This model segments objects by taking both a prompt and an image as inputs. Given that the multi-object tracking dataset provides bounding box annotations but lacks mask labels for pedestrians, we utilize the image and its corresponding pedestrians bounding box labels as input for SAM to  segment the pedestrians in the image. 

\noindent \textbf{Image Inpainting.} In this paper, the algorithm used for image inpainting is based on the Navier-Stokes equation~\cite{bertalmio2001navier}. This algorithm aims to start repairing the image from the edges of the area to be patched, propagating image smoothness along the contour lines, and obtaining the repaired image after all information has been propagated. 

\noindent \textbf{Image Diffusion.} Stable Diffusion is a type of Latent Diffusion Model (LDM)~\cite{rombach2022high}, which is a class of denoising diffusion probability models capable of generating new image. In principle, Stable Diffusion can model conditional distributions. This can be achieved by inputting text, semantic maps, or the other image-to-image transformation task information to control conditional denoising autoencoders. In this paper, we utilize input image to generate new image by adjusting prompt and enhancement coefficient.

\noindent \textbf{Image Merging.} We perform a bitwise AND operation between the image with pedestrian mask and the diffused image, effectively setting the pixel values in the diffused image corresponding to the original pedestrians area to 0, while leaving the pixel values in area outside the pedestrians area are unchanged. This results in the post-processed diffused image. Then, we perform a bitwise OR operation between the post-processed diffused image and the image containing only the pedestrians. Indeed, this operation involves setting the pixel values in the diffused image that correspond to the original pedestrians area to the corresponding pixel values from the image containing only pedestrians, resulting in the output image.

Training models on the augmented data from diffusion model often risks over-emphasizing spurious qualities~\cite{antoniou2017data}. The common solution assigns different sampling probabilities to original and augmented data to manage imbalance~\cite{he2022synthetic}. We apply a similar method to balance original images and the augmented images from DVA. Mathematically, the method can be formulated as:
\begin{equation}
I_i^n= \begin{cases}I_i & , P_i^n \leq T_s \\ \tilde{I}_i & , P_i^n>T_s\end{cases}
\label{eq:T_p}
\end{equation}
where $I_i^n$ represents the image of index $i$ in the $n$-th epoch, $I_i$ represents the original image of index $i$, $\tilde{I}_i$ represents the augmented image of index $i$, $P_i^n$ represents the probability of index $i$ calling the original image in the $n$-th epoch, $T_s$ represents the image selection threshold of calling the original image for each image selection. Given index $i$, with probability $T_s$ the original image $I_i$ is added to the epoch $n$, otherwise its augmented image $\tilde{I}_i$ is added.

\subsection{Group Softmax Module}
\label{sec:GS}
\quad

We observe the issue where the Re-ID have different degrees of feature learning for pedestrian categories with different quantities. It tends to perform better for categories with higher quantities (head classes) and less effectively for categories with fewer quantities (tail classes), which can negatively impact the performance of the Re-ID. To tackle this problem, we propose the Group Softmax (GS) module, as depicted in \cref{fig:Overall pipeline of our strategies}. The GS divides the pedestrian categories into several disjoint groups and performes softmax operation separately for each group. In this way, the pedestrian categories with similar quantities can compete in the same group. Thus, the GS can isolate categories with significant quantity differences, preventing the weights of tail classes from being heavily suppressed by the head classes.

Specifically, we divides the total of $M$ pedestrian classes in the training dataset into $K$ distinct groups according to their number in the training dataset, and the rule of partitioning group formula is: $T_j^l \leq N(i) \leq T_j^h$, where the value of $i$ is from 1 to $M$, the value of $j$ is from 1 to $K$, $N(i)$ is the quantity of the $i$-th pedestrian categories in the training dataset, $T_j^l$ is the lowest quantity thresholds for the $j$-th group, $T_j^h$ is the highest quantity thresholds for the $j$-th group, $M$ represents the number of pedestrian categories, $K$ represents the number of groups. 

To ensure each pedestrian category is only assigned to one group and maintain ordered groups, we specify that the lowest quantity threshold for the $j+1$-th group is equal to the highest quantity threshold for the $j$-th group, i.e., $T_{j+1}^l$ = $T_j^h+1$. To facilitate grouping, we propose the set rule of $T_j^h$ formula is as follows:

\begin{equation}
T_j^h = \frac{j}{K}max(N(i)) (1 \leq i \leq M, 1 \leq j \leq K)
\end{equation}
where $T_j^h$ is the highest quantity thresholds for the $j$-th group, $j$ represents the index of group.

Furthermore, we individually apply softmax processing to each group and utilize the Cross-Entropy Loss to compute the group loss. Then, we calculate the group loss mean as the Re-ID loss, the formula is as follows:
\begin{equation}
L o s s_{\emph{Re-ID}}= - \frac{1}{K} \sum_{j=1}^K \sum_{i \in {G}_j} y_i \log \left(p_i \right)
\end{equation}
where $L o s s_{\emph{Re-ID}}$ represents the Re-ID loss, $K$ represents the number of groups, $j$ represents the index of group, ${G}_j$ represents the $j$-th group, $y_i$ represents the label in $G_{j}$, $p_i$ represents the probability in $G_{j}$. 

\section{Experiments}
\label{sec:Experiments}

\subsection{Datasets and Evaluation Metrics}
\textbf{Datasets:}We conduct extensive experiments on four public MOT benchmarks, i.e., MOT15~\cite{leal2015motchallenge}, MOT16~\cite{milan2016mot16}, MOT17~\cite{milan2016mot16} and MOT20~\cite{dendorfer2020mot20}. Specifically, MOT15 contains 22 sequences, 11 for training and the other 11 for testing, which includes 11286 frames. MOT16 contains 14 sequences, 7 for training and the other 7 for testing, which includes 11235 frames. Compared with MOT16, MOT17 adds the detection bounding boxes of three detectors that DPM, SDP, Faster-RCNN. MOT20 contains 8 sequences captured in the crowded scenes, 4 for training and the other 4 for testing, which includes 13410 frames. In some frames, more than 200 pedestrians are included simultaneously.

\noindent \textbf{Evaluation metrics:}To evaluate, we use the CLEAR metrics~\cite{bernardin2008evaluating}, including multiple object tracking accuracy (MOTA), ID F1 score (IDF1), Higher Order Tracking Accuracy (HOTA), mostly tracker rate (MT), mostly lost rate (ML) and identity switches (IDS). MOTA, IDF1 and HOTA are three important comprehensive metrics. MOTA focuses on detection performance, and IDF1 focuses on association performance. Compared with them, HOTA balances detection performance and association performance.

\subsection{Implementation Details}
\quad
All experiments are trained using an NVIDIA GeForce RTX 3090 GPU.
All models are trained for 30 epochs.
For MOT15, we set the class threshold $T_j$ to 15 for all stationary camera view sequences, the visibility threshold $T_v$ to 1.0, the image selection threshold $T_p$ to 0.8, the prompt of diffusion to ``A street'' for all dynamic camera view sequences, the enhancement coefficient of diffusion to 0.4, and the group number $K$ to 3 for FairMOT and 4 for CSTrack. 
For MOT16 and MOT17, we set $T_j$ to 120 for 02, 04 and 09 sequences, $T_v$ to 1.0, $T_p$ to 0.9, except $T_p$ is 1.0 for CSTrack on MOT17, prompt to ``A street'' for 05, 10 and 13 sequences, ``A mall'' for 11 sequence, enhancement coefficient to 0.4, $K$ to 3 for FairMOT on MOT16, $K$ to 3 for FairMOT on MOT17, 4 for CSTrack on MOT16, and 2 for CSTrack on MOT17. 
For MOT20, since MOT20 is all stationary camera view data, we only need to set the SVA and GS parameters.We set $T_j$ to 1000 for all sequences, $T_v$ to 1.0, and $K$ to 2.
For a fair comparison, we adopt the same training settings as for the baseline to retrain each tracking model with and without our method, where the training datasets include MOT15, MOT16, MOT17 and MOT20.

\subsection{Comparison of long-tail distribution solutions.}
\begin{figure}[t]
 \setlength{\abovecaptionskip}{0.1cm}  
 \setlength{\belowcaptionskip}{-0.15cm} 
  \centering
  \includegraphics[width=1.0\linewidth]{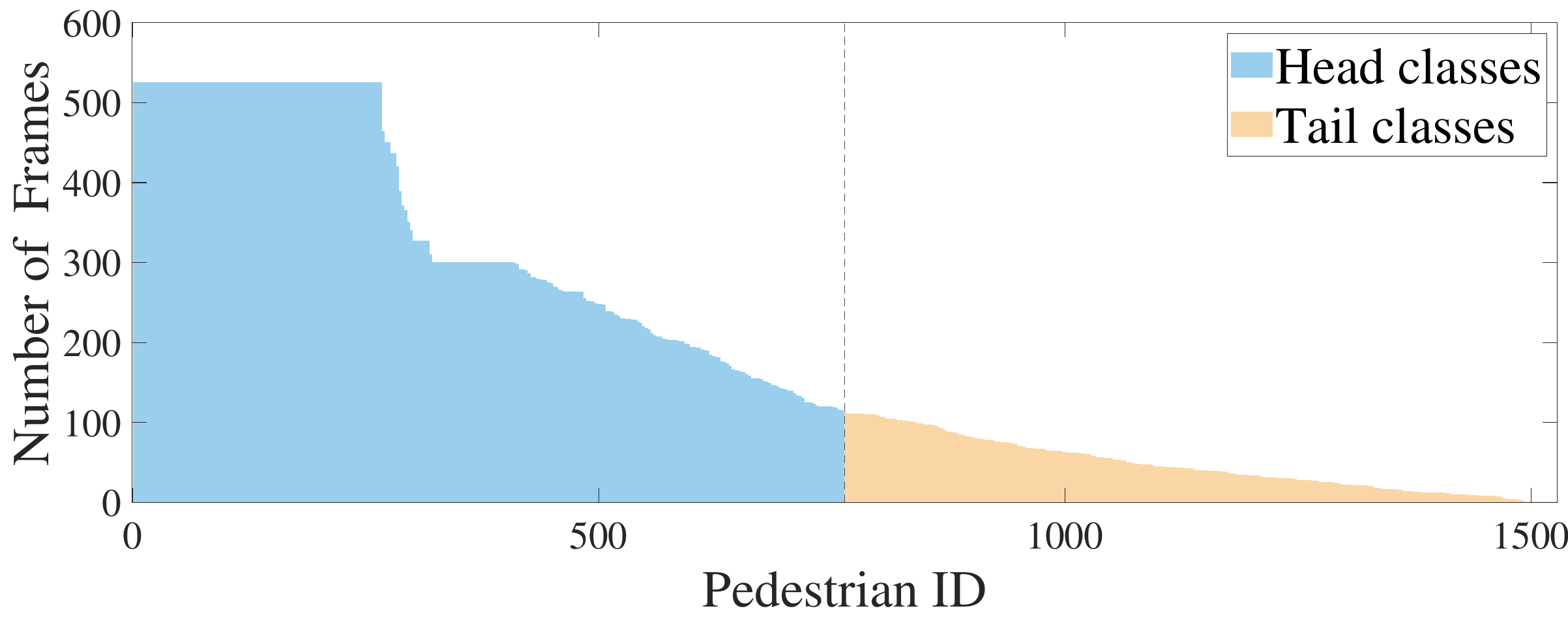} 
  \caption{Division of head classes and tail classes is based on the class average principle on the MOT17 validation set.}
  \label{fig:Division of head classes and tail classes is based on the class average principle on the MOT17 validation set.}
\end{figure}

\quad
We follow the settings of the ablation experiment, count the number of frames of pedestrians with different identities on the MOT17 validation set, and divide all classes into head classes and tail classes according to the class average principle, as shown in \cref{fig:Division of head classes and tail classes is based on the class average principle on the MOT17 validation set.}. We evaluate various long-tail distribution solutions in multiple classes on the MOT17 validation set. As shown in \cref{tab:Comparison of the different methods for improving long-tail distribution on the MOT17 validation set.}, we can observe that some methods boost the MOTA metric but reduce the IDF1 metric on all classes, and the Logit Adjustment method boosts the performance on the all classes but reduces the performance on the tail classes. In comparison, our method achieves the best performance in all classes, head classes and tail classes. 

\begin{table}[t]
 \setlength{\abovecaptionskip}{-0.2cm} 
 \footnotesize 
 \begin{center}
  \setlength{\tabcolsep}{0.75pt} 
  \renewcommand{\arraystretch}{1.0}
  {
  \begin{tabular}{l | cc | cc | cc}
    \toprule
    \multicolumn{1}{l|}{ \multirow{2}*{Method} } & \multicolumn{2}{c|}{All classes} & \multicolumn{2}{c|}{Head classes} & \multicolumn{2}{c}{Tail classes} \\
    \cline{2-7} 
    \multicolumn{1}{l|}{}\rule{0pt}{8pt}& MOTA↑ & IDF1↑ & MOTA↑ & IDF1↑ & MOTA↑ & IDF1↑ \\
    \toprule
    Base & 67.8 & 72.3 & 69.3 & 74.0 & 28.6 & 56.9 \\
    Base(+Focal Loss)   & 68.7 & 70.8 & 70.0 & 72.2 & 26.2 & 56.8  \\
    Base(+Triplet Loss)  & 68.4 & 65.7 & 69.7 & 67.7 & 23.3 & 51.3 \\
    Base(+CB Loss)   & 67.9 & 72.0 & 69.3 & 73.3 & 27.7 & 55.7 \\
    Base(+Logit Adjustment)  & 68.1 & 72.1 & 69.6 & 74.0 & 26.8 & 55.8 \\
    \rowcolor{aliceblue!200} Base(+\textbf{Ours})   & \textbf{69.3} & \textbf{73.4} & \textbf{70.9} & \textbf{75.2} & \textbf{29.6} & \textbf{57.2} \\
   \bottomrule
  \end{tabular}}
 \end{center}
 \caption{Comparison of different methods for improving long-tail distribution on the MOT17 validation set. The
best results are shown in \textbf{bold}. Our method is highlighted in \sethlcolor{aliceblue!200}\hl{blue}.}
 \label{tab:Comparison of the different methods for improving long-tail distribution on the MOT17 validation set.}
\end{table}

\subsection{Comparison with other SOTA} 
\quad Due to the fact that Joint Detection and Tracking involve the concurrent learning of detectors and appearance extractors, it is highly suitable for evaluating our method aimed at training data and Re-ID. To evaluate the effectiveness of our method, we apply them to two state-of-the-art trackers of the Joint Detection and Tracking, FairMOT~\cite{zhang2021fairmot} and CSTrack~\cite{liang2022rethinking}. We evaluate FairMOT and CSTrack on four public MOT benchmarks, i.e., MOT15, MOT16, MOT17 and MOT20. The results are reported in \cref{tab:State-of-the-art}. According to the results, our method can improve the performance of the algorithm on MOTA, IDF1, HOTA and other metrics, especially on the MOT15 and MOT20 benchmarks.

\begin{table}[t]
 \setlength{\abovecaptionskip}{-0.2cm} 
 \setlength{\belowcaptionskip}{-0.2cm}
 \footnotesize
 \begin{center}
  \setlength{\tabcolsep}{0.4mm}{
  \begin{tabular}{lllllll}
   \toprule
    Method & HOTA↑ & IDF1↑ & MOTA↑ & MT↑ & ML↓ & IDS↓ \\
   \midrule
   \multicolumn{7}{@{}l}{\textit{MOT15}} \\ 
   FairMOT~\cite{zhang2021fairmot}  & 45.2 & 59.4 & 53.1 & 39.0\% & 15.4\% & 911  \\
   \rowcolor{aliceblue!200} FairMOT(\textbf{+Ours})  & \textbf{47.0}(\textcolor{red}{+1.8}) & \textbf{61.5}(\textcolor{red}{+2.1}) & \textbf{56.7}(\textcolor{red}{+3.6})  & \textbf{42.7\%} & \textbf{14.0\%} & \textbf{845}  \\
   \midrule
   \multicolumn{7}{@{}l}{\textit{MOT15}} \\ 
   CSTrack~\cite{liang2022rethinking}  & 42.9 & 56.7 & 49.8 & 20.9\%  & 27.2\% & 762  \\
   \rowcolor{aliceblue!200} CSTrack(\textbf{+Ours})  & \textbf{46.4}(\textcolor{red}{+3.5}) & \textbf{60.2}(\textcolor{red}{+3.5}) & \textbf{55.0}(\textcolor{red}{+5.2}) & \textbf{46.7\%} & \textbf{13.0\%} & 920  \\ 
   \midrule
   \multicolumn{7}{@{}l}{\textit{MOT16}}\\
   FairMOT~\cite{zhang2021fairmot}   & 57.7 & 70.8 & 71.0  & 38.2\% & 21.6\% & 1,274  \\
   \rowcolor{aliceblue!200} FairMOT(\textbf{+Ours})  & \textbf{57.8}(\textcolor{red}{+0.1}) & \textbf{71.3}(\textcolor{red}{+0.5}) & \textbf{71.0}(\textcolor{red}{+0}) & \textbf{38.9\%}  & \textbf{21.5\%} & \textbf{1,270}  \\
   \midrule
   \multicolumn{7}{@{}l}{\textit{MOT16}}\\
   CSTrack~\cite{liang2022rethinking} & 53.1 & 67.5 & 65.7 & 30.4\% & 25.0\% & 1,265  \\
   \rowcolor{aliceblue!200} CSTrack(\textbf{+Ours})   & \textbf{53.3}(\textcolor{red}{+0.2}) & 67.5(\textcolor{red}{+0}) & \textbf{66.0}(\textcolor{red}{+0.3}) & 29.5\%  & 25.4\% & 1,303 \\
   \midrule
   \multicolumn{7}{@{}l}{\textit{MOT17}}\\
   FairMOT~\cite{zhang2021fairmot}  & 57.1 & 70.2 & 70.2 & 40.9\% & 18.6\% & 4,329  \\
   \rowcolor{aliceblue!200} FairMOT(\textbf{+Ours})  & \textbf{57.3}(\textcolor{red}{+0.2}) & 70.1(-0.1) & 69.9(-0.3) & \textbf{41.4\%} & \textbf{17.8\%} & 4,776  \\
   \midrule
   \multicolumn{7}{@{}l}{\textit{MOT17}}\\
   CSTrack~\cite{liang2022rethinking} & 52.6 & 66.1 & 65.2 & 30.7\% & 23.8\% & 4,605 \\
   \rowcolor{aliceblue!200} CSTrack(\textbf{+Ours})    & \textbf{53.1}(\textcolor{red}{+0.5}) & \textbf{66.7}(\textcolor{red}{+0.6}) & 65.1(-0.1) & \textbf{32.0\%}  & 25.0\% & \textbf{4,341}  \\
   \midrule
   \multicolumn{7}{@{}l}{\textit{MOT20}}\\
   FairMOT~\cite{zhang2021fairmot} & 52.3 & 65.0 & 56.8 & 67.2\% & 7.3\% & 6,108  \\
   \rowcolor{aliceblue!200} FairMOT(\textbf{+Ours})   & \textbf{54.4}(\textcolor{red}{+2.1}) & \textbf{70.3}(\textcolor{red}{+5.3}) & \textbf{65.9}(\textcolor{red}{\textbf{+9.1}}) & 49.7\% & 12.5\% & \textbf{3,548}  \\
   \midrule
   \multicolumn{7}{@{}l}{\textit{MOT20}}\\
   CSTrack~\cite{liang2022rethinking} & 45.7 & 59.9 & 58.1 & 33.7\% & 21.8\% & 3,645 \\
   \rowcolor{aliceblue!200} CSTrack(\textbf{+Ours})   & \textbf{47.1}(\textcolor{red}{+1.4}) & \textbf{60.4}(\textcolor{red}{+0.5}) & \textbf{58.1}(\textcolor{red}{+0}) & \textbf{35.5\%}  & \textbf{20.9\%} & 4,358  \\
   \bottomrule
  \end{tabular}}
 \end{center}
 \caption{State-of-the-art comparisons on four public MOT benchmarks, i.e., MOT15, MOT16, MOT17 and MOT20. Performance under the private detection on the test set of four public MOT benchmarks, only using themselves train set. All the results are obtained from the official MOT challenge evaluation server. Our better results are marked in \textbf{bold}. The gain vlaues are marked in \textcolor{red}{red}. The best gain vlaue is marked in \textbf{\textcolor{red}{red}}. Our method is highlighted in \sethlcolor{aliceblue!200}\hl{blue}.}
 \label{tab:State-of-the-art}
 \vspace{-1mm}
\end{table}

\noindent \textbf{MOT15:} 
As shown in \cref{tab:State-of-the-art}, the superiority of our method can be fully reflected on the benchmark MOT15. FairMOT is improved by 1.8\% HOTA, 2.1\% IDF1 and 3.6\% MOTA. CSTrack is improved by 3.5\% HOTA, 3.5\% IDF1, 5.2\% MOTA and 25.8\% MT, decreased by 14.2\% ML. 
This demonstrates that our method excellently improves the performance of detection and appearance feature extraction.

\noindent \textbf{MOT16 and MOT17:} 
Compared with MOT15, MOT16 and MOT17 contain more data and more precised annotations. The results in \cref{tab:State-of-the-art} show that 
FairMOT is improved by 0.1\% HOTA on MOT16 and 0.2\% HOTA on MOT17, and CSTrack is improved by 0.2\% HOTA on MOT16 and 0.5\% HOTA on MOT17, after adding our method. HOTA is a comprehensive index that balances detection performance and association performance. This shows that our method can improve the comprehensive ability of the network.

\noindent \textbf{MOT20:} 
Compared with previous MOT benchmarks, MOT20 is more crowded. As shown in \cref{tab:State-of-the-art}, our method delivers extremely outstanding results, with improvements of 2.1\% HOTA, 5.3\% IDF1 and 9.1\% MOTA for FairMOT and 1.4\% HOTA for CSTrack. We speculate that it is because the number of head classes and tail classes in the MOT20 dataset is very different, and our method alleviates the negative impact of the long tail distribution of the MOT20 dataset. Our method achieves extremely superior performance in dense pedestrian scenes.

\begin{table}[t]
 \setlength{\abovecaptionskip}{-0.2cm} 
 \setlength{\belowcaptionskip}{-0.15cm} 
 \footnotesize
 \begin{center}
  \setlength{\tabcolsep}{3.7pt} 
  \renewcommand{\arraystretch}{1.0}
  {
  \begin{tabular}{l | c | cccc}
    \toprule
    Method & Venue & MOTA↑ & IDF1↑ & HOTA↑ & IDS↓ \\
    \toprule
    \multicolumn{5}{l}{\textit{Joint Detection and Tracking framework:}} & \\
    FairMOT~\cite{zhang2021fairmot} & IJCV 2021 & 61.8 & 67.3 & 54.6 & 5,243 \\
    CSTrack~\cite{liang2022rethinking} & TIP 2022 & 66.6 & 68.6 & 54.0 & \textbf{3,196} \\
    RelationTrack~\cite{yu2022relationtrack} & TMM 2022 & 67.2 & 70.5 & 55.1 & 4,243 \\
    MTrack~\cite{yu2022towards} & CVPR 2022 & 63.5 & 69.2 & - & 6,031 \\
    \rowcolor{aliceblue!200} FairMOT(\textbf{+Ours}) & - & \textbf{67.8} & \textbf{70.7} & \textbf{55.4} & 3,505 \\
   \bottomrule
  \end{tabular}}
 \end{center}
 \caption{Comparison with the SOTA methods of the Joint Detection and Tracking framework on the MOT20 test set. The
best results are shown in \textbf{bold}. Our method is highlighted in \sethlcolor{aliceblue!200}\hl{blue}.}
 \label{tab:Comparison with the SOTA methods on the MOT20 test set.}
\end{table}

Then, we evaluate the SOTA methods of the Joint Detection and Tracking framework on the MOT20 test set. As shown in \cref{tab:Comparison with the SOTA methods on the MOT20 test set.}, our method achieves the best performance on comprehensive metrics.

\begin{table}[t]
 \setlength{\abovecaptionskip}{-0.2cm}
 \setlength{\belowcaptionskip}{-0.1cm}
 \footnotesize
 \begin{center}
  \setlength{\tabcolsep}{1.0pt}
  \renewcommand{\arraystretch}{1.2}
  {
  \begin{tabular}{l | c | cc | ccc}
    \toprule
    Method & Training Data & Images & Identities & MOTA↑ & IDF1↑ & IDS↓ \\
    \toprule
    FairMOT~\cite{zhang2021fairmot} & MIX & 77K & 10.4K & 61.8 & 67.3 & 5,243 \\
    \rowcolor{aliceblue!200} FairMOT(\textbf{+Ours}) & MIX & 77K & 10.4K & \textbf{67.8} & \textbf{70.7} & \textbf{3,505} \\
    \rowcolor{aliceblue!200} FairMOT(\textbf{+Ours}) & MOT20 & \textbf{9}K & \textbf{2.2}K & \textbf{65.9} & \textbf{70.3} & \textbf{3,548} \\
   \bottomrule
  \end{tabular}}
 \end{center}
 \caption{Results of the MOT20 test set when using different methods and different datasets for training. ``MIX'' represents the mixed datasets, including MOT20~\cite{dendorfer2020mot20}, ETH~\cite{ess2008mobile}, CityPerson~\cite{zhang2017citypersons}, CalTech~\cite{dollar2009pedestrian}, CUHK-SYSU~\cite{xiao2017joint}, PRW~\cite{zheng2017person} and CrowdHuman~\cite{shao2018crowdhuman} dataset. Our
better results are shown in \textbf{bold}. Our method is highlighted in \sethlcolor{aliceblue!200}\hl{blue}.}
 \label{tab:different methods and different datasets}
 \vspace{-1mm}
\end{table}

Furthermore, to explore the data efficiency benefits of our approach. We used different methods to train models on different data, and the results are shown in \cref{tab:different methods and different datasets}. Notably, our method trained using only the MOT20 data is 4.1\% higher in MOTA and 3.0\% higher in IDF1 than the baseline method trained using MIX data, indicating that our method is particularly effective for data efficiency.

\subsection{Ablation Study} 
\quad Using FairMOT~\cite{zhang2021fairmot} as the baseline tracker, we perform a series of ablation study on MOT17 dataset to demonstrate the effectiveness of our method from different aspects. Since MOTChallenge does not provide the validation set, we divide the training datasets of MOT17 into two parts, the first one half as training set and the other half as validation set. the training datasets of MOT17 include 02, 04, 05, 09, 10, 11 and 13 sequences. 02, 04 and 09 sequences belong to stationary camera view data. 05, 10, 11 and 13 sequences belong to dynamic camera view data. Thus, we apply SVA to 02, 04 and 09 sequences, and DVA to 05, 10, 11 and 13 sequences. All the models are trained for 30 epochs on the training set of MOT17.

\noindent \textbf{Impact of each component.}
As shown in \cref{tab:Impact}, all the components have boosted the tracking performance effectively. Our method can obtain 1.5\% MOTA and 1.1\% IDF1 gains (\normalsize{\textcircled{\scriptsize{1}}}\normalsize \enspace vs. \normalsize{\textcircled{\scriptsize{5}}}\normalsize). Among them, the SVA increases 0.5\% MOTA and 0.5\% IDF1, and DVA can further improve MOTA to 69.0\% and IDF1 to 73.0\%. Due to GS, our method boosts the performance to 69.3\% on MOTA and 73.4\% on IDF1.

\begin{table}[t]
\setlength{\abovecaptionskip}{-0.2cm}
\setlength{\belowcaptionskip}{-0.15cm} 
\begin{center}
  \setlength  \tabcolsep{1.55pt}
    \begin{tabular}{l ccc cc}
    \toprule
    Method & SVA & DVA & GS & MOTA↑ & IDF1↑ \\
    \toprule
    \Circled{1} Base   & \ding{56} & \ding{56} & \ding{56}  & 67.8  & 72.3 \\ 
    \Circled{2} Base(+SVA) & \ding{52} & \ding{56} & \ding{56} & 68.3  & 72.8 \\
    \Circled{3} Base(+DVA)  & \ding{56} & \ding{52}  & \ding{56} & 68.8  & 72.7  \\
    \Circled{4} Base(+SVA+DVA)  & \ding{52} & \ding{52}  & \ding{56} & 69.0  & 73.0  \\
    \Circled{5} Base(+SVA+DVA+GS)   & \ding{52} & \ding{52}  & \ding{52} & \textbf{69.3} & \textbf{73.4}  \\
    \bottomrule
    \end{tabular}
\end{center}
    \caption{Impact of each proposed component on validation set of MOT17. (SVA: Stationary Camera View Data Augmentation, DVA: Dynamic Camera View data Augmentation, GS: Group Softmax. The best results are marked in \textbf{bold}.)}
    \label{tab:Impact}
\end{table}

\begin{table}[t]
 \setlength{\abovecaptionskip}{-0.2cm}
 \setlength{\belowcaptionskip}{0.3cm}
 \footnotesize
 \begin{center}
  \setlength{\tabcolsep}{5.0pt}
  \renewcommand{\arraystretch}{1.0}
  {
  \begin{tabular}{cccccc}
   \toprule
    $T_s$ & MOTA↑ & IDF1↑ & FP↓ & FN↓ & IDS↓ \\
   \midrule
    0.0 & 68.0 & 70.2 & \textcolor{red}{2,511} & 14,329 & 475 \\
    0.1 & 68.0 & 70.2 & 2,945 & 13,868 & 462 \\
    0.2 & 68.4 & 70.2 & 2,778 & 13,810 & 461 \\
    0.3 & 68.1 & 69.7 & 2,670 & 14,090 & 479 \\
    0.4 & 68.1 & 71.3 & 2,991 & 13,787 & \textcolor{red}{451} \\
    0.5 & 68.5 & 70.8 & 2,663 & 13,844 & 495 \\
    0.6 & 68.0 & 69.8 & 2,785 & 14,016 & 515 \\
    0.7 & \textcolor{blue}{69.0} & \textcolor{blue}{72.5} & 2,685 & 13,591 & 470 \\
    0.8 & \textcolor{blue}{69.0} & 70.6 & 2,670 & 13,609 & \textcolor{blue}{453} \\
    1.0 & 68.3 & 71.0 & 3,224 & \textcolor{red}{13,426} & 458 \\
    \midrule
    0.9 & \textcolor{red}{69.3} & \textcolor{red}{73.4} & \textcolor{blue}{2,640} & \textcolor{blue}{13,504} & 455 \\
   \bottomrule
  \end{tabular}}
 \end{center}
 \caption{Comparison of different image selection thresholds $T_s$. The top two results are highlighted with \textcolor{red}{red} and \textcolor{blue}{blue}.} 
 \label{tab:image selection threshold $T_s$}
 \vspace{-4mm}
\end{table} 

\begin{figure}[htbp]
 \setlength{\abovecaptionskip}{0.3cm} 
 \setlength{\belowcaptionskip}{-0.1cm} 
    \centering
    \subfloat[Baseline]{\label{fig:a}\includegraphics[width=3.27in]{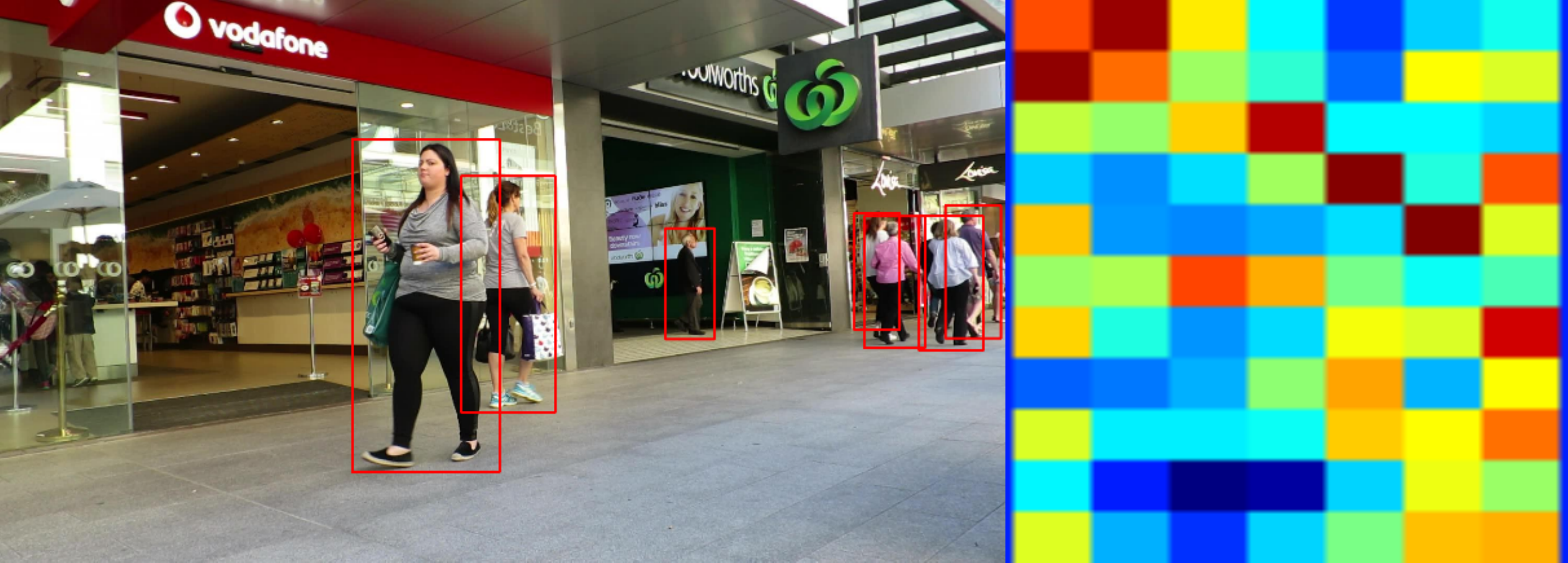}}\\
    \subfloat[Baseline+GS]{\label{fig:b}\includegraphics[width=3.27in]{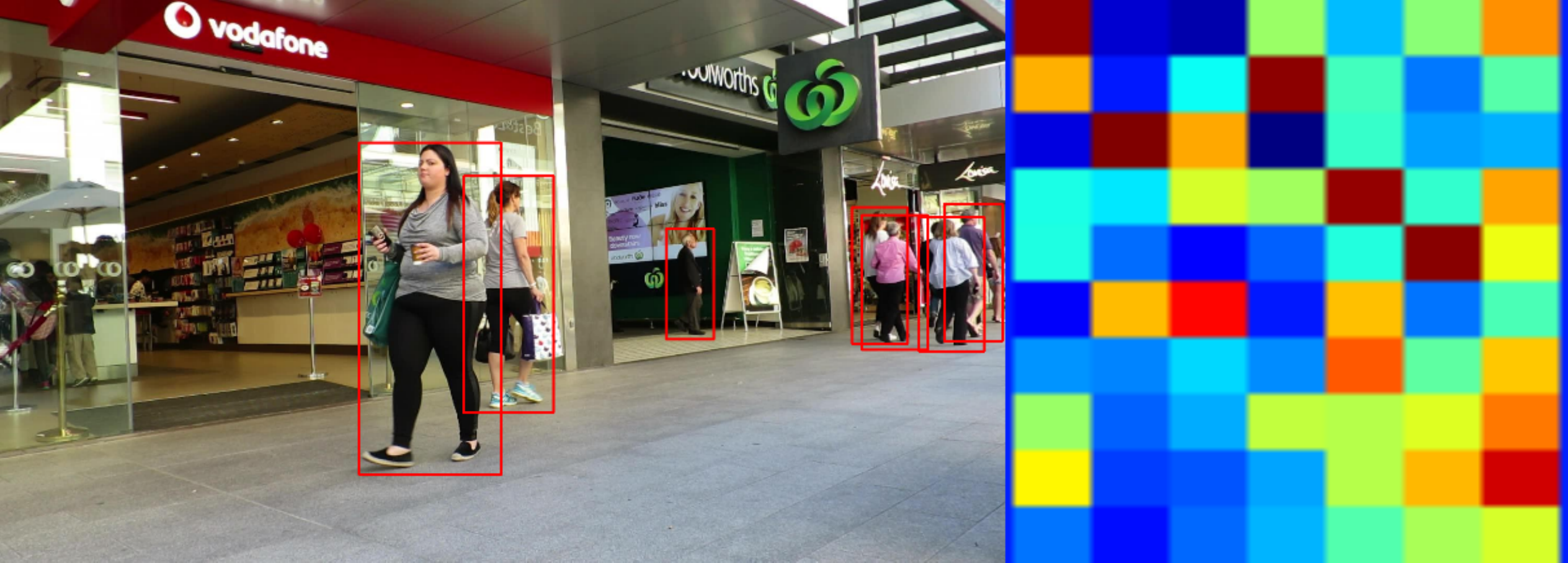}}\\
    \caption{Visualization about the cosine metric matrix between appearance vectors of the current example frame and the track templates. Red color indicates a higher association, and blue indicates a lesser association.}
    \label{fig:Re-ID embeddings}
\end{figure}

\noindent \textbf{Analysis of DVA.} 
As introduced in \cref{sec:DVA}. We select the image selection threshold $T_s$ in \cref{eq:T_p} to analyze its impact. We change the image selection threshold $T_s$ from 0.1 to 1.0 at intervals of 0.1. The results by using different image selection threshold $T_s$ are illustrated in \cref{tab:image selection threshold $T_s$}. We can observe that the best experimental results occur when the image selection threshold $T_s$ is 0.9. This phenomenon shows that using a small amount of enhanced images by DVA can help improve the performance of network. We speculate that the reason is that the network learns too much about the features of the enhanced image by DVA, which leads to ignoring the scene features of the original image.

\noindent \textbf{Analysis of GS.}
As introduced in \cref{sec:GS}. 
For a intuitive comparison, we construct the cosine metric matrix between appearance vectors of the current example frame and the track templates in \cref{fig:Re-ID embeddings}. 
During the matching process, the ideal situation is that there is at most one red color in each row and column of the cosine metric matrix, and the rest are blue. 
We can observe that the GS can significantly improve the association ability.

\section{Conclusion}
\label{sec:Conclusion}

\quad
In this study, we note a significant imbalance in the distribution of trajectory lengths across different pedestrians, revealing the long-tail distribution issue within existing MOT datasets.
To address it, we propose our method, which focuses on two key aspects: information augmentation and module improvement. Specifically, we introduce two data augmentation approach tailored for viewpoint states, including SVA and DVA, and the GS module for Re-ID. Notably, our work represents the pioneering effort in tackling the long-tail distribution in the realm of MOT. Using two SOTA multi-object trackers, we have verified the effectiveness of our method on MOTChallenge benchmarks. The experimental results demonstrate that our method effectively mitigates the impact of long-tail distribution on MOT.

\section*{Acknowledgements}
\label{sec:Acknowledgements}
\quad
This work was supported by the National Natural Science Foundation of China under Grant 62176096.

{
    \small
    \bibliographystyle{ieeenat_fullname}
    \bibliography{main}

\begin{thebibliography}{69}
\providecommand{\natexlab}[1]{#1}
\providecommand{\url}[1]{\texttt{#1}}
\expandafter\ifx\csname urlstyle\endcsname\relax
  \providecommand{\doi}[1]{doi: #1}\else
  \providecommand{\doi}{doi: \begingroup \urlstyle{rm}\Url}\fi

\bibitem[Antoniou et~al.(2017)Antoniou, Storkey, and Edwards]{antoniou2017data}
Antreas Antoniou, Amos Storkey, and Harrison Edwards.
\newblock Data augmentation generative adversarial networks.
\newblock \emph{arXiv preprint arXiv:1711.04340}, 2017.

\bibitem[Bashar et~al.(2022)Bashar, Islam, Hussain, Hasan, Rahman, and Kabir]{bashar2022multiple}
Mk Bashar, Samia Islam, Kashifa~Kawaakib Hussain, Md~Bakhtiar Hasan, ABM Rahman, and Md~Hasanul Kabir.
\newblock Multiple object tracking in recent times: a literature review.
\newblock \emph{arXiv preprint arXiv:2209.04796}, 2022.

\bibitem[Bergmann et~al.(2019)Bergmann, Meinhardt, and Leal-Taixe]{bergmann2019tracking}
Philipp Bergmann, Tim Meinhardt, and Laura Leal-Taixe.
\newblock Tracking without bells and whistles.
\newblock In \emph{Proceedings of the IEEE/CVF International Conference on Computer Vision}, pages 941--951, 2019.

\bibitem[Bernardin and Stiefelhagen(2008)]{bernardin2008evaluating}
Keni Bernardin and Rainer Stiefelhagen.
\newblock Evaluating multiple object tracking performance: the clear mot metrics.
\newblock \emph{EURASIP Journal on Image and Video Processing}, 2008:\penalty0 1--10, 2008.

\bibitem[Bertalmio et~al.(2001)Bertalmio, Bertozzi, and Sapiro]{bertalmio2001navier}
Marcelo Bertalmio, Andrea~L Bertozzi, and Guillermo Sapiro.
\newblock Navier-stokes, fluid dynamics, and image and video inpainting.
\newblock In \emph{Proceedings of the 2001 IEEE Computer Society Conference on Computer Vision and Pattern Recognition. CVPR 2001}, pages I--I. IEEE, 2001.

\bibitem[Bewley et~al.(2016)Bewley, Ge, Ott, Ramos, and Upcroft]{bewley2016simple}
Alex Bewley, Zongyuan Ge, Lionel Ott, Fabio Ramos, and Ben Upcroft.
\newblock Simple online and realtime tracking.
\newblock In \emph{2016 IEEE international conference on image processing (ICIP)}, pages 3464--3468. IEEE, 2016.

\bibitem[Cao et~al.(2020)Cao, Zhu, Huang, Guo, and Lei]{cao2020domain}
Dong Cao, Xiangyu Zhu, Xingyu Huang, Jianzhu Guo, and Zhen Lei.
\newblock Domain balancing: Face recognition on long-tailed domains.
\newblock In \emph{Proceedings of the IEEE/CVF Conference on Computer Vision and Pattern Recognition}, pages 5671--5679, 2020.

\bibitem[Cao et~al.(2019)Cao, Wei, Gaidon, Arechiga, and Ma]{cao2019learning}
Kaidi Cao, Colin Wei, Adrien Gaidon, Nikos Arechiga, and Tengyu Ma.
\newblock Learning imbalanced datasets with label-distribution-aware margin loss.
\newblock \emph{Advances in neural information processing systems}, 32, 2019.

\bibitem[Chou et~al.(2020)Chou, Chang, Pan, Wei, and Juan]{chou2020remix}
Hsin-Ping Chou, Shih-Chieh Chang, Jia-Yu Pan, Wei Wei, and Da-Cheng Juan.
\newblock Remix: rebalanced mixup.
\newblock In \emph{Computer Vision--ECCV 2020 Workshops: Glasgow, UK, August 23--28, 2020, Proceedings, Part VI 16}, pages 95--110. Springer, 2020.

\bibitem[Cui et~al.(2021)Cui, Zhong, Liu, Yu, and Jia]{cui2021parametric}
Jiequan Cui, Zhisheng Zhong, Shu Liu, Bei Yu, and Jiaya Jia.
\newblock Parametric contrastive learning.
\newblock In \emph{Proceedings of the IEEE/CVF international conference on computer vision}, pages 715--724, 2021.

\bibitem[Cui et~al.(2019)Cui, Jia, Lin, Song, and Belongie]{cui2019class}
Yin Cui, Menglin Jia, Tsung-Yi Lin, Yang Song, and Serge Belongie.
\newblock Class-balanced loss based on effective number of samples.
\newblock In \emph{Proceedings of the IEEE/CVF conference on computer vision and pattern recognition}, pages 9268--9277, 2019.

\bibitem[Dendorfer et~al.(2020)Dendorfer, Rezatofighi, Milan, Shi, Cremers, Reid, Roth, Schindler, and Leal-Taix{\'e}]{dendorfer2020mot20}
Patrick Dendorfer, Hamid Rezatofighi, Anton Milan, Javen Shi, Daniel Cremers, Ian Reid, Stefan Roth, Konrad Schindler, and Laura Leal-Taix{\'e}.
\newblock Mot20: A benchmark for multi object tracking in crowded scenes.
\newblock \emph{arXiv preprint arXiv:2003.09003}, 2020.

\bibitem[Desai et~al.(2021)Desai, Wu, Tripathi, and Vasconcelos]{desai2021learning}
Alakh Desai, Tz-Ying Wu, Subarna Tripathi, and Nuno Vasconcelos.
\newblock Learning of visual relations: The devil is in the tails.
\newblock In \emph{Proceedings of the IEEE/CVF International Conference on Computer Vision}, pages 15404--15413, 2021.

\bibitem[Doll{\'a}r et~al.(2009)Doll{\'a}r, Wojek, Schiele, and Perona]{dollar2009pedestrian}
Piotr Doll{\'a}r, Christian Wojek, Bernt Schiele, and Pietro Perona.
\newblock Pedestrian detection: A benchmark.
\newblock In \emph{2009 IEEE conference on computer vision and pattern recognition}, pages 304--311. IEEE, 2009.

\bibitem[Du et~al.(2023)Du, Zhao, Song, Zhao, Su, Gong, and Meng]{du2023strongsort}
Yunhao Du, Zhicheng Zhao, Yang Song, Yanyun Zhao, Fei Su, Tao Gong, and Hongying Meng.
\newblock Strongsort: Make deepsort great again.
\newblock \emph{IEEE Transactions on Multimedia}, 2023.

\bibitem[Ess et~al.(2008)Ess, Leibe, Schindler, and Van~Gool]{ess2008mobile}
Andreas Ess, Bastian Leibe, Konrad Schindler, and Luc Van~Gool.
\newblock A mobile vision system for robust multi-person tracking.
\newblock In \emph{2008 IEEE Conference on Computer Vision and Pattern Recognition}, pages 1--8. IEEE, 2008.

\bibitem[Feng et~al.(2021)Feng, Zhong, and Huang]{feng2021exploring}
Chengjian Feng, Yujie Zhong, and Weilin Huang.
\newblock Exploring classification equilibrium in long-tailed object detection.
\newblock In \emph{Proceedings of the IEEE/CVF International conference on computer vision}, pages 3417--3426, 2021.

\bibitem[Fischer et~al.(2023)Fischer, Huang, Pang, Qiu, Chen, Darrell, and Yu]{fischer2023qdtrack}
Tobias Fischer, Thomas~E Huang, Jiangmiao Pang, Linlu Qiu, Haofeng Chen, Trevor Darrell, and Fisher Yu.
\newblock Qdtrack: Quasi-dense similarity learning for appearance-only multiple object tracking.
\newblock \emph{IEEE Transactions on Pattern Analysis and Machine Intelligence}, 2023.

\bibitem[Guo and Wang(2021)]{guo2021long}
Hao Guo and Song Wang.
\newblock Long-tailed multi-label visual recognition by collaborative training on uniform and re-balanced samplings.
\newblock In \emph{Proceedings of the IEEE/CVF Conference on Computer Vision and Pattern Recognition}, pages 15089--15098, 2021.

\bibitem[Han et~al.(2022)Han, Huang, Wang, Yu, Liu, and Pan]{han2022mat}
Shoudong Han, Piao Huang, Hongwei Wang, En Yu, Donghaisheng Liu, and Xiaofeng Pan.
\newblock Mat: Motion-aware multi-object tracking.
\newblock \emph{Neurocomputing}, 476:\penalty0 75--86, 2022.

\bibitem[He et~al.(2019)He, Girshick, and Doll{\'a}r]{he2019rethinking}
Kaiming He, Ross Girshick, and Piotr Doll{\'a}r.
\newblock Rethinking imagenet pre-training.
\newblock In \emph{Proceedings of the IEEE/CVF International Conference on Computer Vision}, pages 4918--4927, 2019.

\bibitem[He et~al.(2020)He, Fan, Wu, Xie, and Girshick]{he2020momentum}
Kaiming He, Haoqi Fan, Yuxin Wu, Saining Xie, and Ross Girshick.
\newblock Momentum contrast for unsupervised visual representation learning.
\newblock In \emph{Proceedings of the IEEE/CVF conference on computer vision and pattern recognition}, pages 9729--9738, 2020.

\bibitem[He et~al.(2022)He, Sun, Yu, Xue, Zhang, Torr, Bai, and Qi]{he2022synthetic}
Ruifei He, Shuyang Sun, Xin Yu, Chuhui Xue, Wenqing Zhang, Philip Torr, Song Bai, and Xiaojuan Qi.
\newblock Is synthetic data from generative models ready for image recognition?
\newblock \emph{arXiv preprint arXiv:2210.07574}, 2022.

\bibitem[Kang et~al.(2019)Kang, Xie, Rohrbach, Yan, Gordo, Feng, and Kalantidis]{kang2019decoupling}
Bingyi Kang, Saining Xie, Marcus Rohrbach, Zhicheng Yan, Albert Gordo, Jiashi Feng, and Yannis Kalantidis.
\newblock Decoupling representation and classifier for long-tailed recognition.
\newblock \emph{arXiv preprint arXiv:1910.09217}, 2019.

\bibitem[Kang et~al.(2020)Kang, Li, Xie, Yuan, and Feng]{kang2020exploring}
Bingyi Kang, Yu Li, Sa Xie, Zehuan Yuan, and Jiashi Feng.
\newblock Exploring balanced feature spaces for representation learning.
\newblock In \emph{International Conference on Learning Representations}, 2020.

\bibitem[Kirillov et~al.(2023)Kirillov, Mintun, Ravi, Mao, Rolland, Gustafson, Xiao, Whitehead, Berg, Lo, et~al.]{kirillov2023segment}
Alexander Kirillov, Eric Mintun, Nikhila Ravi, Hanzi Mao, Chloe Rolland, Laura Gustafson, Tete Xiao, Spencer Whitehead, Alexander~C Berg, Wan-Yen Lo, et~al.
\newblock Segment anything.
\newblock \emph{arXiv preprint arXiv:2304.02643}, 2023.

\bibitem[Leal-Taix{\'e} et~al.(2015)Leal-Taix{\'e}, Milan, Reid, Roth, and Schindler]{leal2015motchallenge}
Laura Leal-Taix{\'e}, Anton Milan, Ian Reid, Stefan Roth, and Konrad Schindler.
\newblock Motchallenge 2015: Towards a benchmark for multi-target tracking.
\newblock \emph{arXiv preprint arXiv:1504.01942}, 2015.

\bibitem[Li et~al.(2021)Li, Gong, Liu, Wang, Qiao, and Cheng]{li2021metasaug}
Shuang Li, Kaixiong Gong, Chi~Harold Liu, Yulin Wang, Feng Qiao, and Xinjing Cheng.
\newblock Metasaug: Meta semantic augmentation for long-tailed visual recognition.
\newblock In \emph{Proceedings of the IEEE/CVF conference on computer vision and pattern recognition}, pages 5212--5221, 2021.

\bibitem[Li et~al.(2020)Li, Wang, Kang, Tang, Wang, Li, and Feng]{li2020overcoming}
Yu Li, Tao Wang, Bingyi Kang, Sheng Tang, Chunfeng Wang, Jintao Li, and Jiashi Feng.
\newblock Overcoming classifier imbalance for long-tail object detection with balanced group softmax.
\newblock In \emph{Proceedings of the IEEE/CVF conference on computer vision and pattern recognition}, pages 10991--11000, 2020.

\bibitem[Liang et~al.(2022)Liang, Zhang, Zhou, Li, Zhu, and Hu]{liang2022rethinking}
Chao Liang, Zhipeng Zhang, Xue Zhou, Bing Li, Shuyuan Zhu, and Weiming Hu.
\newblock Rethinking the competition between detection and reid in multiobject tracking.
\newblock \emph{IEEE Transactions on Image Processing}, 31:\penalty0 3182--3196, 2022.

\bibitem[Liu et~al.(2021)Liu, Li, Kang, Hua, and Vasconcelos]{liu2021gistnet}
Bo Liu, Haoxiang Li, Hao Kang, Gang Hua, and Nuno Vasconcelos.
\newblock Gistnet: a geometric structure transfer network for long-tailed recognition.
\newblock In \emph{Proceedings of the IEEE/CVF International Conference on Computer Vision}, pages 8209--8218, 2021.

\bibitem[Liu et~al.(2020)Liu, Sun, Han, Dou, and Li]{liu2020deep}
Jialun Liu, Yifan Sun, Chuchu Han, Zhaopeng Dou, and Wenhui Li.
\newblock Deep representation learning on long-tailed data: A learnable embedding augmentation perspective.
\newblock In \emph{Proceedings of the IEEE/CVF conference on computer vision and pattern recognition}, pages 2970--2979, 2020.

\bibitem[Meinhardt et~al.(2022)Meinhardt, Kirillov, Leal-Taixe, and Feichtenhofer]{meinhardt2022trackformer}
Tim Meinhardt, Alexander Kirillov, Laura Leal-Taixe, and Christoph Feichtenhofer.
\newblock Trackformer: Multi-object tracking with transformers.
\newblock In \emph{Proceedings of the IEEE/CVF conference on computer vision and pattern recognition}, pages 8844--8854, 2022.

\bibitem[Milan et~al.(2016)Milan, Leal-Taix{\'e}, Reid, Roth, and Schindler]{milan2016mot16}
Anton Milan, Laura Leal-Taix{\'e}, Ian Reid, Stefan Roth, and Konrad Schindler.
\newblock Mot16: A benchmark for multi-object tracking.
\newblock \emph{arXiv preprint arXiv:1603.00831}, 2016.

\bibitem[Ren et~al.(2020)Ren, Yu, Ma, Zhao, Yi, et~al.]{ren2020balanced}
Jiawei Ren, Cunjun Yu, Xiao Ma, Haiyu Zhao, Shuai Yi, et~al.
\newblock Balanced meta-softmax for long-tailed visual recognition.
\newblock \emph{Advances in neural information processing systems}, 33:\penalty0 4175--4186, 2020.

\bibitem[Rombach et~al.(2022)Rombach, Blattmann, Lorenz, Esser, and Ommer]{rombach2022high}
Robin Rombach, Andreas Blattmann, Dominik Lorenz, Patrick Esser, and Bj{\"o}rn Ommer.
\newblock High-resolution image synthesis with latent diffusion models.
\newblock In \emph{Proceedings of the IEEE/CVF conference on computer vision and pattern recognition}, pages 10684--10695, 2022.

\bibitem[Samuel and Chechik(2021)]{samuel2021distributional}
Dvir Samuel and Gal Chechik.
\newblock Distributional robustness loss for long-tail learning.
\newblock In \emph{Proceedings of the IEEE/CVF International Conference on Computer Vision}, pages 9495--9504, 2021.

\bibitem[Shao et~al.(2018)Shao, Zhao, Li, Xiao, Yu, Zhang, and Sun]{shao2018crowdhuman}
Shuai Shao, Zijian Zhao, Boxun Li, Tete Xiao, Gang Yu, Xiangyu Zhang, and Jian Sun.
\newblock Crowdhuman: A benchmark for detecting human in a crowd.
\newblock \emph{arXiv preprint arXiv:1805.00123}, 2018.

\bibitem[Sun et~al.(2020)Sun, Cao, Jiang, Zhang, Xie, Yuan, Wang, and Luo]{sun2020transtrack}
Peize Sun, Jinkun Cao, Yi Jiang, Rufeng Zhang, Enze Xie, Zehuan Yuan, Changhu Wang, and Ping Luo.
\newblock Transtrack: Multiple object tracking with transformer.
\newblock \emph{arXiv preprint arXiv:2012.15460}, 2020.

\bibitem[Tao et~al.(2023)Tao, Sun, Yang, Chen, Wang, Yang, Du, and Zheng]{tao2023local}
Yingfan Tao, Jingna Sun, Hao Yang, Li Chen, Xu Wang, Wenming Yang, Daniel Du, and Min Zheng.
\newblock Local and global logit adjustments for long-tailed learning.
\newblock In \emph{Proceedings of the IEEE/CVF International Conference on Computer Vision}, pages 11783--11792, 2023.

\bibitem[Tokmakov et~al.(2021)Tokmakov, Li, Burgard, and Gaidon]{tokmakov2021learning}
Pavel Tokmakov, Jie Li, Wolfram Burgard, and Adrien Gaidon.
\newblock Learning to track with object permanence.
\newblock In \emph{Proceedings of the IEEE/CVF International Conference on Computer Vision}, pages 10860--10869, 2021.

\bibitem[Wang et~al.(2023)Wang, Fei, Wang, Li, Bao, Wu, Zhao, and Shen]{wang2023balancing}
Yuchao Wang, Jingjing Fei, Haochen Wang, Wei Li, Tianpeng Bao, Liwei Wu, Rui Zhao, and Yujun Shen.
\newblock Balancing logit variation for long-tailed semantic segmentation.
\newblock In \emph{Proceedings of the IEEE/CVF Conference on Computer Vision and Pattern Recognition}, pages 19561--19573, 2023.

\bibitem[Wang et~al.(2020)Wang, Zheng, Liu, and Wang]{wang2019towards}
Zhongdao Wang, Liang Zheng, Yixuan Liu, and Shengjin Wang.
\newblock Towards real-time multi-object tracking.
\newblock \emph{The European Conference on Computer Vision (ECCV)}, 2020.

\bibitem[Wojke et~al.(2017)Wojke, Bewley, and Paulus]{wojke2017simple}
Nicolai Wojke, Alex Bewley, and Dietrich Paulus.
\newblock Simple online and realtime tracking with a deep association metric.
\newblock In \emph{2017 IEEE international conference on image processing (ICIP)}, pages 3645--3649. IEEE, 2017.

\bibitem[Xiang et~al.(2020)Xiang, Ding, and Han]{xiang2020learning}
Liuyu Xiang, Guiguang Ding, and Jungong Han.
\newblock Learning from multiple experts: Self-paced knowledge distillation for long-tailed classification.
\newblock In \emph{Computer Vision--ECCV 2020: 16th European Conference, Glasgow, UK, August 23--28, 2020, Proceedings, Part V 16}, pages 247--263. Springer, 2020.

\bibitem[Xiao et~al.(2017)Xiao, Li, Wang, Lin, and Wang]{xiao2017joint}
Tong Xiao, Shuang Li, Bochao Wang, Liang Lin, and Xiaogang Wang.
\newblock Joint detection and identification feature learning for person search.
\newblock In \emph{Proceedings of the IEEE conference on computer vision and pattern recognition}, pages 3415--3424, 2017.

\bibitem[Ye et~al.(2020)Ye, Chen, Zhan, and Chao]{ye2020identifying}
Han-Jia Ye, Hong-You Chen, De-Chuan Zhan, and Wei-Lun Chao.
\newblock Identifying and compensating for feature deviation in imbalanced deep learning.
\newblock \emph{arXiv preprint arXiv:2001.01385}, 2020.

\bibitem[Yin et~al.(2019)Yin, Yu, Sohn, Liu, and Chandraker]{yin2019feature}
Xi Yin, Xiang Yu, Kihyuk Sohn, Xiaoming Liu, and Manmohan Chandraker.
\newblock Feature transfer learning for face recognition with under-represented data.
\newblock In \emph{Proceedings of the IEEE/CVF conference on computer vision and pattern recognition}, pages 5704--5713, 2019.

\bibitem[Yu et~al.(2022{\natexlab{a}})Yu, Li, and Han]{yu2022towards}
En Yu, Zhuoling Li, and Shoudong Han.
\newblock Towards discriminative representation: Multi-view trajectory contrastive learning for online multi-object tracking.
\newblock In \emph{Proceedings of the IEEE/CVF Conference on Computer Vision and Pattern Recognition}, pages 8834--8843, 2022{\natexlab{a}}.

\bibitem[Yu et~al.(2022{\natexlab{b}})Yu, Li, Han, and Wang]{yu2022relationtrack}
En Yu, Zhuoling Li, Shoudong Han, and Hongwei Wang.
\newblock Relationtrack: Relation-aware multiple object tracking with decoupled representation.
\newblock \emph{IEEE Transactions on Multimedia}, 2022{\natexlab{b}}.

\bibitem[Yu et~al.(2023{\natexlab{a}})Yu, Liu, Li, Yang, Li, Han, and Tao]{yu2023generalizing}
En Yu, Songtao Liu, Zhuoling Li, Jinrong Yang, Zeming Li, Shoudong Han, and Wenbing Tao.
\newblock Generalizing multiple object tracking to unseen domains by introducing natural language representation.
\newblock In \emph{Proceedings of the AAAI Conference on Artificial Intelligence}, pages 3304--3312, 2023{\natexlab{a}}.

\bibitem[Yu et~al.(2023{\natexlab{b}})Yu, Wang, Li, Zhang, Zhang, and Tao]{yu2023motrv3}
En Yu, Tiancai Wang, Zhuoling Li, Yuang Zhang, Xiangyu Zhang, and Wenbing Tao.
\newblock Motrv3: Release-fetch supervision for end-to-end multi-object tracking.
\newblock \emph{arXiv preprint arXiv:2305.14298}, 2023{\natexlab{b}}.

\bibitem[Zang et~al.(2021)Zang, Huang, and Loy]{zang2021fasa}
Yuhang Zang, Chen Huang, and Chen~Change Loy.
\newblock Fasa: Feature augmentation and sampling adaptation for long-tailed instance segmentation.
\newblock In \emph{Proceedings of the IEEE/CVF International Conference on Computer Vision}, pages 3457--3466, 2021.

\bibitem[Zeng et~al.(2022)Zeng, Dong, Zhang, Wang, Zhang, and Wei]{zeng2022motr}
Fangao Zeng, Bin Dong, Yuang Zhang, Tiancai Wang, Xiangyu Zhang, and Yichen Wei.
\newblock Motr: End-to-end multiple-object tracking with transformer.
\newblock In \emph{European Conference on Computer Vision}, pages 659--675. Springer, 2022.

\bibitem[Zhang et~al.(2017)Zhang, Benenson, and Schiele]{zhang2017citypersons}
Shanshan Zhang, Rodrigo Benenson, and Bernt Schiele.
\newblock Citypersons: A diverse dataset for pedestrian detection.
\newblock In \emph{Proceedings of the IEEE conference on computer vision and pattern recognition}, pages 3213--3221, 2017.

\bibitem[Zhang et~al.(2021{\natexlab{a}})Zhang, Li, Yan, He, and Sun]{zhang2021distribution}
Songyang Zhang, Zeming Li, Shipeng Yan, Xuming He, and Jian Sun.
\newblock Distribution alignment: A unified framework for long-tail visual recognition.
\newblock In \emph{Proceedings of the IEEE/CVF conference on computer vision and pattern recognition}, pages 2361--2370, 2021{\natexlab{a}}.

\bibitem[Zhang et~al.(2021{\natexlab{b}})Zhang, Wu, Weng, Fu, Chen, Jiang, and Davis]{zhang2021videolt}
Xing Zhang, Zuxuan Wu, Zejia Weng, Huazhu Fu, Jingjing Chen, Yu-Gang Jiang, and Larry~S Davis.
\newblock Videolt: Large-scale long-tailed video recognition.
\newblock In \emph{Proceedings of the IEEE/CVF International Conference on Computer Vision}, pages 7960--7969, 2021{\natexlab{b}}.

\bibitem[Zhang et~al.(2021{\natexlab{c}})Zhang, Wang, Wang, Zeng, and Liu]{zhang2021fairmot}
Yifu Zhang, Chunyu Wang, Xinggang Wang, Wenjun Zeng, and Wenyu Liu.
\newblock Fairmot: On the fairness of detection and re-identification in multiple object tracking.
\newblock \emph{International Journal of Computer Vision}, 129:\penalty0 3069--3087, 2021{\natexlab{c}}.

\bibitem[Zhang et~al.(2022)Zhang, Sun, Jiang, Yu, Weng, Yuan, Luo, Liu, and Wang]{zhang2022bytetrack}
Yifu Zhang, Peize Sun, Yi Jiang, Dongdong Yu, Fucheng Weng, Zehuan Yuan, Ping Luo, Wenyu Liu, and Xinggang Wang.
\newblock Bytetrack: Multi-object tracking by associating every detection box.
\newblock In \emph{European Conference on Computer Vision}, pages 1--21. Springer, 2022.

\bibitem[Zhang et~al.(2023)Zhang, Wang, and Zhang]{zhang2023motrv2}
Yuang Zhang, Tiancai Wang, and Xiangyu Zhang.
\newblock Motrv2: Bootstrapping end-to-end multi-object tracking by pretrained object detectors.
\newblock In \emph{Proceedings of the IEEE/CVF Conference on Computer Vision and Pattern Recognition}, pages 22056--22065, 2023.

\bibitem[Zhao et~al.(2022)Zhao, Chen, Tan, Huang, and Zhu]{zhao2022adaptive}
Yan Zhao, Weicong Chen, Xu Tan, Kai Huang, and Jihong Zhu.
\newblock Adaptive logit adjustment loss for long-tailed visual recognition.
\newblock In \emph{Proceedings of the AAAI Conference on Artificial Intelligence}, pages 3472--3480, 2022.

\bibitem[Zheng et~al.(2017)Zheng, Zhang, Sun, Chandraker, Yang, and Tian]{zheng2017person}
Liang Zheng, Hengheng Zhang, Shaoyan Sun, Manmohan Chandraker, Yi Yang, and Qi Tian.
\newblock Person re-identification in the wild.
\newblock In \emph{Proceedings of the IEEE conference on computer vision and pattern recognition}, pages 1367--1376, 2017.

\bibitem[Zhong et~al.(2019)Zhong, Deng, Wang, Hu, Peng, Tao, and Huang]{zhong2019unequal}
Yaoyao Zhong, Weihong Deng, Mei Wang, Jiani Hu, Jianteng Peng, Xunqiang Tao, and Yaohai Huang.
\newblock Unequal-training for deep face recognition with long-tailed noisy data.
\newblock In \emph{Proceedings of the IEEE/CVF conference on computer vision and pattern recognition}, pages 7812--7821, 2019.

\bibitem[Zhong et~al.(2021)Zhong, Cui, Liu, and Jia]{zhong2021improving}
Zhisheng Zhong, Jiequan Cui, Shu Liu, and Jiaya Jia.
\newblock Improving calibration for long-tailed recognition.
\newblock In \emph{Proceedings of the IEEE/CVF conference on computer vision and pattern recognition}, pages 16489--16498, 2021.

\bibitem[Zhou et~al.(2020{\natexlab{a}})Zhou, Cui, Wei, and Chen]{zhou2020bbn}
Boyan Zhou, Quan Cui, Xiu-Shen Wei, and Zhao-Min Chen.
\newblock Bbn: Bilateral-branch network with cumulative learning for long-tailed visual recognition.
\newblock In \emph{Proceedings of the IEEE/CVF conference on computer vision and pattern recognition}, pages 9719--9728, 2020{\natexlab{a}}.

\bibitem[Zhou et~al.(2020{\natexlab{b}})Zhou, Koltun, and Kr{\"a}henb{\"u}hl]{zhou2020tracking}
Xingyi Zhou, Vladlen Koltun, and Philipp Kr{\"a}henb{\"u}hl.
\newblock Tracking objects as points.
\newblock In \emph{European conference on computer vision}, pages 474--490. Springer, 2020{\natexlab{b}}.

\bibitem[Zhou et~al.(2022)Zhou, Yin, Koltun, and Kr{\"a}henb{\"u}hl]{zhou2022global}
Xingyi Zhou, Tianwei Yin, Vladlen Koltun, and Philipp Kr{\"a}henb{\"u}hl.
\newblock Global tracking transformers.
\newblock In \emph{Proceedings of the IEEE/CVF Conference on Computer Vision and Pattern Recognition}, pages 8771--8780, 2022.

\bibitem[Zhu and Yang(2020)]{zhu2020inflated}
Linchao Zhu and Yi Yang.
\newblock Inflated episodic memory with region self-attention for long-tailed visual recognition.
\newblock In \emph{Proceedings of the IEEE/CVF Conference on Computer Vision and Pattern Recognition}, pages 4344--4353, 2020.

\bibitem[Zoph et~al.(2020)Zoph, Ghiasi, Lin, Cui, Liu, Cubuk, and Le]{zoph2020rethinking}
Barret Zoph, Golnaz Ghiasi, Tsung-Yi Lin, Yin Cui, Hanxiao Liu, Ekin~Dogus Cubuk, and Quoc Le.
\newblock Rethinking pre-training and self-training.
\newblock \emph{Advances in neural information processing systems}, 33:\penalty0 3833--3845, 2020.

\end{thebibliography}
}

\clearpage

\section{Appendix}
\label{sec:Appendix}
\quad

In this appendix, we provide more details about our method. Our method comprises 3 modules: Stationary Camera View Data Augmentation (SVA), Dynamic Camera View Data Augmentation (DVA), and Group Softmax (GS).
Specifically, \cref{sec:Data Efficiency} further explores the data utilization efficiency of our method. \cref{sec:Analysis of Performance Differences} analyzes the performance differences of our method on different datasets. \cref{sec:Additional Ablation Study} provides more experiments to confirm our method. \cref{sec:Visualization} shows many visualization examples. \cref{sec:Questions and Replies} shows many questions and replies.

\subsection{Data Efficiency}
\label{sec:Data Efficiency}

\quad
To further explore data utilization efficiency, we further use less data. We use the FairMOT algorithm added to our method to train on the last 3/4 and 1/2 of the MOT20 training set. In addition, in order to explore the outstanding performance of our method, we use the FairMOT algorithm added to our method to train on the mixed data, which is ``MIX'' data, and test on the MOT20 test set. We find that the FairMOT only trained for 20 epochs on the MOT20 when training on the mixed data.
Therefore, to compare fairness, we also test the model trained for 20 epochs.
As shown in \cref{tab:different methods and less data}, using the mixed data, our method achieves extremely better performance than the baseline on the MOT20 test set, with a 6.0\% improvement in MOTA, a 3.4\% improvement in IDF1, and a 1738 reduction in IDS. In addition, using only 5.2\% of the mixed image data, our method achieves comparable performance to the baseline trained on the mixed data on the MOT20 test set.

\begin{table*}
\hsize=\textwidth
\centering
 \setlength{\abovecaptionskip}{-0.15cm}  
 \setlength{\belowcaptionskip}{+0cm}  
 \footnotesize
 \begin{center}
  \setlength{\tabcolsep}{3.5pt}
  \renewcommand{\arraystretch}{1.2} 
  {
  \begin{tabular}{l | c | rr | ccc | ccc}
    \toprule
    Method & Training Data & Images & Identities & CH training epochs & ECP training epochs & MOT20 training epochs & MOTA↑ & IDF1↑ & IDS↓ \\
    \toprule
    FairMOT*~\cite{zhang2021fairmot} & MIX & 77K & 10.4K & 30 & 60 & 20 & 61.8 & 67.3 & 5,243 \\
    FairMOT~\cite{zhang2021fairmot} & MOT20 & 9K & 2.2K & 0 & 0 & 20 & 57.5 & 66.1 & 5,793 \\
    FairMOT~\cite{zhang2021fairmot} & MOT20 & 9K & 2.2K & 0 & 0 & 30 & 56.8 & 65.9 & 6,108 \\
    \rowcolor{aliceblue!200} FairMOT(\textbf{+Ours}) & MIX & 77K & 10.4K & 30 & 60 & 20 & \textcolor{red}{\textbf{67.8}} & \textcolor{red}{\textbf{70.7}} & \textbf{3,505} \\
    \rowcolor{aliceblue!200} FairMOT(\textbf{+Ours}) & MOT20 & 9K & 2.2K & 0 & 0 & 20 & \textbf{66.6} & \textbf{70.4} & \textcolor{red}{\textbf{3,334}} \\
    \rowcolor{aliceblue!200} FairMOT(\textbf{+Ours}) & MOT20 & 9K & 2.2K & 0 & 0 & 30 & \textbf{65.9} & \textbf{70.3} & \textbf{3,548} \\
    \rowcolor{aliceblue!200} FairMOT(\textbf{+Ours}) & 3/4 MOT20 & 6K & 1.8K & 0 & 0 & 20 & \textbf{64.6} & \textbf{67.3} & \textbf{4,225} \\
    \rowcolor{aliceblue!200} FairMOT(\textbf{+Ours}) & 3/4 MOT20 & 6K & 1.8K & 0 & 0 & 30 & \textbf{64.3} & \textbf{67.3} & \textbf{3,801} \\
    \rowcolor{aliceblue!200} FairMOT(\textbf{+Ours}) & 1/2 MOT20 & 4K & 1.4K & 0 & 0 & 20 & \textbf{62.7} & 64.6 & \textbf{3,766} \\
    \rowcolor{aliceblue!200} FairMOT(\textbf{+Ours}) & 1/2 MOT20 & 4K & 1.4K & 0 & 0 & 30 & \textbf{62.9} & 65.8 & \textbf{3,849} \\
   \bottomrule
  \end{tabular}}
 \end{center}
 \caption{Performance under the private detection on the MOT20 test set when using different methods and different datasets for training. ``MIX'' represents the mixed datasets, including MOT20~\cite{dendorfer2020mot20}, ETH~\cite{ess2008mobile}, CityPerson~\cite{zhang2017citypersons}, CalTech~\cite{dollar2009pedestrian}, CUHK-SYSU~\cite{xiao2017joint}, PRW~\cite{zheng2017person} and CrowdHuman~\cite{shao2018crowdhuman} dataset.``CH'' represents CrowdHuman dataset. ``ECP'' represents ETH, CityPerson, CalTech, CUHK-SYSU and PRW datasets. Our method is highlighted in \sethlcolor{aliceblue!200}\hl{blue}. The better results are marked in \textbf{bold}. The best results are marked in \textcolor{red}{red}. All results are obtained from the official MOT challenge evaluation serve, except the result marked with ``*'' is obtained from the FairMOT~\cite{zhang2021fairmot} paper.}
 \label{tab:different methods and less data}
 \vspace{-4mm}
\end{table*}

\begin{figure}
 \setlength{\abovecaptionskip}{0.1cm}  
 \setlength{\belowcaptionskip}{-0.4cm}  
  \centering
   \includegraphics[width=1.0\linewidth]{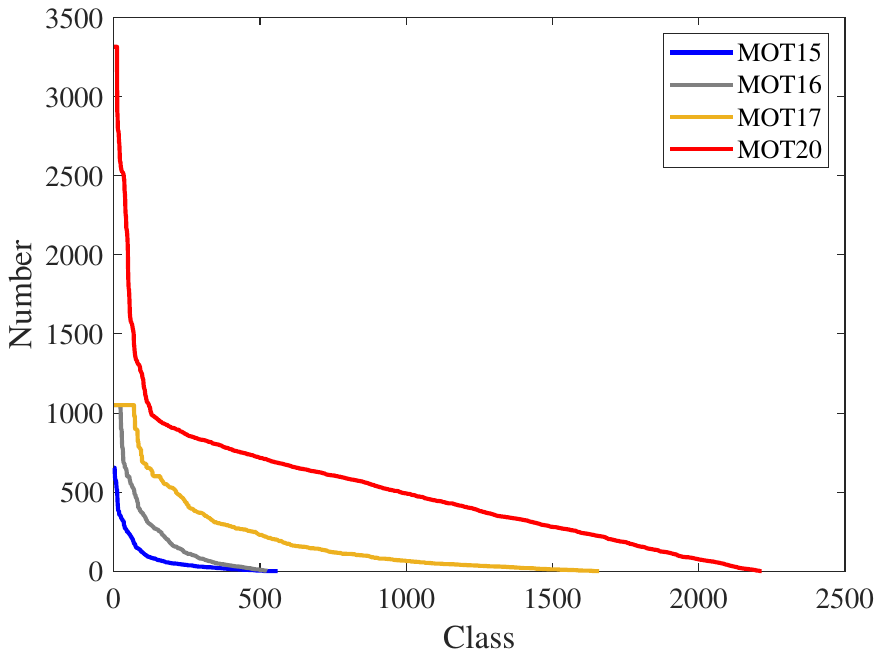}
   \caption{
   The number of pedestrian classes in the MOT15, MOT16, MOT17 and MOT20. We stipulate that the number is the number of frames in which the pedestrian appears, and the pedestrian classes represent pedestrians with different identities.
   }
   \label{fig:The number of pedestrian classes 15 16 17 20.}
\end{figure}

\subsection{Analysis of Performance Differences}
\label{sec:Analysis of Performance Differences}

\quad
In order to explain the differences in the performance improvement of our method on the MOT15, MOT16, MOT17 and MOT20 datasets, we analyze the reasons from the perspective of data.

From the overall data analysis, we plot the number of pedestrian classes of MOT15, MOT16, MOT17 and MOT20 into the same image, which is illustrated in \cref{fig:The number of pedestrian classes 15 16 17 20.}. It can be observed that the data volume of MOT20 changes the most drastically. The number of a few categories reaches more than 3,000, while 90\% of the categories do not reach 1,000, which has serious long-tail distribution characteristics. Like MOT20, MOT15 also has the similar sharp decline in the number of classes. Compared with MOT15 and MOT20, the number of classes in MOT16 and MOT17 dropped more slowly.

\begin{figure*}[ht]
\begin{subfigure}{.19\linewidth}
  \centering
  \includegraphics[width=1.0\linewidth]{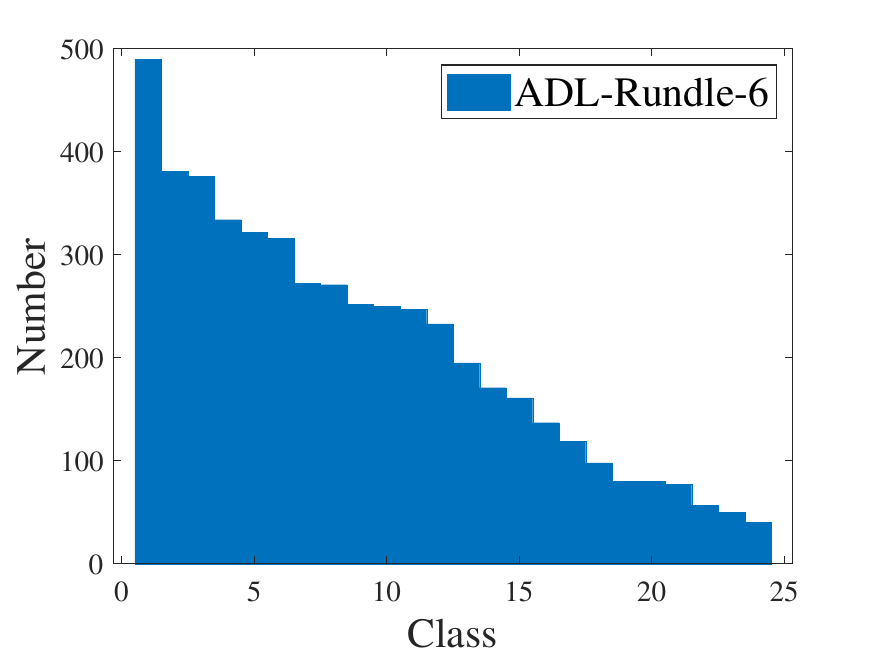}
  \caption{ADL-Rundle-6}
  \label{fig:ADL-Rundle-6}
\end{subfigure}
\hfil
\begin{subfigure}{.19\linewidth}
  \centering
  \includegraphics[width=1.0\linewidth]{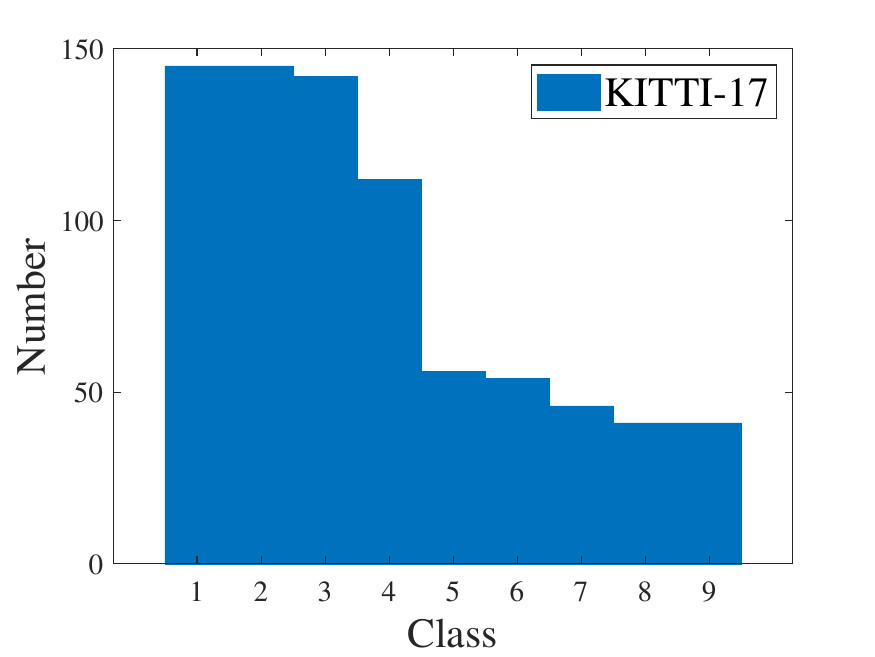}
  \caption{KITTI-17}
  \label{fig:KITTI-17}
\end{subfigure}
\hfil
\begin{subfigure}{.19\linewidth}
  \centering
  \includegraphics[width=1.0\linewidth]{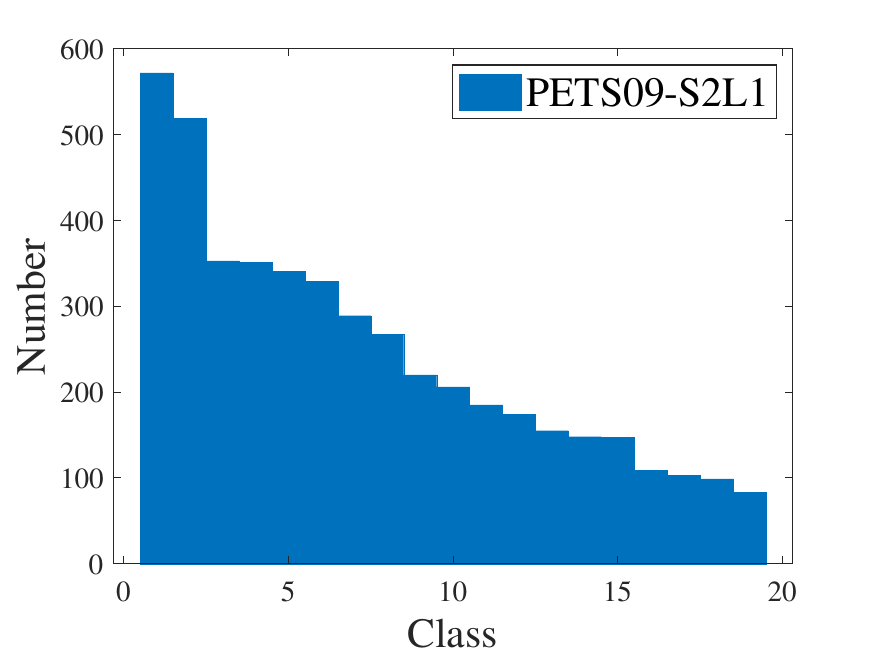}
  \caption{PETS09-S2L1}
  \label{fig:PETS09-S2L1}
\end{subfigure}
\hfil
\begin{subfigure}{.19\linewidth}
  \centering
  \includegraphics[width=1.0\linewidth]{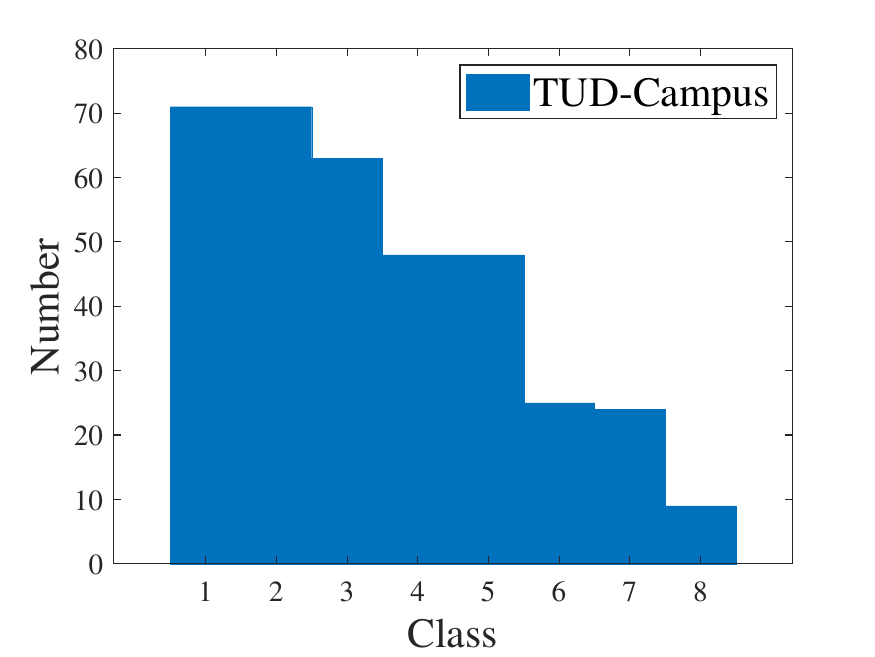}
  \caption{TUD-Campus}
  \label{fig:TUD-Campus}
\end{subfigure}
\hfil
\begin{subfigure}{.19\linewidth}
  \centering
  \includegraphics[width=1.0\linewidth]{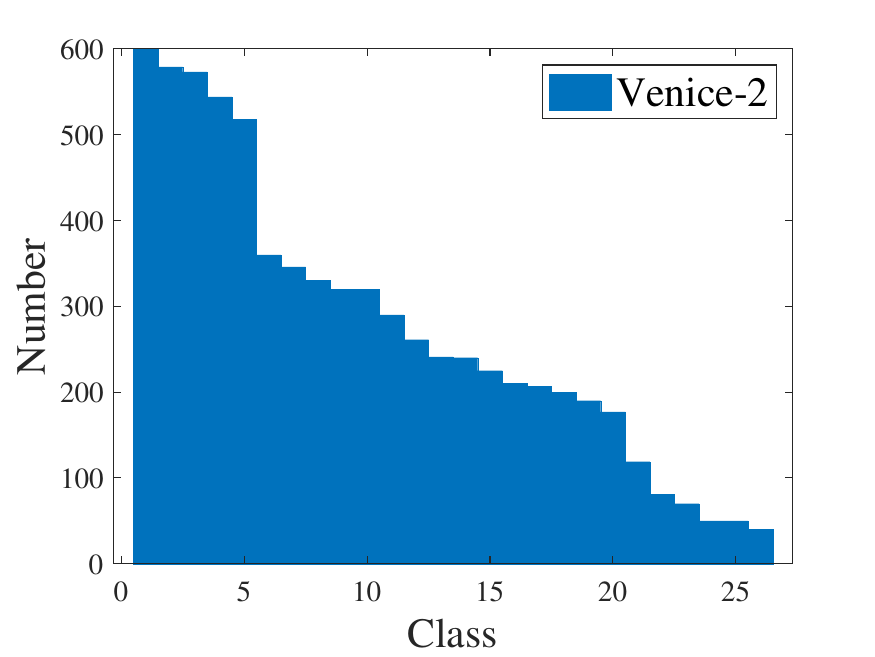}
  \caption{Venice-2}
  \label{fig:Venice-2}
\end{subfigure}
\caption{The number of pedestrian classes for stationary camera view sequences in the MOT15.}
\label{fig:The number of pedestrian classes for stationary camera view sequences in the MOT15.}
\end{figure*}

\begin{figure*}[ht]
\begin{subfigure}{.33\linewidth}
  \centering
  \includegraphics[width=1.0\linewidth]{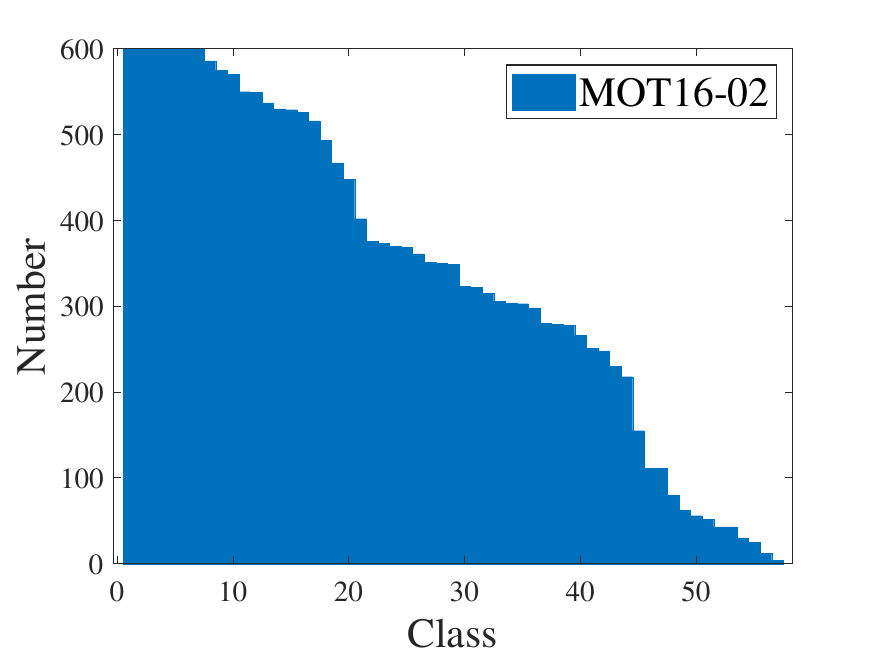}
  \caption{MOT16-02}
  \label{fig:MOT16-02}
\end{subfigure}
\hfil
\begin{subfigure}{.33\linewidth}
  \centering
  \includegraphics[width=1.0\linewidth]{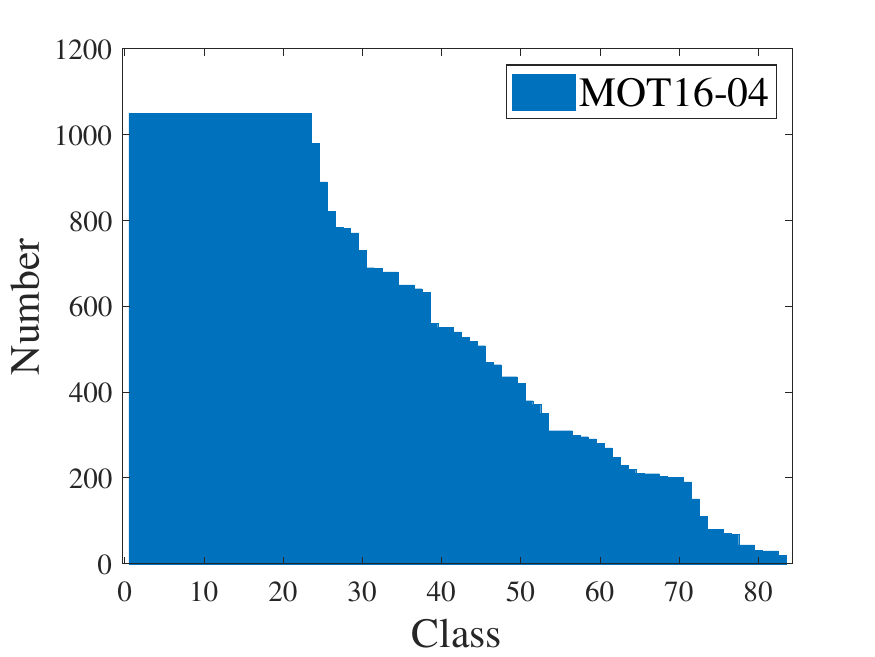}
  \caption{MOT16-04}
  \label{fig:MOT16-04}
\end{subfigure}
\hfil
\begin{subfigure}{.33\linewidth}
  \centering
  \includegraphics[width=1.0\linewidth]{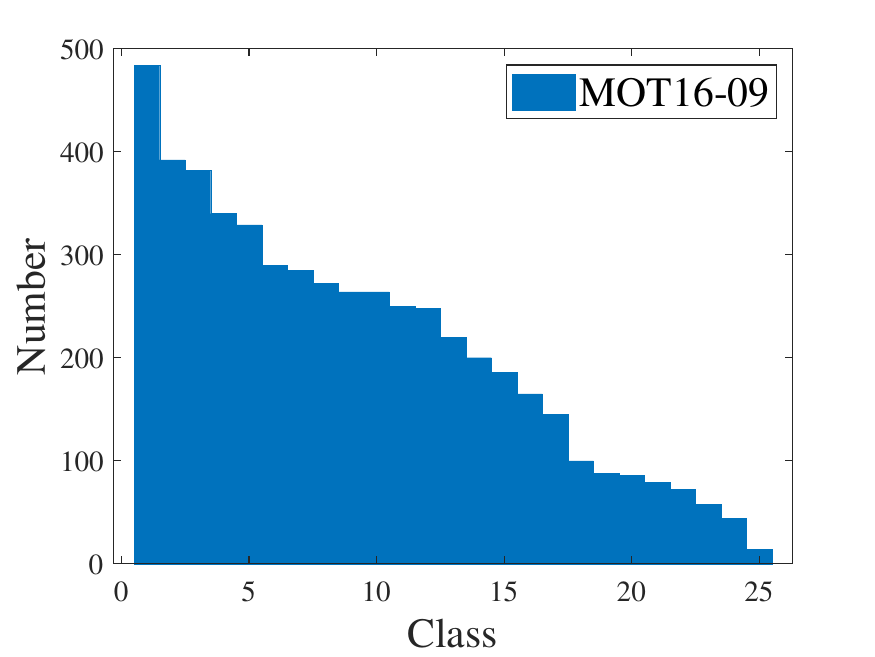}
  \caption{MOT16-09}
  \label{fig:MOT16-09}
\end{subfigure}
\caption{The number of pedestrian classes for stationary camera view sequences in the MOT16.}
\label{fig:The number of pedestrian classes for stationary camera view sequences in the MOT16.}
\end{figure*}

\begin{figure*}[ht]
\begin{subfigure}{.24\linewidth}
  \centering
  \includegraphics[width=1.0\linewidth]{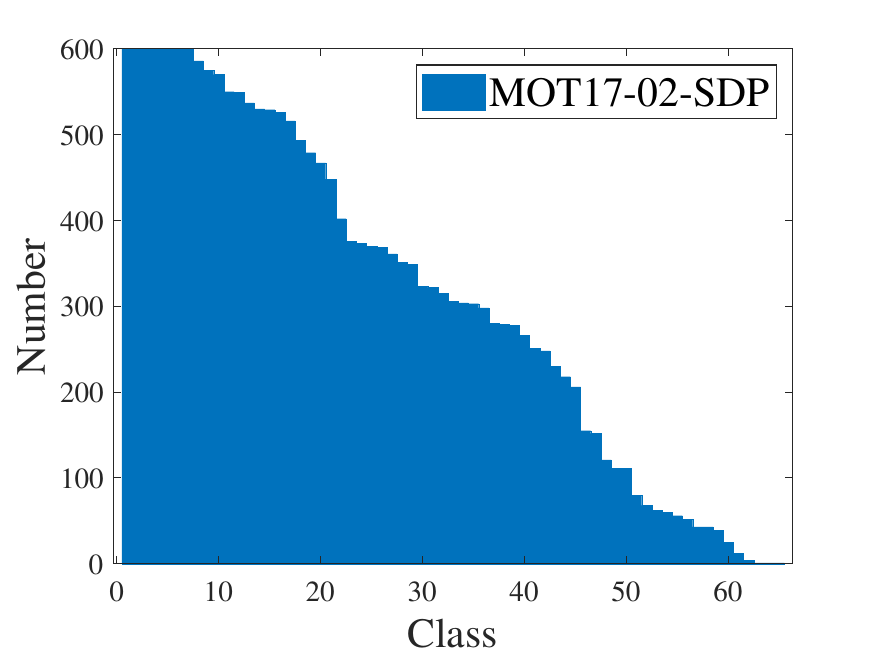}
  \caption{MOT17-02-SDP}
  \label{fig:MOT17-02-SDP}
\end{subfigure}
\hfil
\begin{subfigure}{.24\linewidth}
  \centering
  \includegraphics[width=1.0\linewidth]{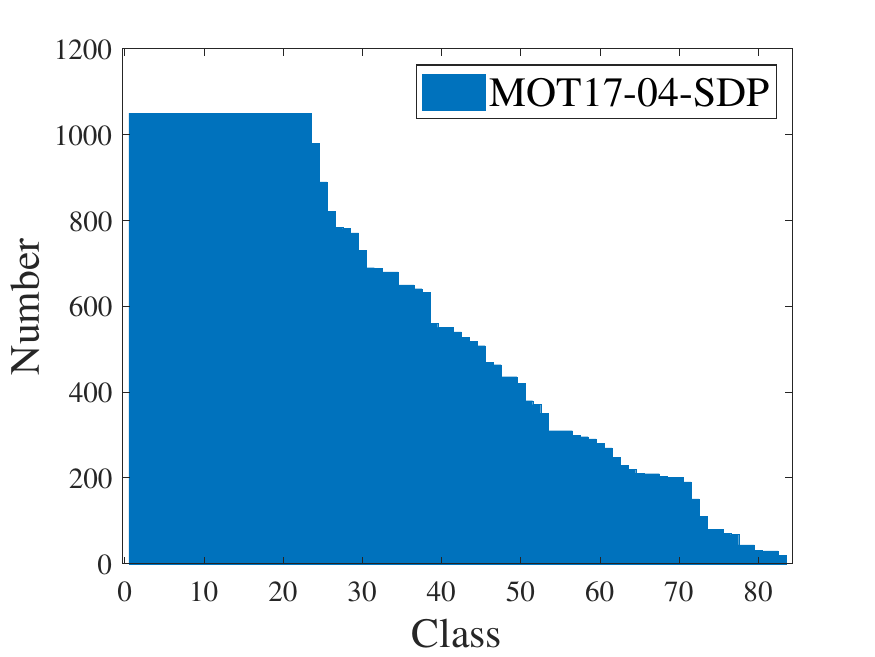}
  \caption{MOT17-04-SDP}
  \label{fig:MOT17-04-SDP}
\end{subfigure}
\hfil
\begin{subfigure}{.24\linewidth}
  \centering
  \includegraphics[width=1.0\linewidth]{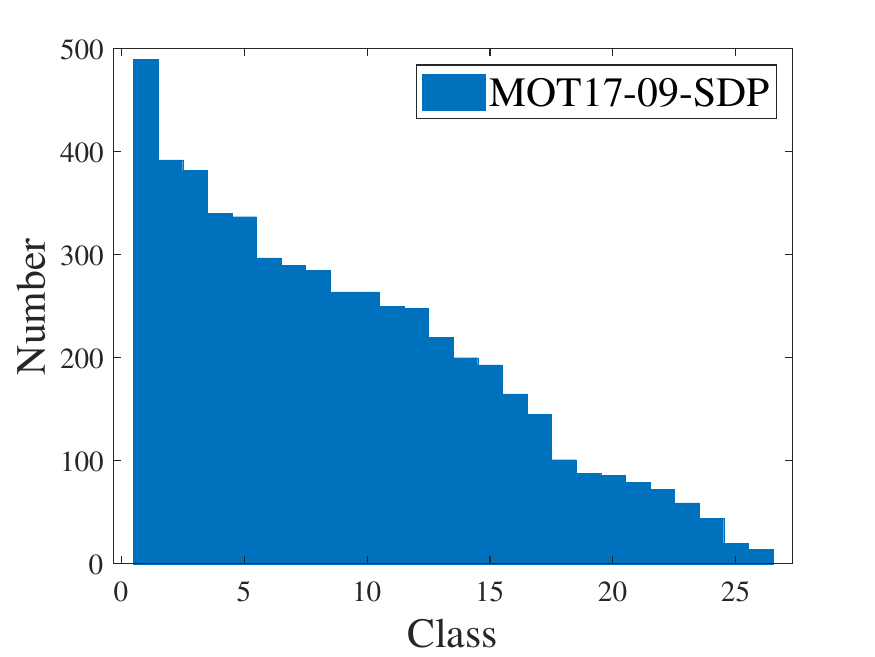}
  \caption{MOT17-09-SDP}
  \label{fig:MOT17-09-SDP}
\end{subfigure}
\hfil
\begin{subfigure}{.24\linewidth}
  \centering
  \includegraphics[width=1.0\linewidth]{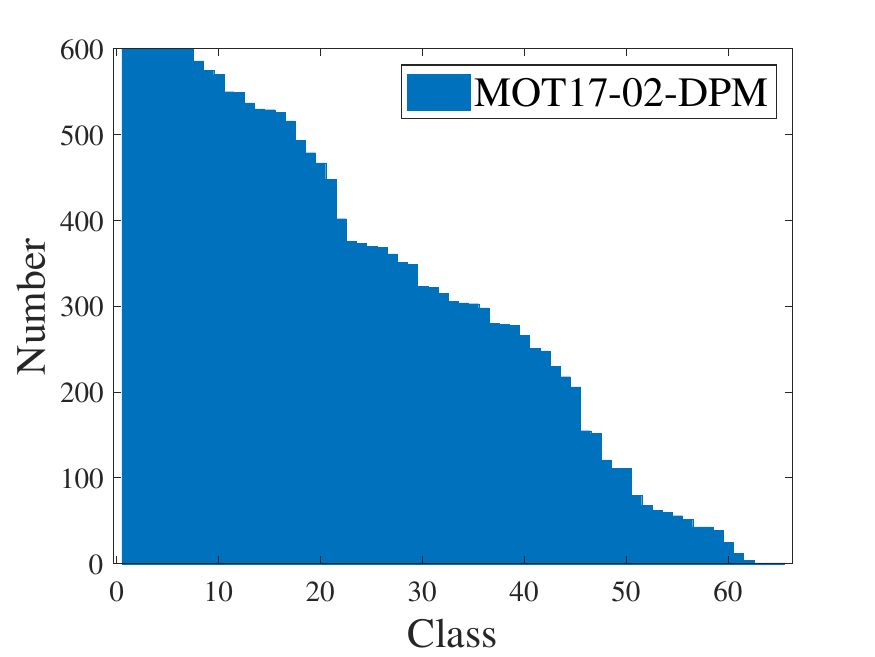}
  \caption{MOT17-02-DPM}
  \label{fig:MOT17-02-DPM}
\end{subfigure}
\newline
\begin{subfigure}{.19\linewidth}
  \centering
  \includegraphics[width=1.0\linewidth]{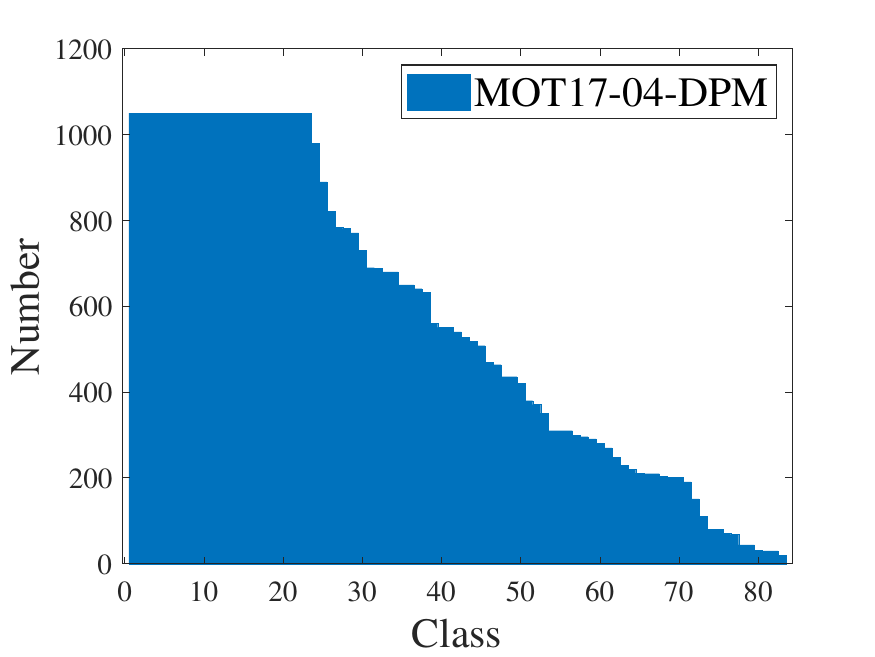}
  \caption{MOT17-04-DPM}
  \label{fig:MOT17-04-DPM}
\end{subfigure}
\hfil
\begin{subfigure}{.19\linewidth}
  \centering
  \includegraphics[width=1.0\linewidth]{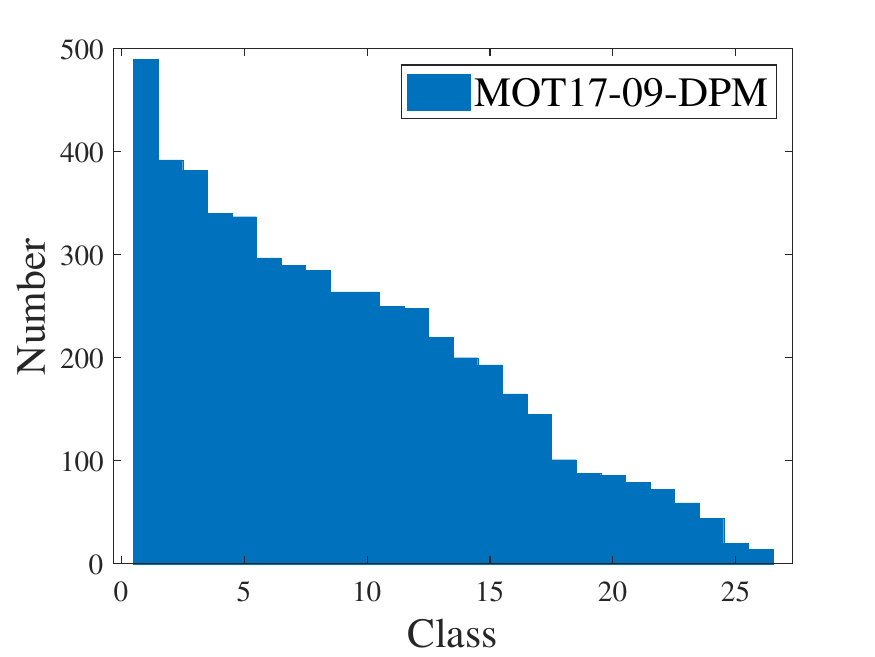}
  \caption{MOT17-09-DPM}
  \label{fig:MOT17-09-DPM}
\end{subfigure}
\hfil
\begin{subfigure}{.19\linewidth}
  \centering
  \includegraphics[width=1.0\linewidth]{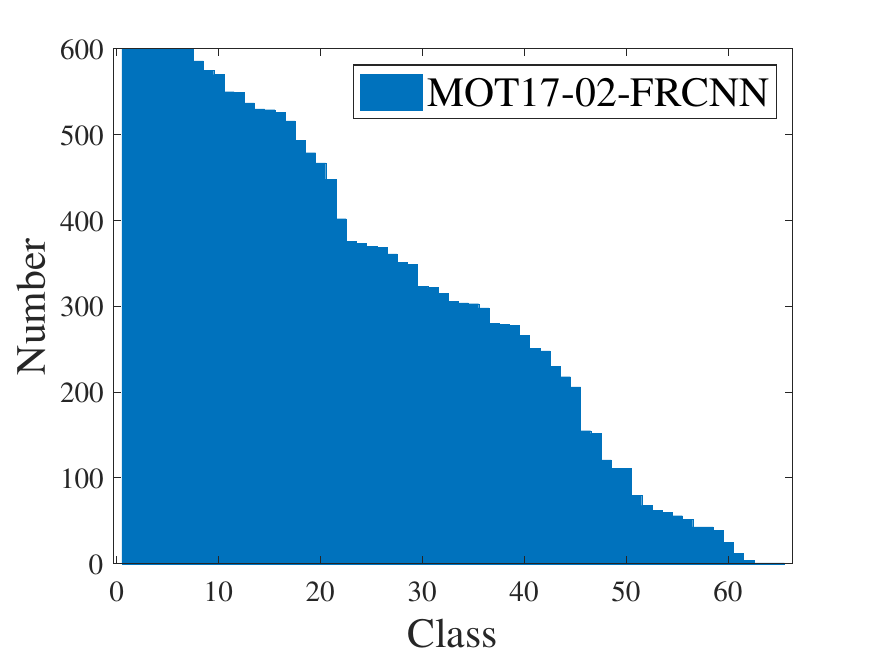}
  \caption{MOT17-02-FRCNN}
  \label{fig:MOT17-02-FRCNN}
\end{subfigure}
\hfil
\begin{subfigure}{.19\linewidth}
  \centering
  \includegraphics[width=1.0\linewidth]{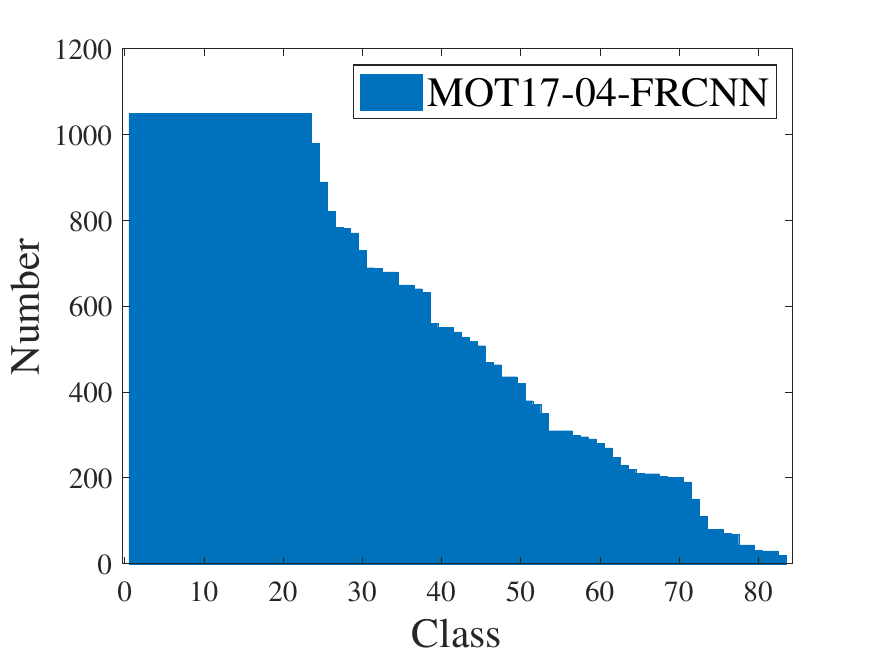}
  \caption{MOT17-04-FRCNN}
  \label{fig:MOT17-04-FRCNN}
\end{subfigure}
\hfil
\begin{subfigure}{.19\linewidth}
  \centering
  \includegraphics[width=1.0\linewidth]{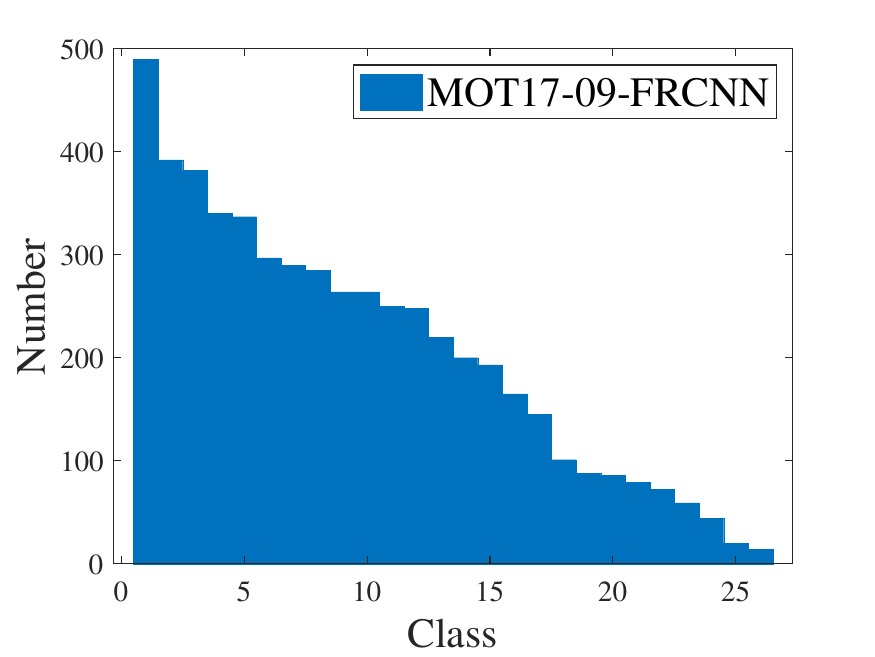}
  \caption{MOT17-09-FRCNN}
  \label{fig:MOT17-09-FRCNN}
\end{subfigure}
\caption{The number of pedestrian classes for stationary camera view sequences in the MOT17.}
\label{fig:The number of pedestrian classes for stationary camera view sequences in the MOT17.}
\end{figure*}

\begin{figure*}[ht]
\begin{subfigure}{.24\linewidth}
  \centering
  \includegraphics[width=1.0\linewidth]{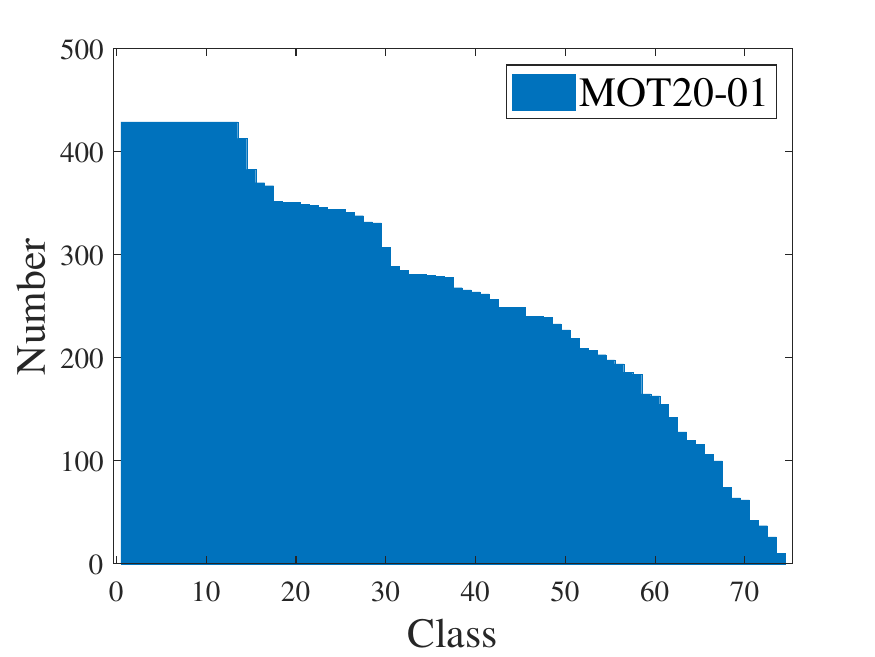}
  \caption{MOT20-01}
  \label{fig:MOT20-01}
\end{subfigure}
\hfil
\begin{subfigure}{.24\linewidth}
  \centering
  \includegraphics[width=1.0\linewidth]{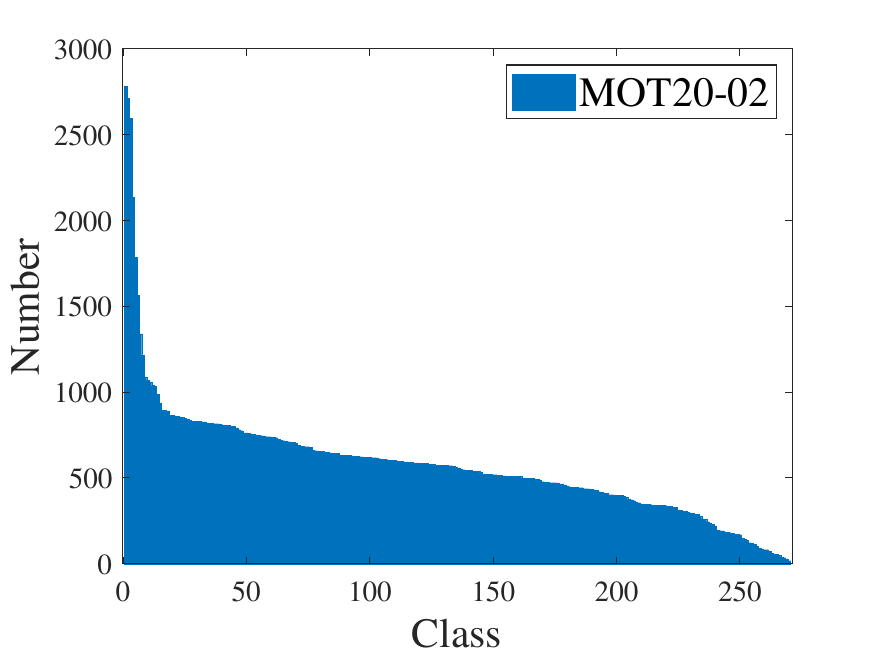}
  \caption{MOT20-02}
  \label{fig:MOT20-02}
\end{subfigure}
\hfil
\begin{subfigure}{.24\linewidth}
  \centering
  \includegraphics[width=1.0\linewidth]{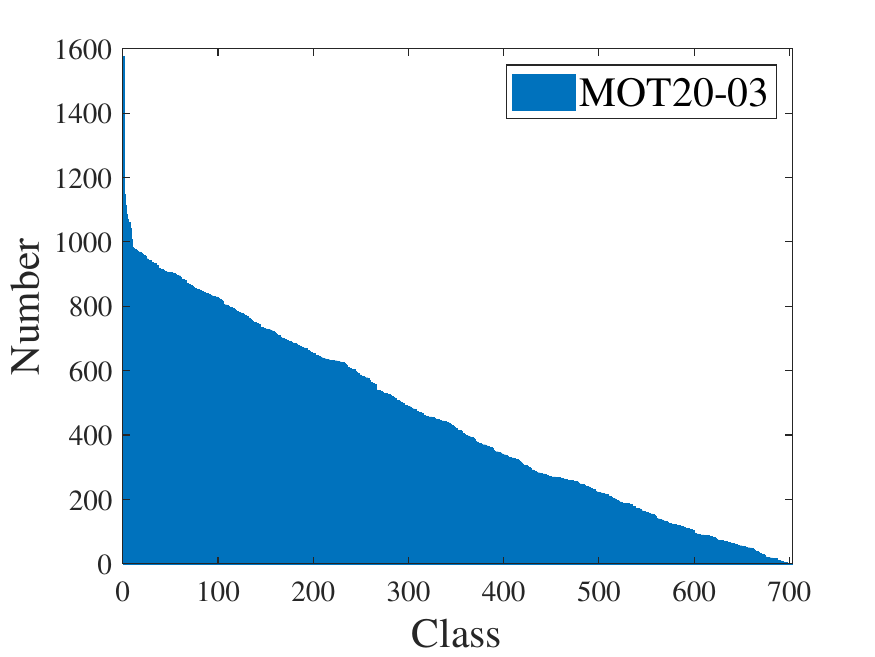}
  \caption{MOT20-03}
  \label{fig:MOT20-03}
\end{subfigure}
\hfil
\begin{subfigure}{.24\linewidth}
  \centering
  \includegraphics[width=1.0\linewidth]{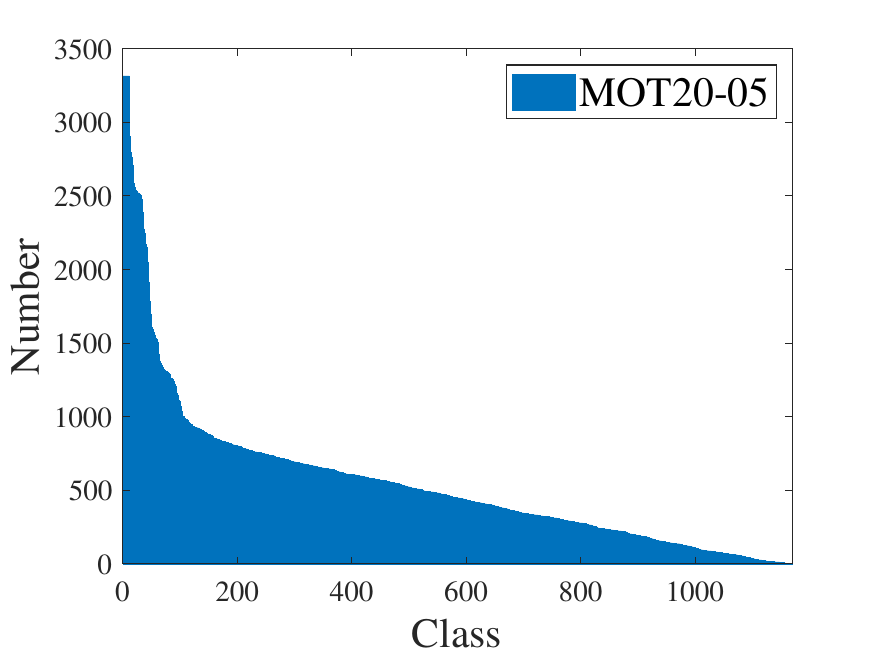}
  \caption{MOT20-05}
  \label{fig:MOT20-05}
\end{subfigure}
\caption{The number of pedestrian classes for stationary camera view sequences in the MOT20.}
\label{fig:The number of pedestrian classes for stationary camera view sequences in the MOT20.}
\end{figure*}

From the analysis of stationary camera view data, we plot the number of pedestrian classes belonging to all stationary camera view sequences of MOT15, MOT16, MOT17 and MOT20. As shown in \cref{fig:The number of pedestrian classes for stationary camera view sequences in the MOT15.}, we can observe that for the KITTI-17, PETS09-S2L11, and Venice-2 sequences in MOT15, the number of pedestrian classes will decrease sharply in a certain interval, and an obvious boundary will appear. As shown in \cref{fig:The number of pedestrian classes for stationary camera view sequences in the MOT16.} and \cref{fig:The number of pedestrian classes for stationary camera view sequences in the MOT17.}, we can observe that there are no obvious boundary in all sequences of MOT16 and MOT17, and the changes in the number of pedestrian classes are relatively gentle. As shown in \cref{fig:The number of pedestrian classes for stationary camera view sequences in the MOT20.}, we can observe that in the MOT20-02 and MOT20-05 sequences in MOT20, the number of pedestrian classes will drop extremely sharply in a certain interval, and there will be an extremely serious demarcation, which has serious long-tail distribution characteristics.

Based on the above analysis results, we can conclude that because the MOT20 dataset has the most long-tail distribution characteristics, it can explain that our method achieves the best indicator on the MOT20 dataset.

\subsection{Additional Ablation Study}
\label{sec:Additional Ablation Study}
\quad

In this subsection, we present more additional ablation study results to further analyze our method. 

\noindent \textbf{Analysis of visibility thresholds $T_v$ in SVA.}
As introduced in \cref{sec:SVA}. To demonstrate the effectiveness of $T_v$ in the proposed target trajectory prediction continuation, we compare it with different visibility thresholds from 0.1 to 1.0 at intervals of 0.1. The same parameter settings are used for a fair comparison. As shown in \cref{tab:visibility thresholds}, $T_v$ takes 0.9 to achieve the best results. We speculate that appropriate adoption of fully visible pedestrian can improve network learning.

\begin{table}[t]
 \setlength{\abovecaptionskip}{-0.2cm} 
 \setlength{\belowcaptionskip}{0.2cm} 
 \footnotesize
 \begin{center}
  \setlength{\tabcolsep}{5.0pt}
  \renewcommand{\arraystretch}{1.0}
  {
  \begin{tabular}{cccccc}
   \toprule
    $T_v$ & MOTA↑ & IDF1↑ & FP↓ & FN↓ & IDS↓  \\
   \midrule
    0.1 & 68.6 & 72.5 & 2,809 & 13,702 & 437 \\
    0.2 & \textcolor{blue}{69.2} & 71.3 & 2,732 & \textcolor{blue}{13,496} & 442 \\
    0.3 & 69.0 & 69.5 & 2,792 & 13,528 & 422 \\
    0.4 & 68.9 & 73.0 & 2,660 & 13,728 & \textcolor{red}{403} \\
    0.5 & 68.9 & 70.9 & 2,912 & \textcolor{red}{13,429} & 466 \\
    0.6 & 68.7 & 70.8 & \textcolor{blue}{2,615} & 13,815 & 469 \\
    0.7 & \textcolor{blue}{69.2} & 72.0 & \textcolor{red}{2,587} & 13,664 & \textcolor{blue}{411} \\
    0.8 & 68.8 & \textcolor{blue}{73.2} & 2,676 & 13,740 & 446 \\
    1.0 & 69.0 & 72.7 & 2,670 & 13,689 & 420 \\
    \midrule
    0.9 & \textcolor{red}{69.3} & \textcolor{red}{73.4} & 2,640 & 13,504 & 455 \\
   \bottomrule
  \end{tabular}}
 \end{center}
 \caption{Comparison of different visibility thresholds $T_v$. The top two results are highlighted with \textcolor{red}{red} and \textcolor{blue}{blue}.}
 \label{tab:visibility thresholds}
 \vspace{-5mm}
\end{table}

\begin{table}[t]
 \footnotesize
\setlength{\abovecaptionskip}{-0.2cm}
\setlength{\belowcaptionskip}{0.2cm}
 \begin{center}
  \setlength{\tabcolsep}{3.0pt}{
  \renewcommand{\arraystretch}{1.06}
  \begin{tabular}{ccccc}
   \toprule
    Enhancement Coefficient    & MOTA↑ & IDF1↑ & IDS↓ \\
   \midrule
    0.1 & 68.9 & 71.0 & 508 \\
    0.2 & 68.7 & 71.0 & \textcolor{blue}{442} \\
    0.3 & \textcolor{red}{69.3} & \textcolor{blue}{72.0} & 463 \\
    0.5 & 68.7 & 70.7 & 465 \\
    0.6 & 68.8 & 71.1 & 447 \\
    0.7 & 68.9 & 71.9 & 488 \\
    0.8 & 69.0 & 71.0 & 457 \\
    0.9 & \textcolor{blue}{69.2} & 70.6 & \textcolor{red}{419} \\
   \midrule
    0.4 & \textcolor{red}{69.3} & \textcolor{red}{73.4} & 455 \\
   \bottomrule
  \end{tabular}}
 \end{center}
 \caption{Comparison of different enhancement coefficients. The top two results are highlighted with \textcolor{red}{red} and \textcolor{blue}{blue}.}
 \label{tab:enhancement coefficient}
 \vspace{-6mm}
\end{table}

\noindent \textbf{Analysis of enhancement coefficient in DVA.} We analyze how different enhancement coefficients of diffusion to affect the performance of the network. We change the enhancement coefficient from 0.1 to 0.9 at intervals of 0.1. As shown in \cref{tab:enhancement coefficient}, the best experimental results occur when the enhancement coefficient is 0.4. We can observe that the effect of large enhancement coefficients was generally worse than that of small enhancement coefficients. 

\noindent \textbf{Analysis of prompt in DVA.} As introduced in \cref{sec:DVA}, we select the prompt of diffusion to analyze the impact on the results. For the MOT17 training set, sequences of 05, 10 and 13 belong to the street scene, and 11 sequence belongs to the mall scene. Thus, we set up five types of prompts for the 05, 10 and 13 sequences, including ``A street'', ``A empty street'', ``A crowded street'', ``A street with no pedestrians'' and `` ''. We also set up five types of prompts for the 11 sequence, including ``A mall'', ``A empty mall'', ``A crowded mall'', ``A mall with no pedestrians'' and `` ''.
As shown in \cref{tab:prompt}, ``A street'' for 05, 10 and 13, ``A mall'' for 11 leads to the best results. We speculate that the simple prompt is powerful for diffusion. Just the simple prompt is all you need.

\begin{table}[t]
 \footnotesize
\setlength{\abovecaptionskip}{-0.2cm} 
\setlength{\belowcaptionskip}{+0.2cm} 
 \begin{center}
  \setlength{\tabcolsep}{2.4pt}{ 
  \renewcommand{\arraystretch}{1.0} 
  \begin{tabular}{cccccc}
   \toprule
    Prompt & MOTA↑ & IDF1↑ & FP↓ & FN↓ & IDS↓ \\
    \midrule
    $S_{2}$ for 05, 10 and 13, $M_{2}$ for 11 & \textcolor{red}{69.4} & 71.2 & \textcolor{red}{2,594} & 13,512 & \textcolor{red}{436} \\
    $S_{3}$ for 05, 10 and 13, $M_{3}$ for 11 & 69.2 & 71.4 & 2,713 & 13,454 & 464 \\
    $S_{4}$ for 05, 10 and 13, $M_{4}$ for 11 & 69.2 & \textcolor{blue}{73.0} & 2,762 & \textcolor{red}{13,449} & 456 \\
    $S_{5}$ for 05, 10 and 13, $M_{5}$ for 11 & 68.7 & 71.6 & 2,953 & 13,506 & 483 \\
    \midrule
    $S_{1}$ for 05, 10 and 13, $M_{1}$ for 11 & \textcolor{blue}{69.3} & \textcolor{red}{73.4} & \textcolor{blue}{2,640} & \textcolor{blue}{13,504} & \textcolor{blue}{455} \\
    \bottomrule
  \end{tabular}}
 \end{center}
 \caption{Comparison of different prompts. ($S_{1}$:``A street'', $S_{2}$:``A empty street'', $S_{3}$:``A crowded street'', $S_{4}$:``A street with no pedestrians'', $S_{5}$:`` '', $M_{1}$:``A mall'', $M_{2}$:``A empty mall'', $M_{3}$:``A crowded mall'', $M_{4}$:``A mall with no pedestrians'' and $M_{5}$:`` ''.The top two results are highlighted with \textcolor{red}{red} and \textcolor{blue}{blue}.)}
 \label{tab:prompt}%
 \vspace{-5mm} 
\end{table}

\begin{table}[t]
 \setlength{\abovecaptionskip}{-0.2cm}  
 \setlength{\belowcaptionskip}{+0.2cm}  
 \footnotesize
 \begin{center}
  \setlength{\tabcolsep}{2.0pt}
  \renewcommand{\arraystretch}{1.0} 
  {
  \begin{tabular}{cccccc}
   \toprule
    $K$ & MOTA↑ & IDF1↑ & FP↓ & FN↓ & IDS↓ \\
   \midrule
    2 & 69.0 & \textcolor{blue}{72.0} & 2,651 & 13,680 & \textcolor{blue}{398} \\
    4 & 68.9 & 70.8 & 2,796 & 13,581 & 431 \\
    5 & \textcolor{red}{69.4} & 70.5 & 2,682 & \textcolor{blue}{13,443} & 432 \\
    6 & 69.2 & 71.4 & \textcolor{red}{2,524} & 13,751 & \textcolor{red}{394} \\
    7 & 68.7 & 70.0 & 3,053 & \textcolor{red}{13,383} & 500 \\
    8 & 69.1 & 70.4 & 2,714 & 13,492 & 488 \\
   \midrule
    3 & \textcolor{blue}{69.3} & \textcolor{red}{73.4} & \textcolor{blue}{2,640} & 13,504 & 455 \\
   \bottomrule
  \end{tabular}}
 \end{center}
 \caption{Comparison of different group number $K$. The top two results are highlighted with \textcolor{red}{red} and \textcolor{blue}{blue}.} 
 \label{tab:Group Softmax Module for Re-ID}%
 \vspace{-4mm}
\end{table}

\noindent \textbf{Analysis of group number $K$ in GS.}
As introduced in \cref{sec:GS}. We select the group number $K$ for analysis. To demonstrate the effectiveness of different group number, we compare it with different group number, including 2, 3, 4, 5, 6, 7 and 8. As shown in \cref{tab:Group Softmax Module for Re-ID}, we can observe that the group number $K$ takes 3 to achieve the best results. We speculate that too much group number is not conducive to network learning and the group number $K$ is a sensitive hyper-parameter. 

\subsection{Visualization}
\label{sec:Visualization}

\begin{figure}[ht]
\begin{subfigure}{1.0\linewidth}
  \centering
  \includegraphics[width=0.8\linewidth]{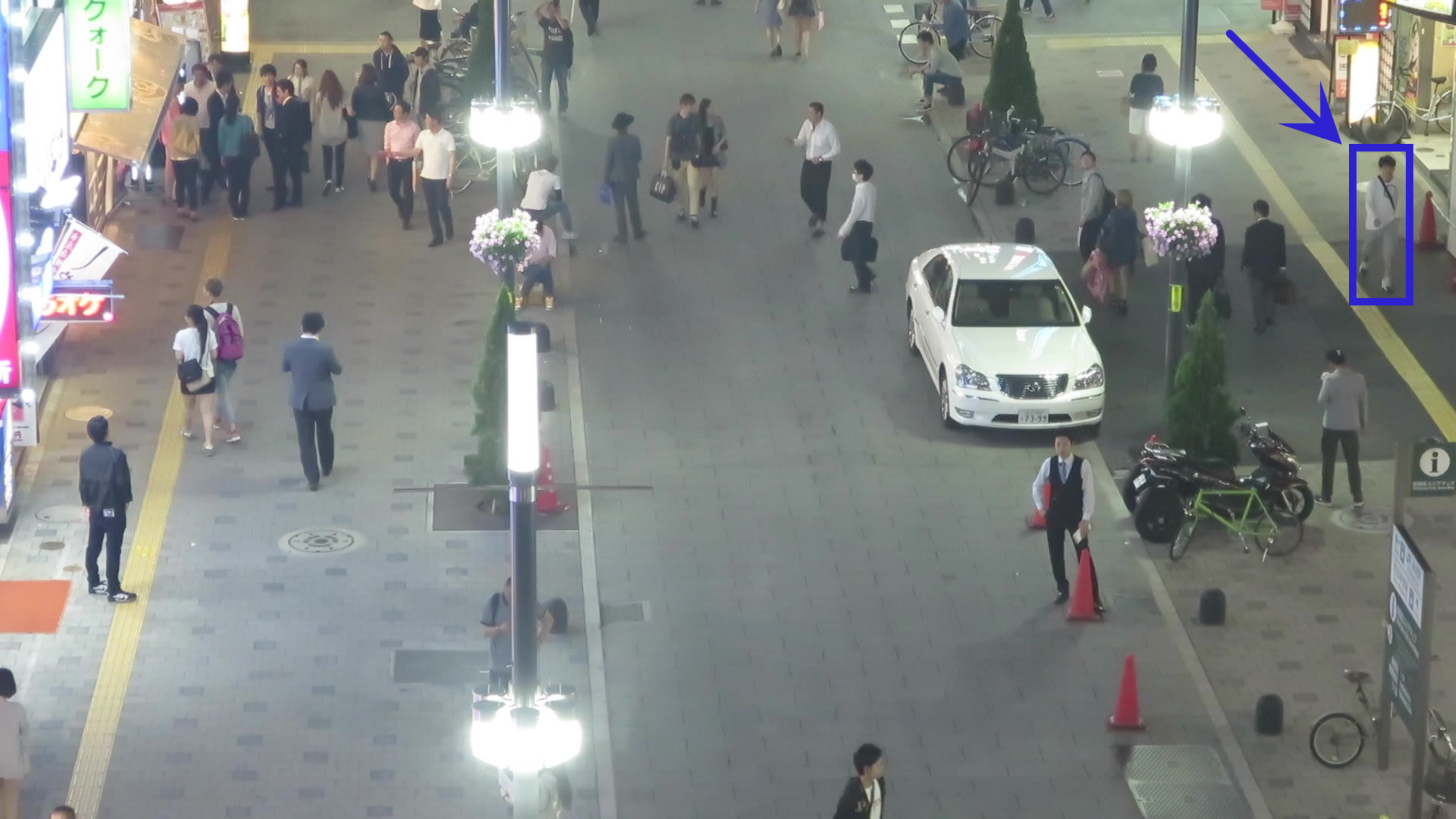}
  \caption{MOT17-04, frame: 000001}
  \label{fig:MOT17-04, frame: 000001}
\end{subfigure}
\newline
\begin{subfigure}{1.0\linewidth}
  \centering
  \includegraphics[width=0.8\linewidth]{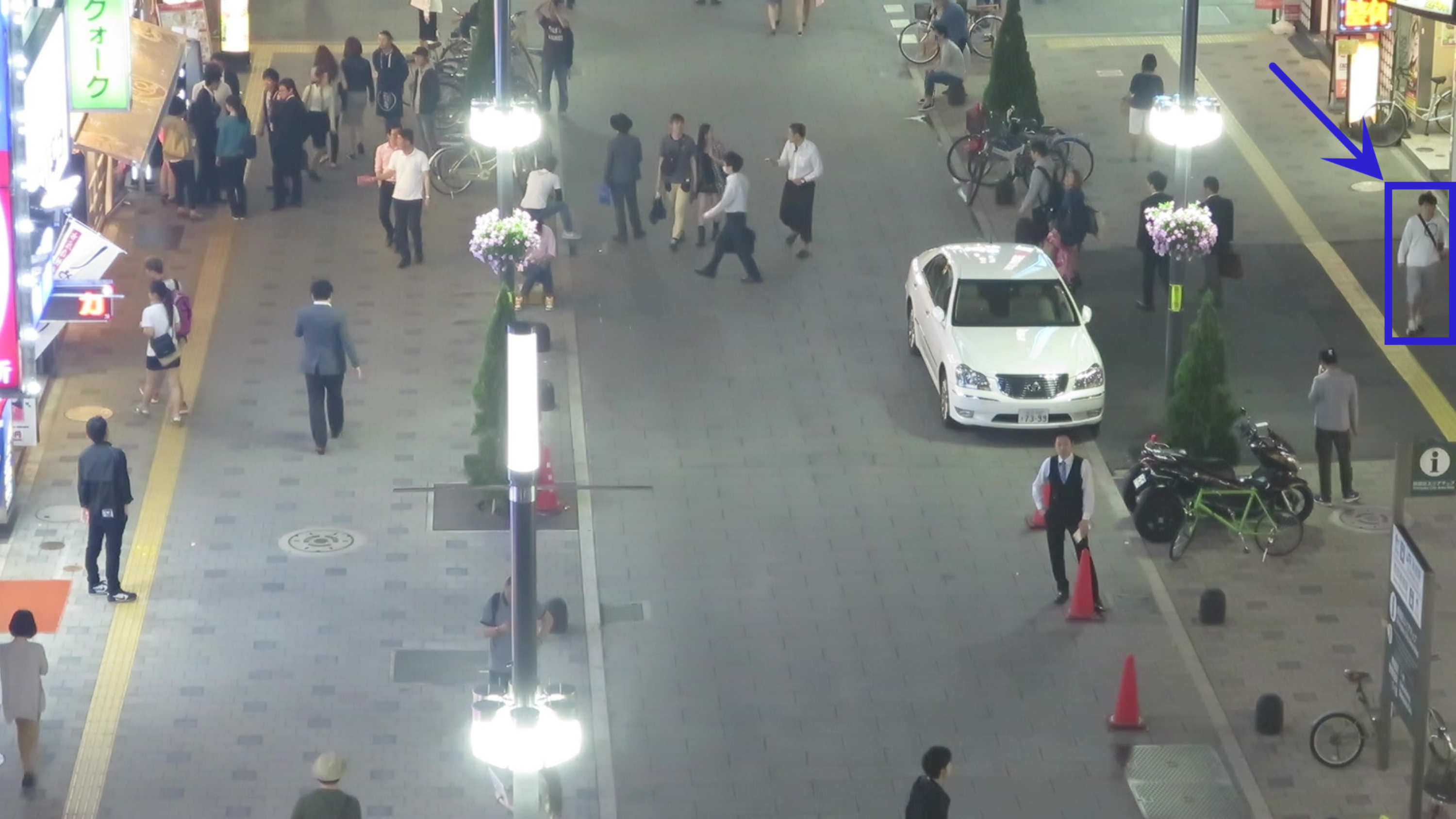}
  \caption{MOT17-04, frame: 000040}
  \label{fig:MOT17-04, frame: 000040}
\end{subfigure}
\newline
\begin{subfigure}{1.0\linewidth}
  \centering
  \includegraphics[width=0.8\linewidth]{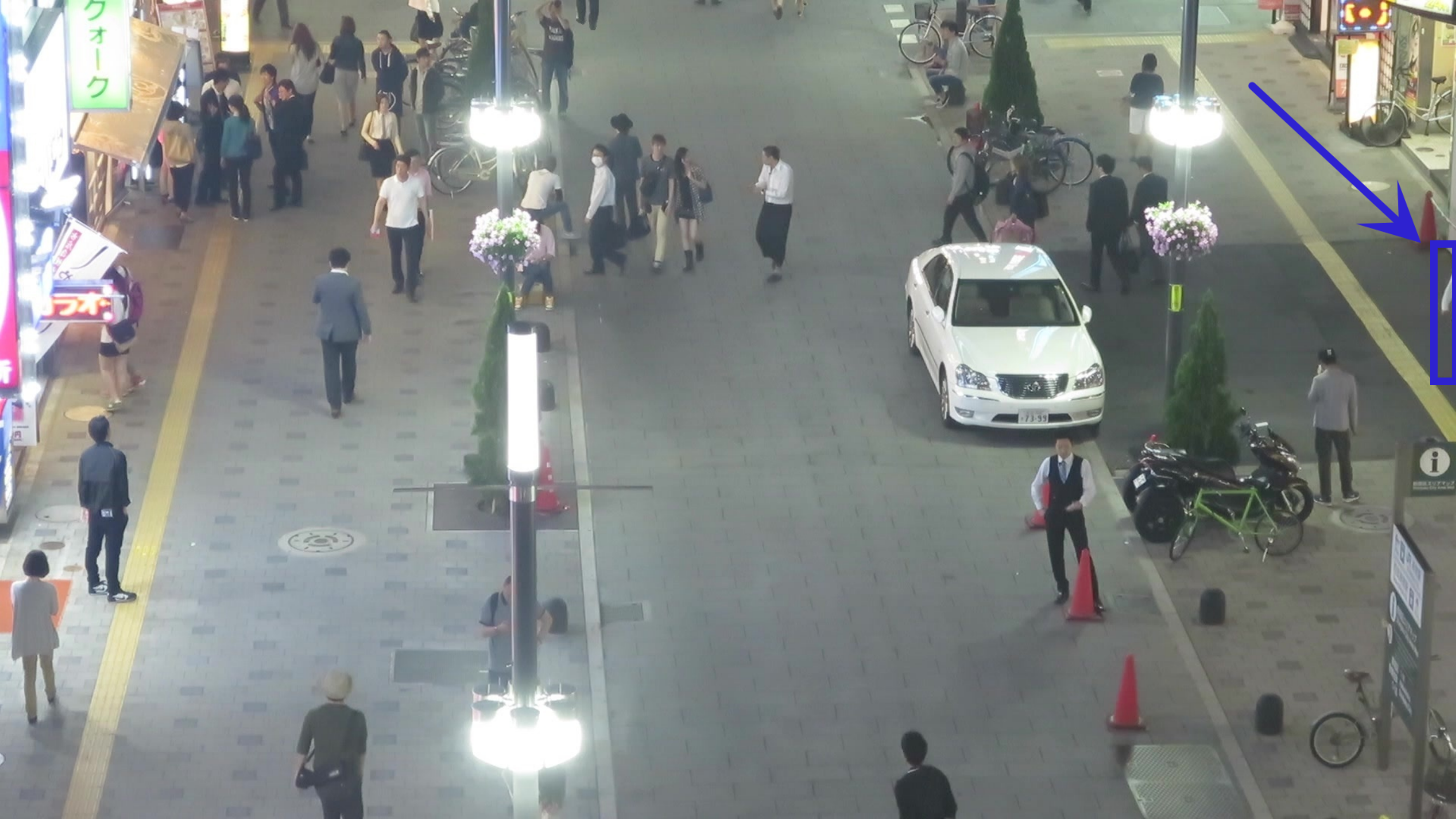}
  \caption{MOT17-04, frame: 000080}
  \label{fig:MOT17-04, frame: 000080}
\end{subfigure}
\newline
\begin{subfigure}{1.0\linewidth}
  \centering
  \includegraphics[width=0.8\linewidth]{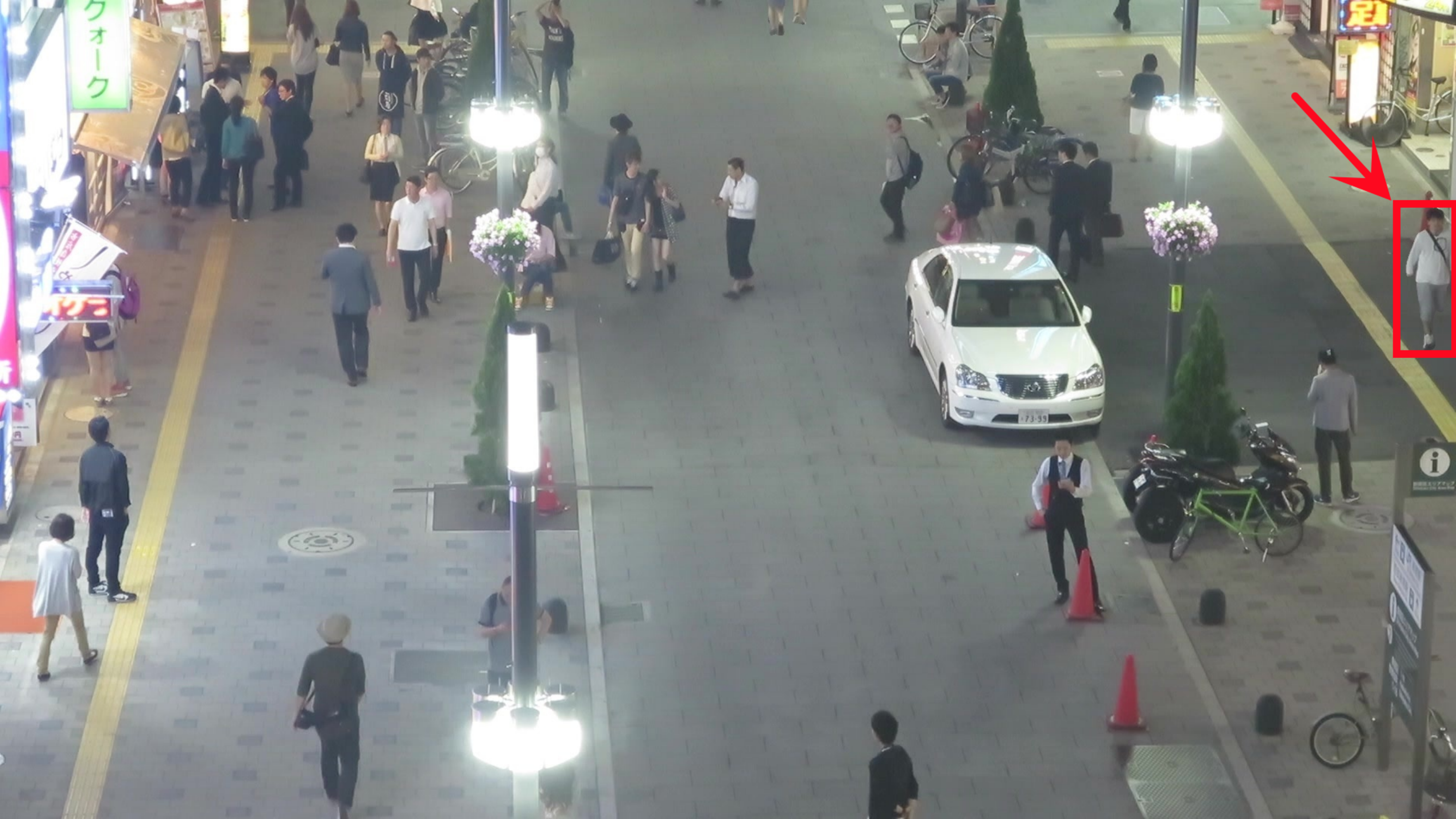}
  \caption{MOT17-04, frame: 000110}
  \label{fig:MOT17-04, frame: 000110}
\end{subfigure}
\newline
\begin{subfigure}{1.0\linewidth}
  \centering
  \includegraphics[width=0.8\linewidth]{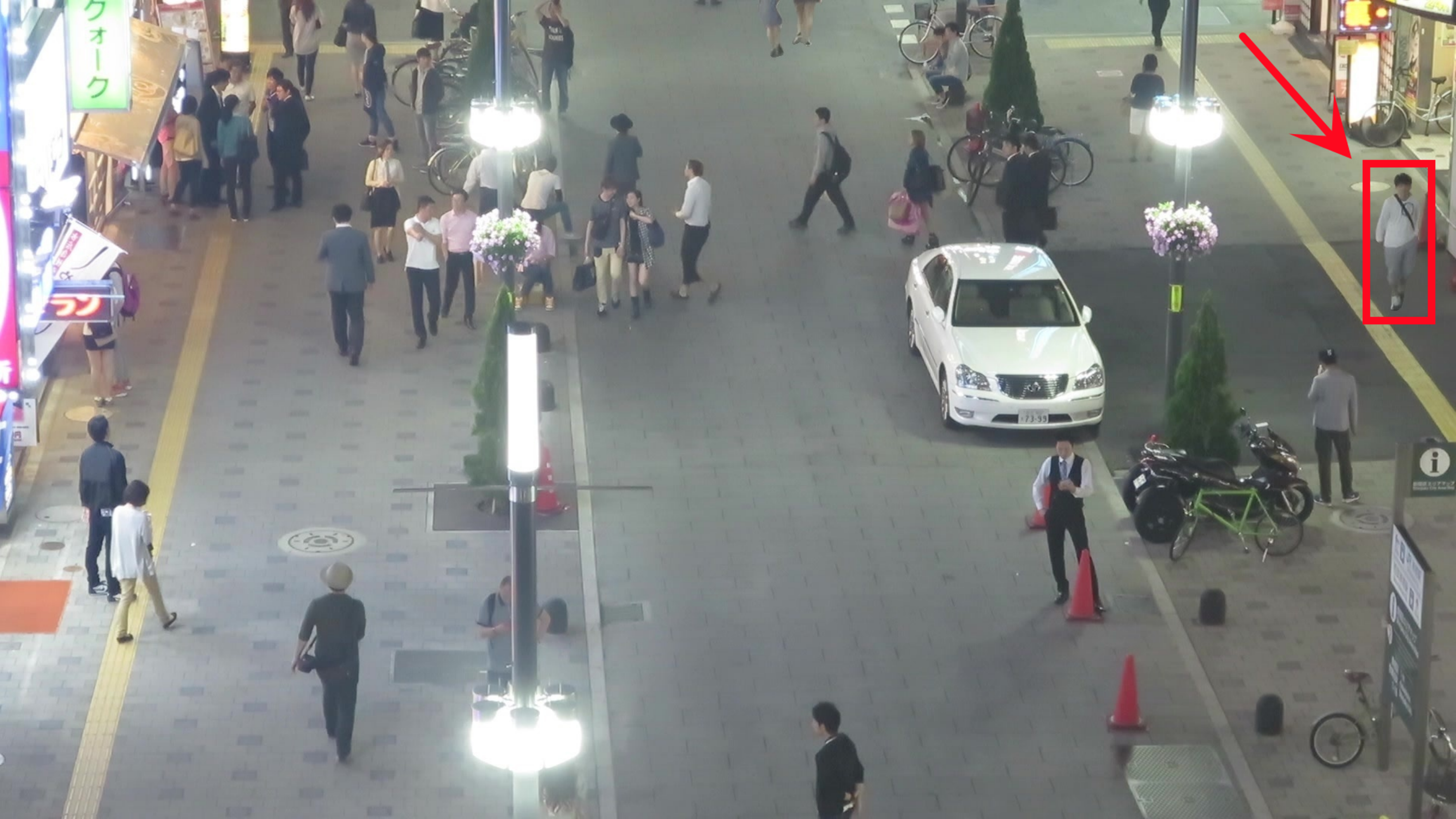}
  \caption{MOT17-04, frame: 000140}
  \label{fig:MOT17-04, frame: 000140}
\end{subfigure}
\caption{Example of backtracking continuation in SVA. The pedestrian in \textcolor{red}{red} box represents the backtracked pedestrian, and the pedestrian in \textcolor{blue}{blue} box represents the original pedestrian.}
\label{fig:Example of backtracking continuation in the Stationary Camera View Data Augmentation (SVA).}
\end{figure}

\begin{figure}[ht]
\begin{subfigure}{1.0\linewidth}
  \centering
  \includegraphics[width=0.8\linewidth]{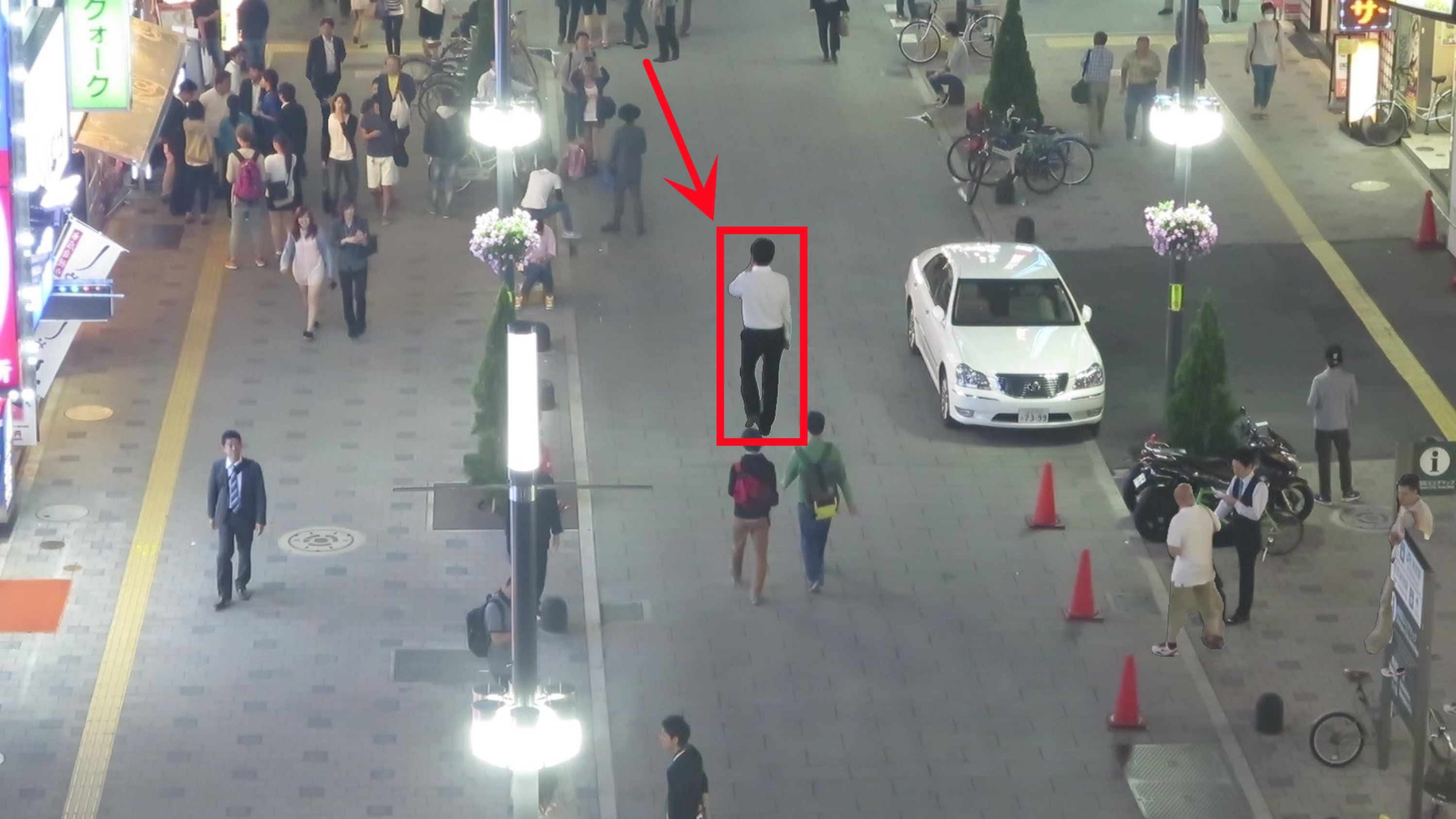}
  \caption{MOT17-04, frame: 000830}
  \label{fig:MOT17-04, frame: 000830}
\end{subfigure}
\newline
\begin{subfigure}{1.0\linewidth}
  \centering
  \includegraphics[width=0.8\linewidth]{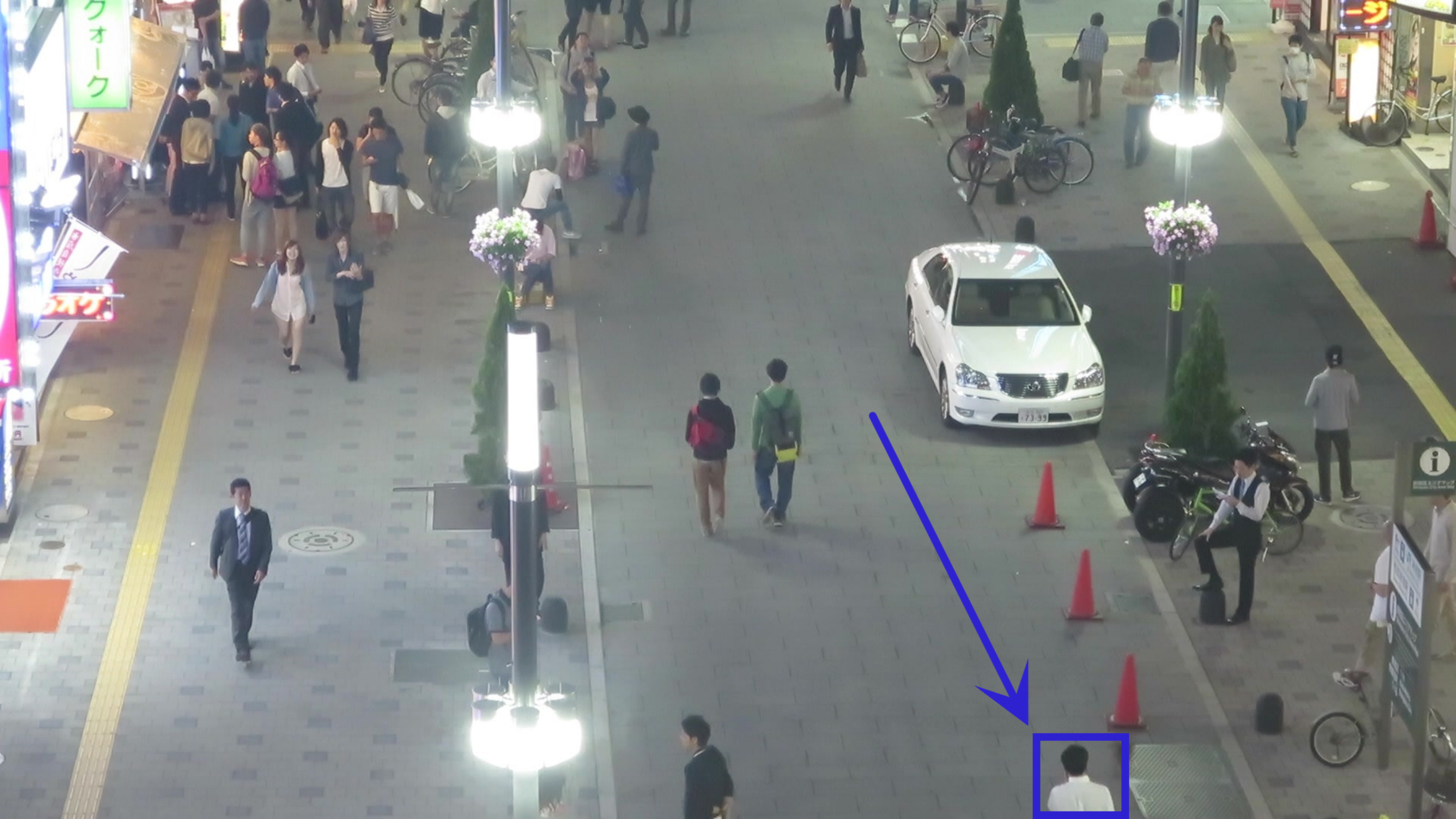}
  \caption{MOT17-04, frame: 000870}
  \label{fig:MOT17-04, frame: 000870}
\end{subfigure}
\newline
\begin{subfigure}{1.0\linewidth}
  \centering
  \includegraphics[width=0.8\linewidth]{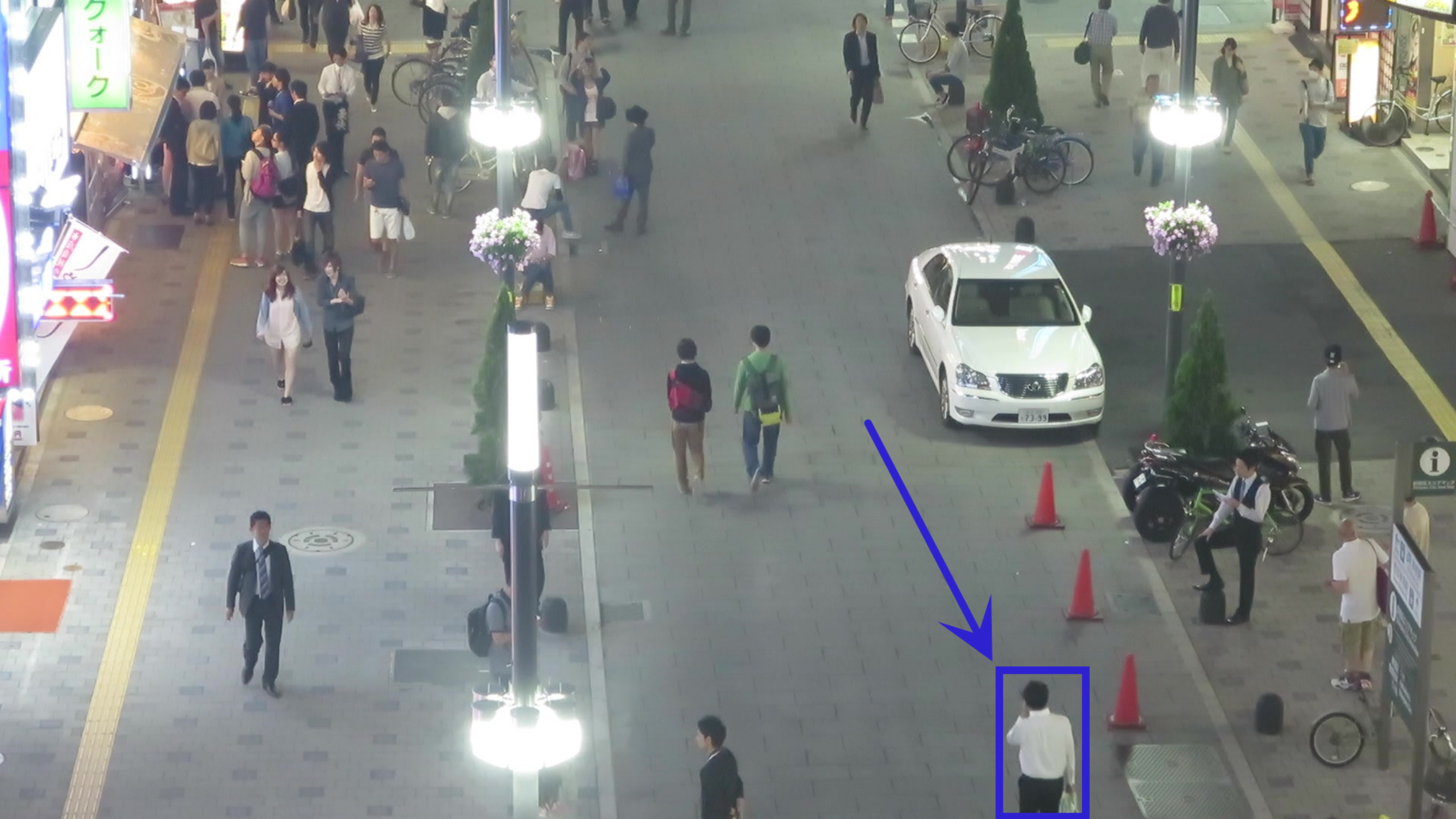}
  \caption{MOT17-04, frame: 000900}
  \label{fig:MOT17-04, frame: 000900}
\end{subfigure}
\newline
\begin{subfigure}{1.0\linewidth}
  \centering
  \includegraphics[width=0.8\linewidth]{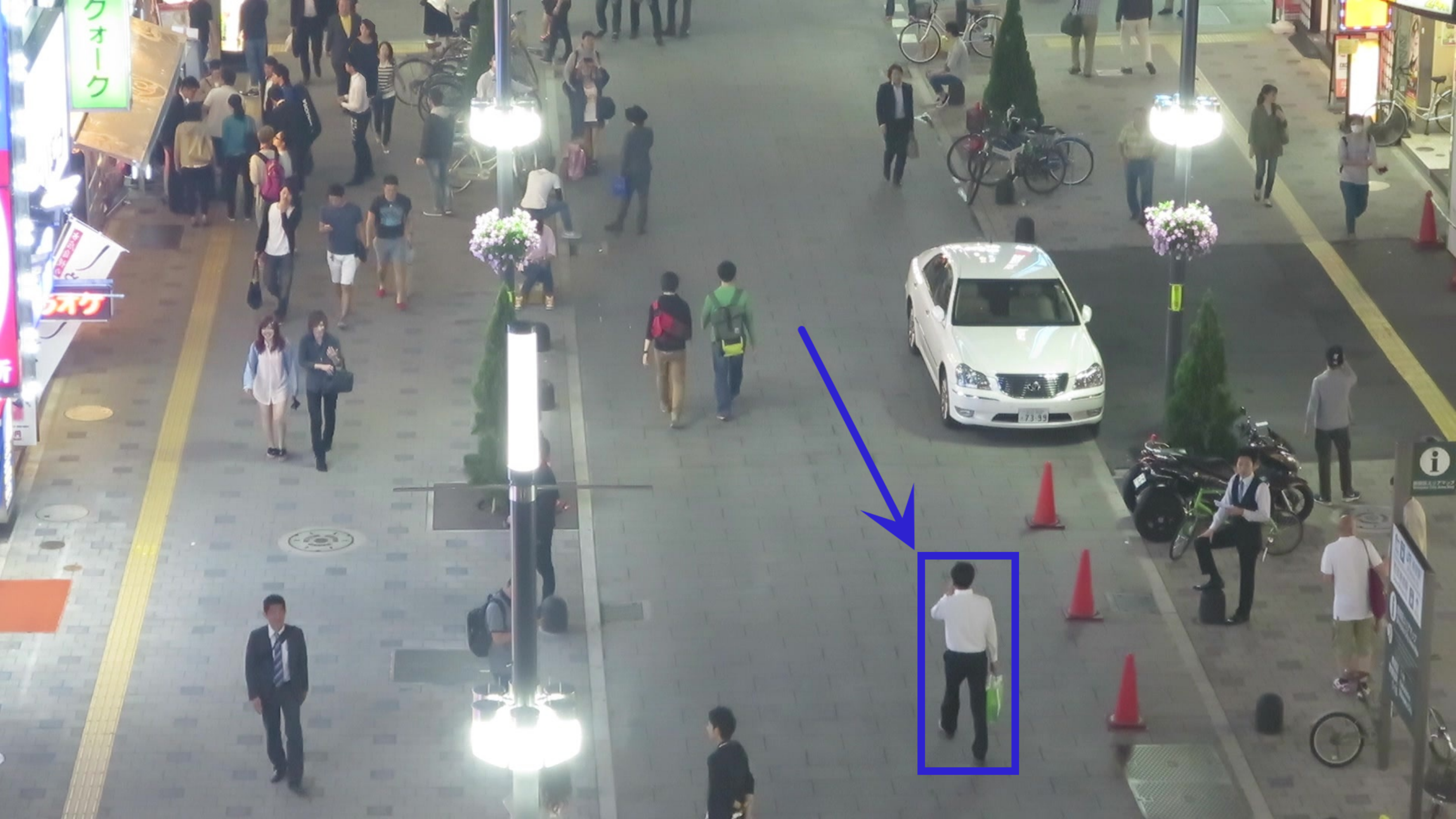}
  \caption{MOT17-04, frame: 000960}
  \label{fig:MOT17-04, frame: 000960}
\end{subfigure}
\newline
\begin{subfigure}{1.0\linewidth}
  \centering
  \includegraphics[width=0.8\linewidth]{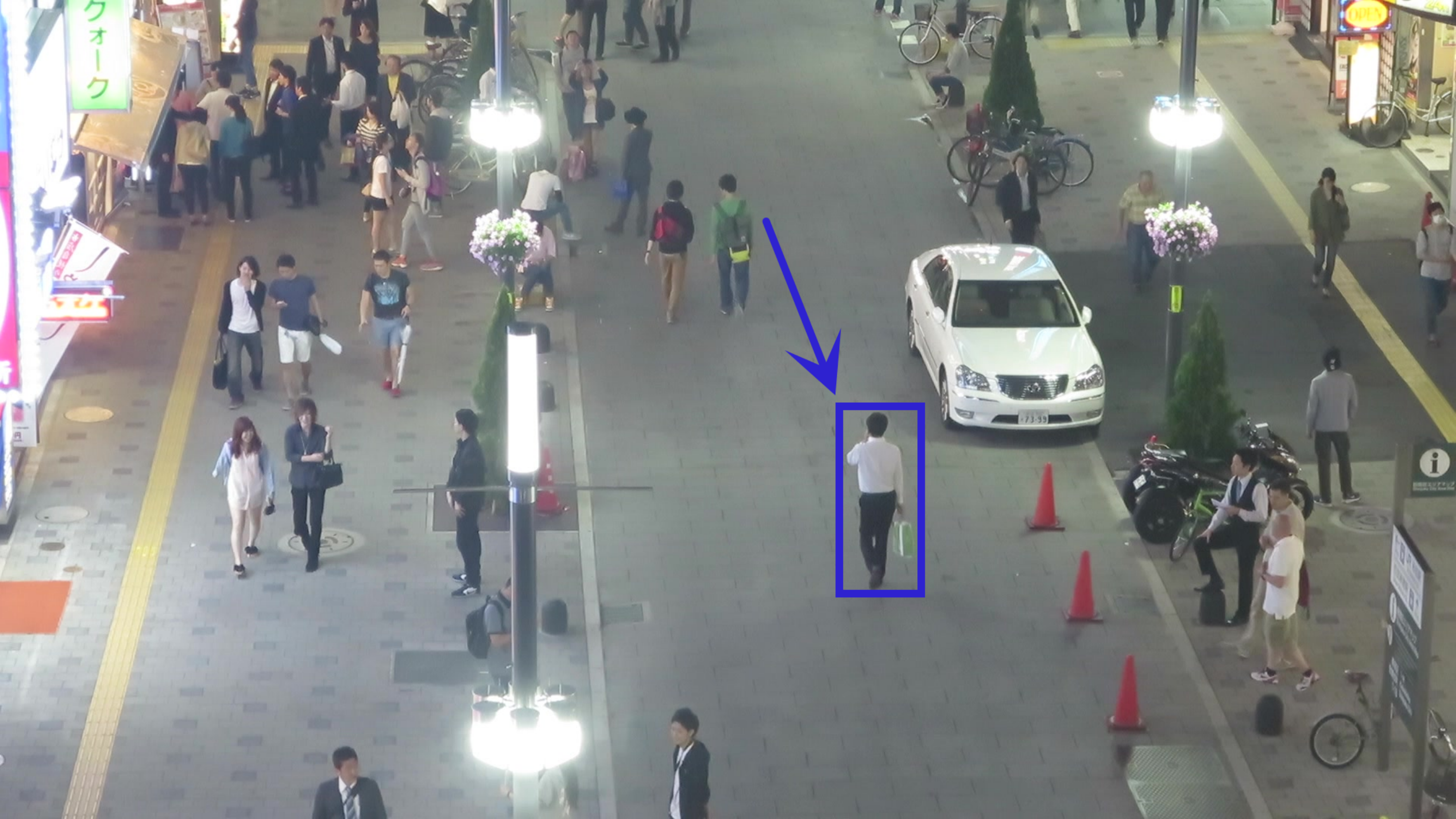}
  \caption{MOT17-04, frame: 001050}
  \label{fig:MOT17-04, frame: 001050}
\end{subfigure}
\caption{Example of prediction continuation in SVA. The pedestrian in \textcolor{red}{red} box represents the predicted pedestrian, and the pedestrian in \textcolor{blue}{blue} box represents the original pedestrian.}
\label{fig:Example of prediction continuation in the Stationary Camera View Data Augmentation (SVA).}
\end{figure}

\begin{figure*}[ht]
\begin{subfigure}{.49\linewidth}
  \centering
  \includegraphics[width=0.92\linewidth]{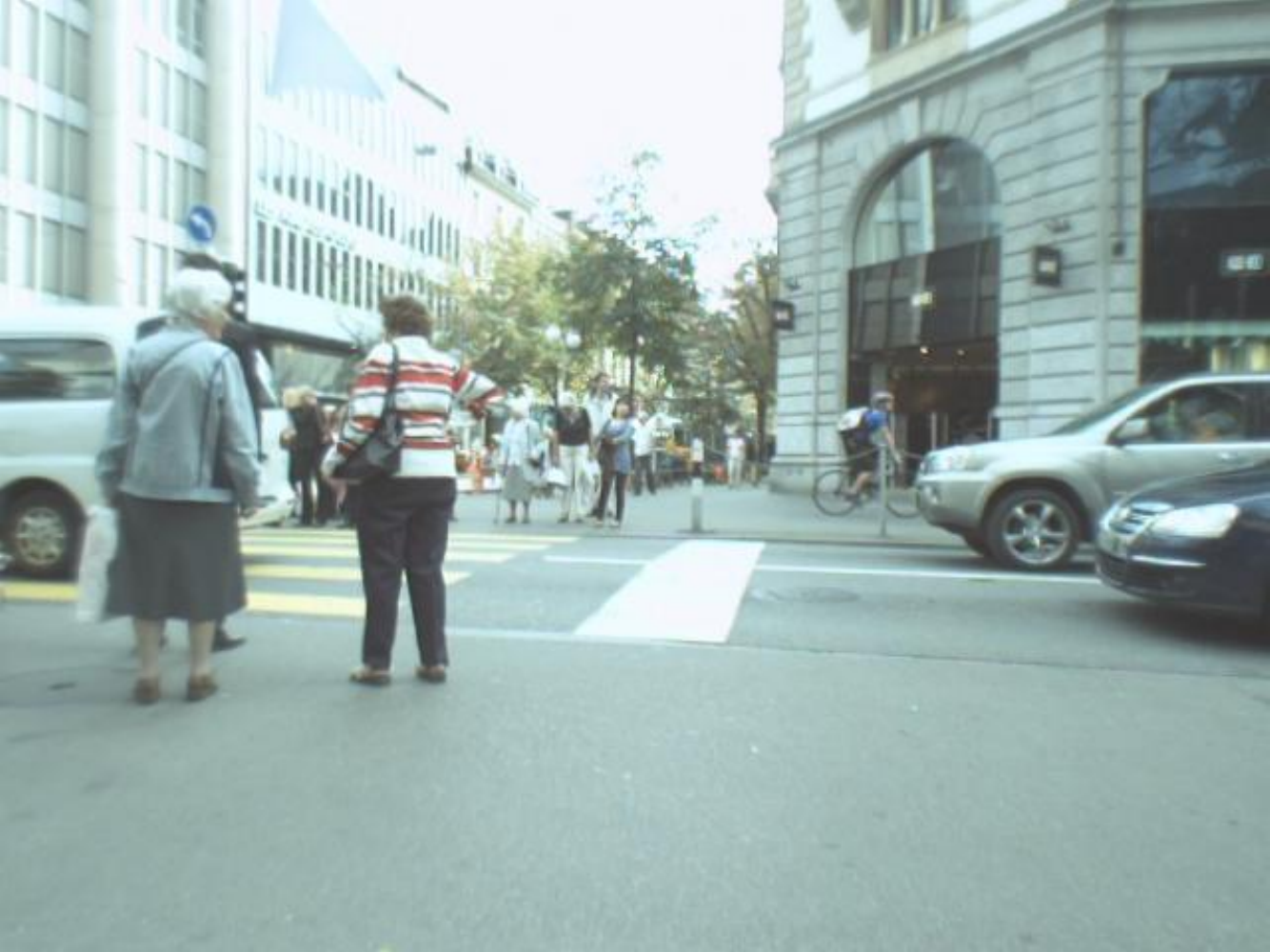}
  \caption{The original image, MOT17-05, frame: 000010}
  \label{fig:The original image, MOT17-05, frame: 000010}
\end{subfigure}
\hfill
\begin{subfigure}{.49\linewidth}
  \centering
  \includegraphics[width=0.92\linewidth]{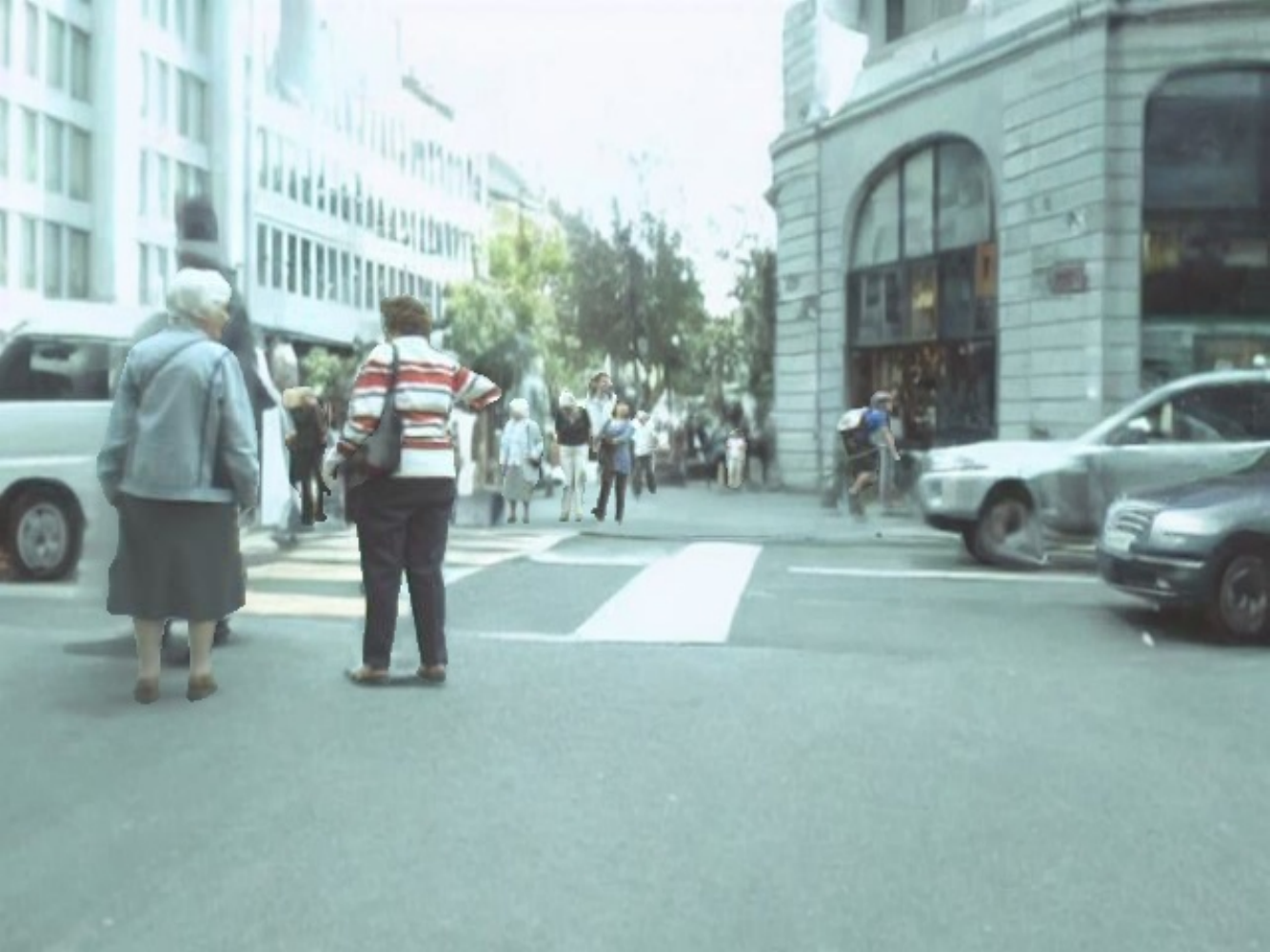}
  \caption{The image after DVA processing, MOT17-05, frame: 000010}
  \label{fig:The image after DVA processing, MOT17-05, frame: 000010}
\end{subfigure}
\newline
\begin{subfigure}{.49\linewidth}
  \centering
  \includegraphics[width=0.92\linewidth]{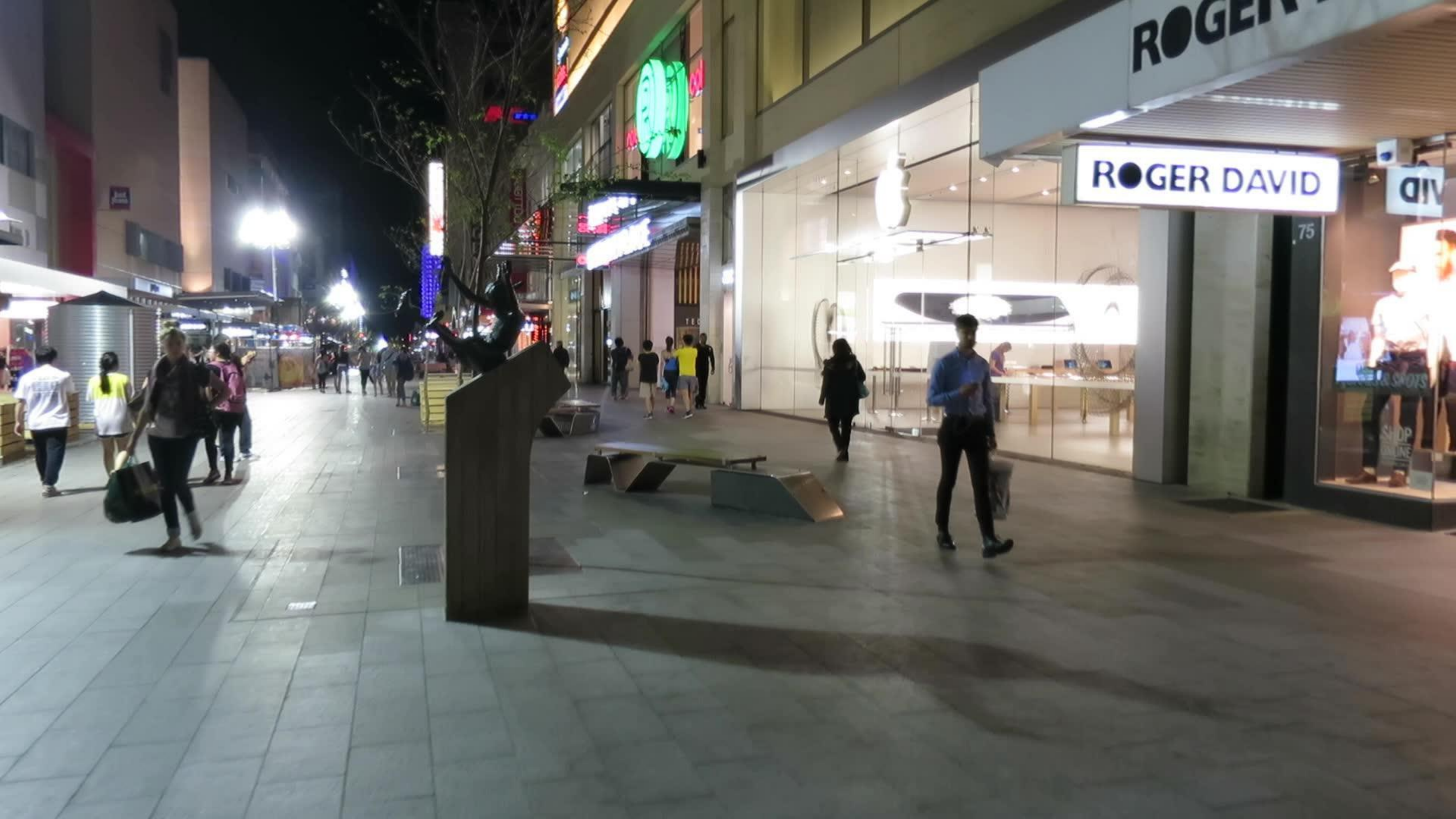}
  \caption{The original image, MOT17-10, frame: 000043}
  \label{fig:The original image, MOT17-10, frame: 000043}
\end{subfigure}
\hfill
\begin{subfigure}{.49\linewidth}
  \centering
  \includegraphics[width=0.92\linewidth]{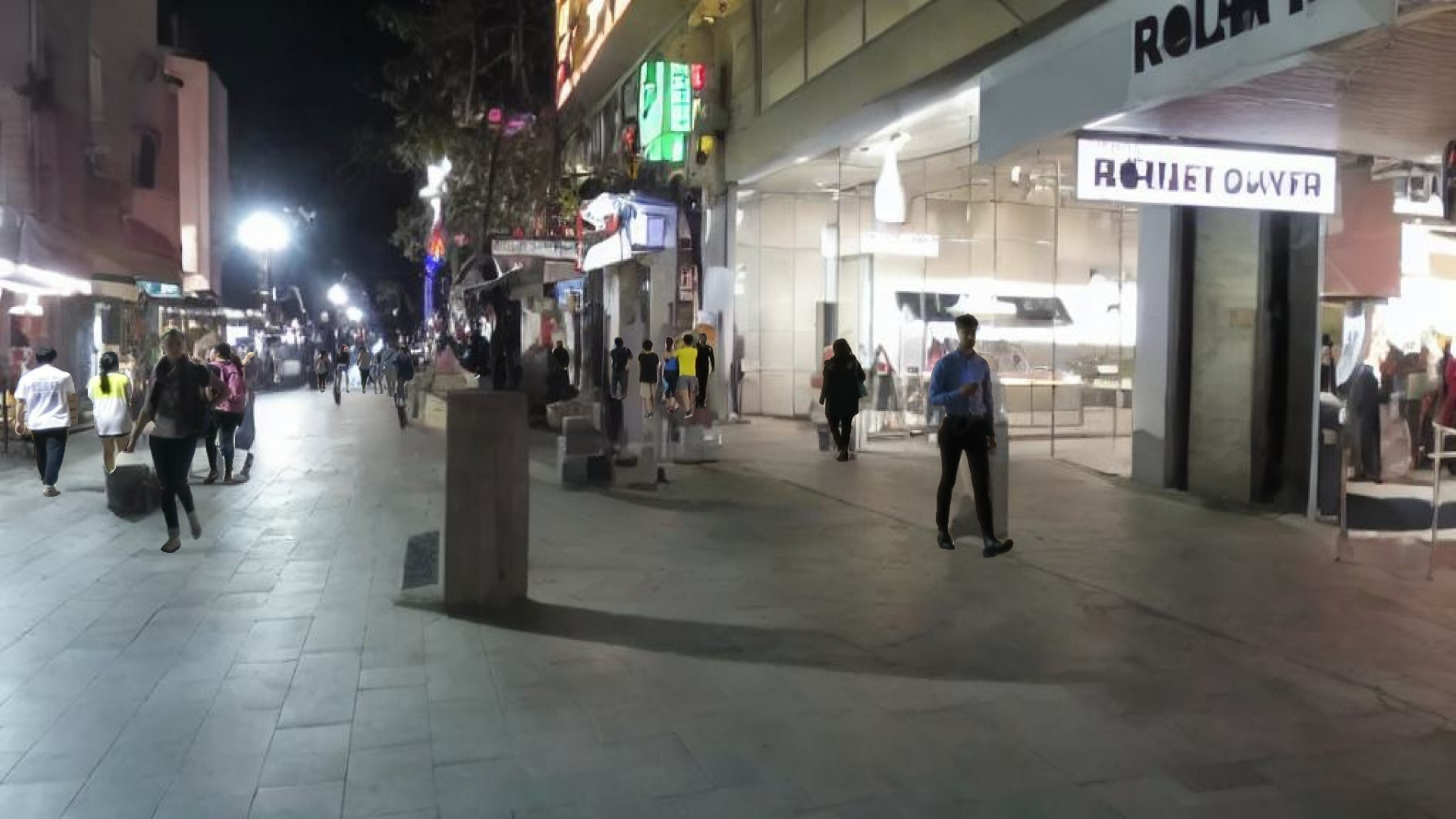}
  \caption{The image after DVA processing, MOT17-10, frame: 000043}
  \label{fig:The image after DVA processing, MOT17-10, frame: 000043}
\end{subfigure}
\newline
\begin{subfigure}{.49\linewidth}
  \centering
  \includegraphics[width=0.92\linewidth]{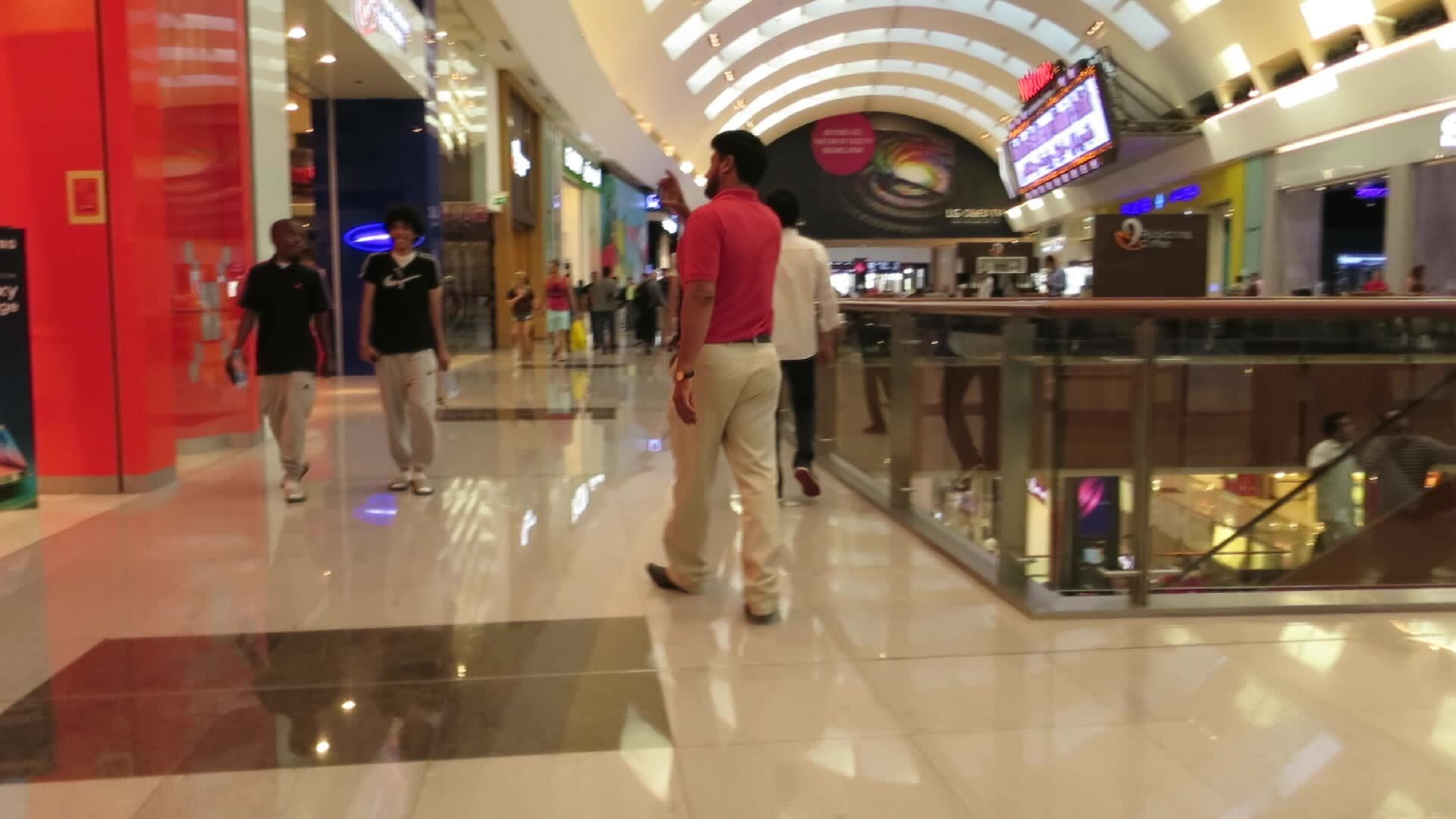}
  \caption{The original image, MOT17-11, frame: 000020}
  \label{fig:The original image, MOT17-11, frame: 000020}
\end{subfigure}
\hfill
\begin{subfigure}{.49\linewidth}
  \centering
  \includegraphics[width=0.92\linewidth]{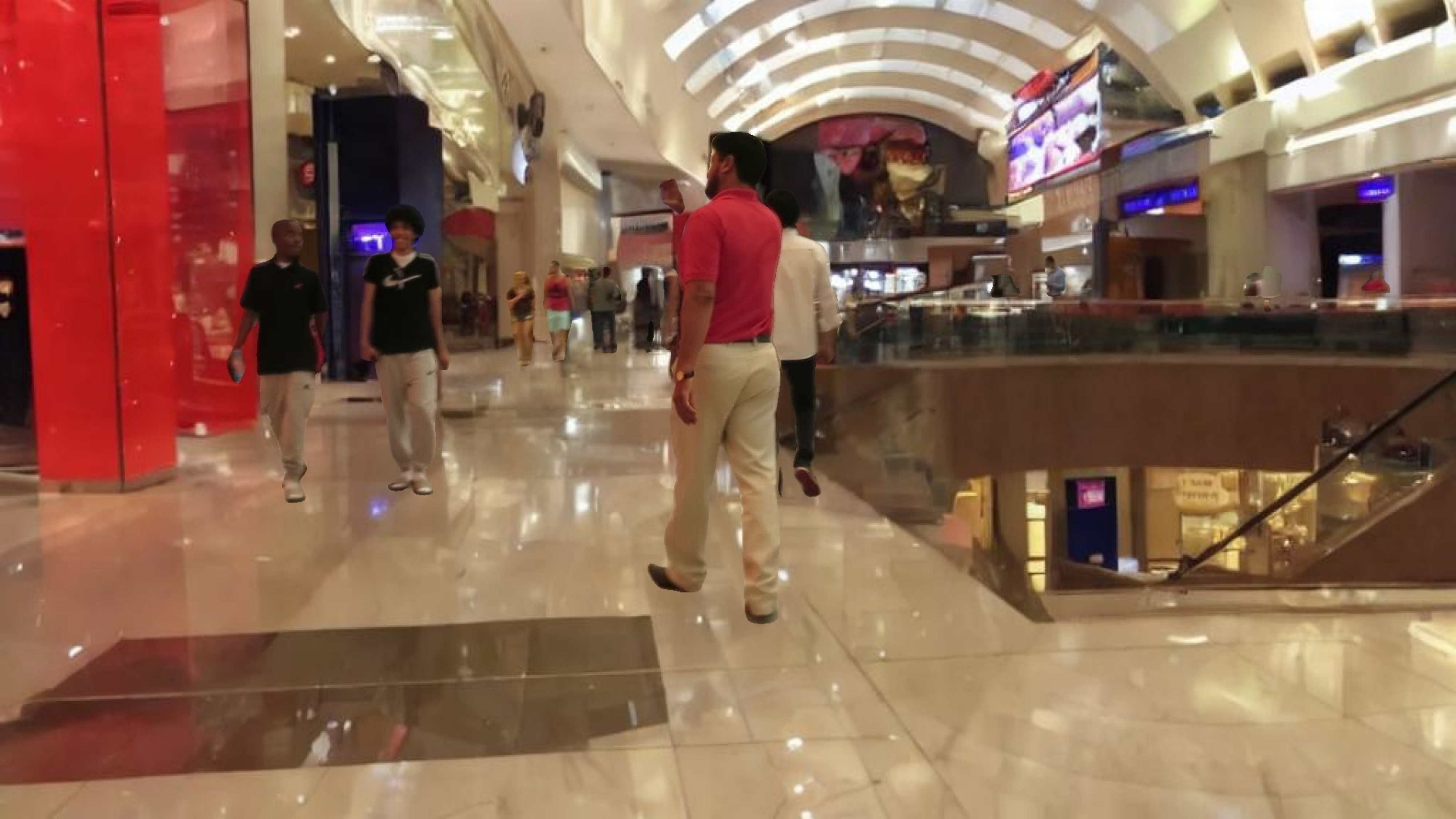}
  \caption{The image after DVA processing, MOT17-11, frame: 000020}
  \label{fig:The image after DVA processing, MOT17-11, frame: 000020}
\end{subfigure}
\newline
\begin{subfigure}{.49\linewidth}
  \centering
  \includegraphics[width=0.92\linewidth]{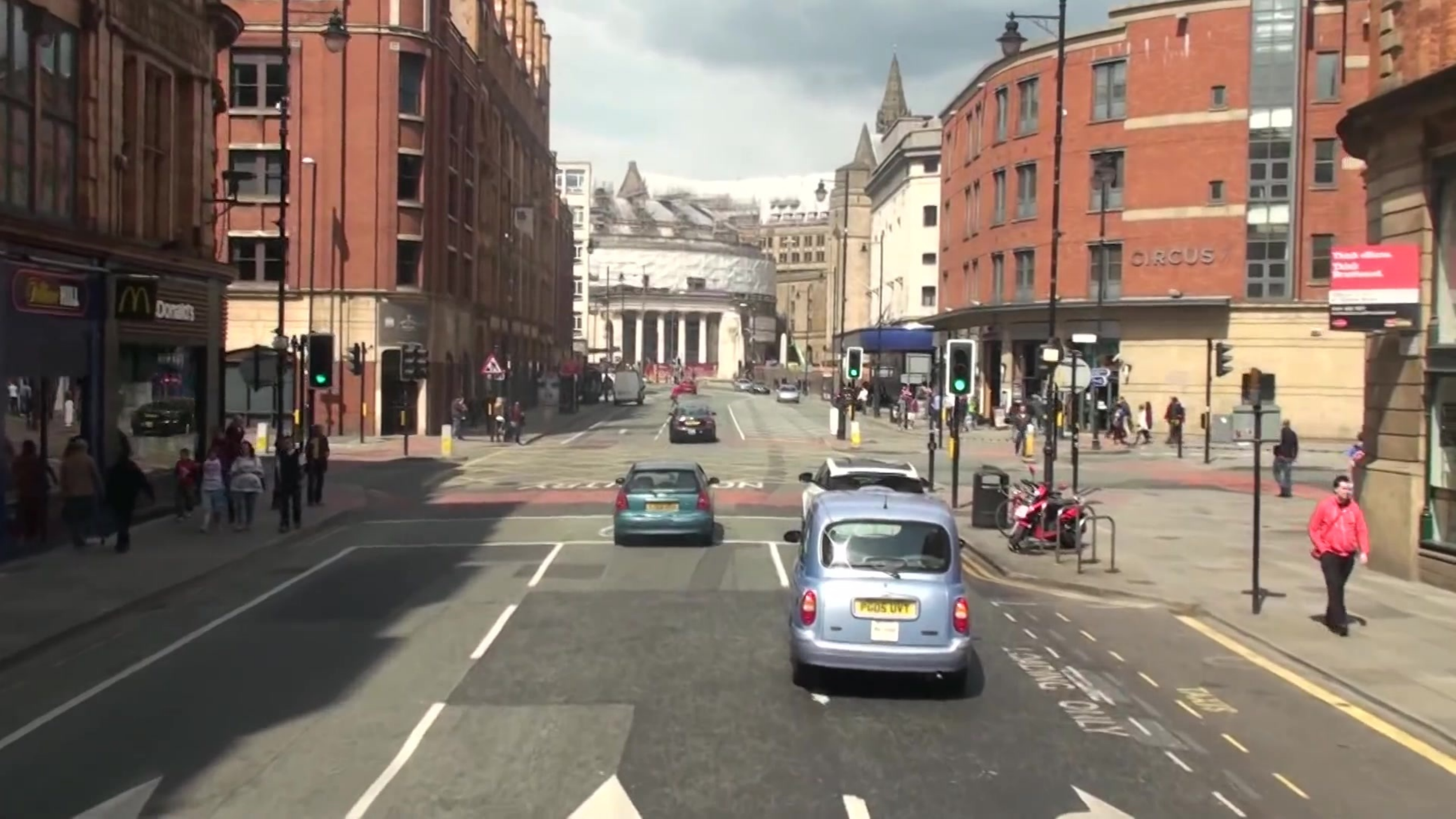}
  \caption{The original image, MOT17-13, frame: 000106}
  \label{fig:The original image, MOT17-13, frame: 000106}
\end{subfigure}
\hfill
\begin{subfigure}{.49\linewidth}
  \centering
  \includegraphics[width=0.92\linewidth]{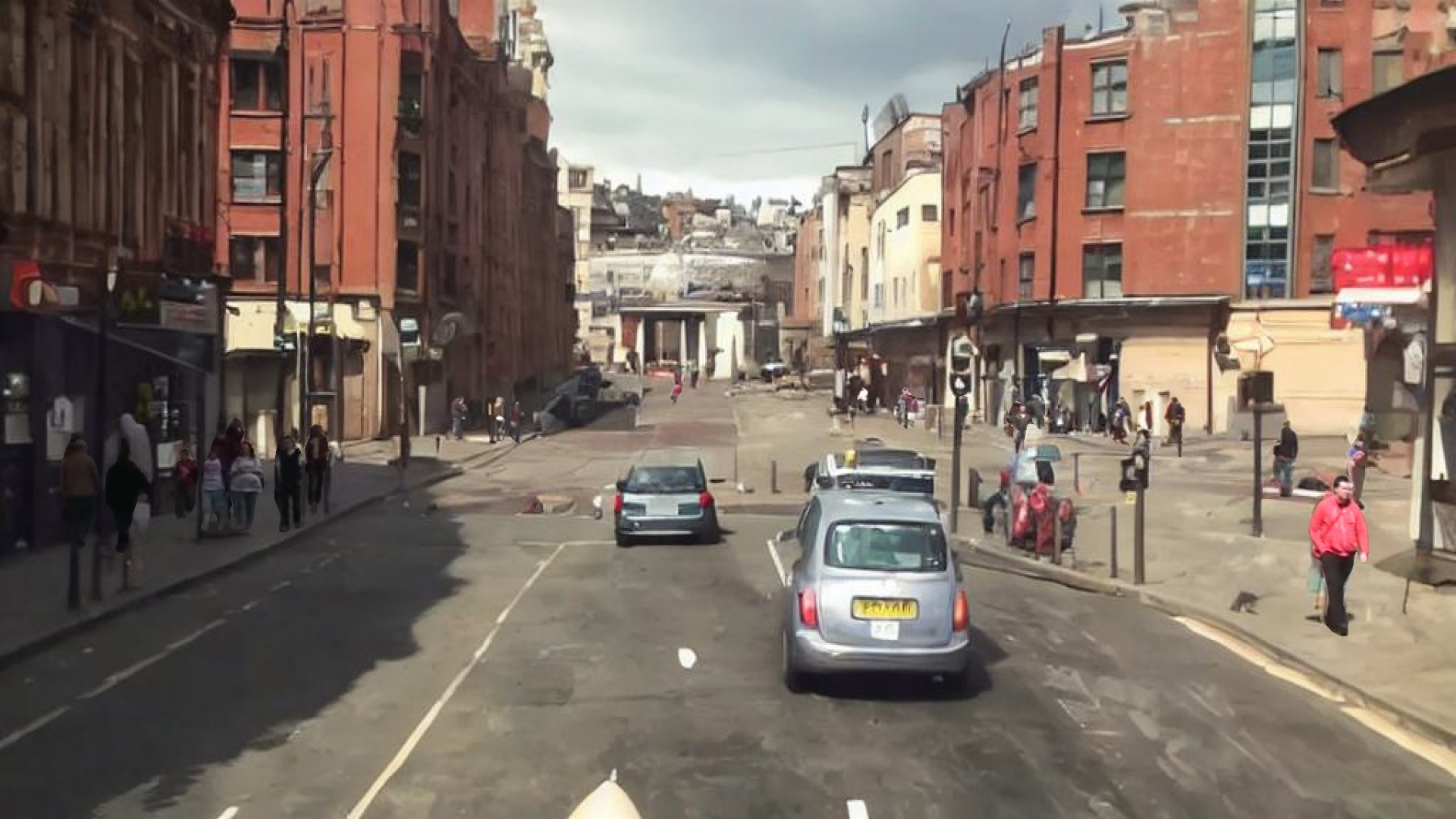}
  \caption{The image after DVA processing, MOT17-13, frame: 000106}
  \label{fig:The image after DVA processing, MOT17-13, frame: 000106}
\end{subfigure}
\caption{Comparison of the original image and the image processed by DVA.}
\label{fig:Comparison of the original image and the image processed by DVA.}
\end{figure*}

\begin{figure*}[ht]
\begin{subfigure}{.32\linewidth}
  \centering
  \includegraphics[width=1.0\linewidth]{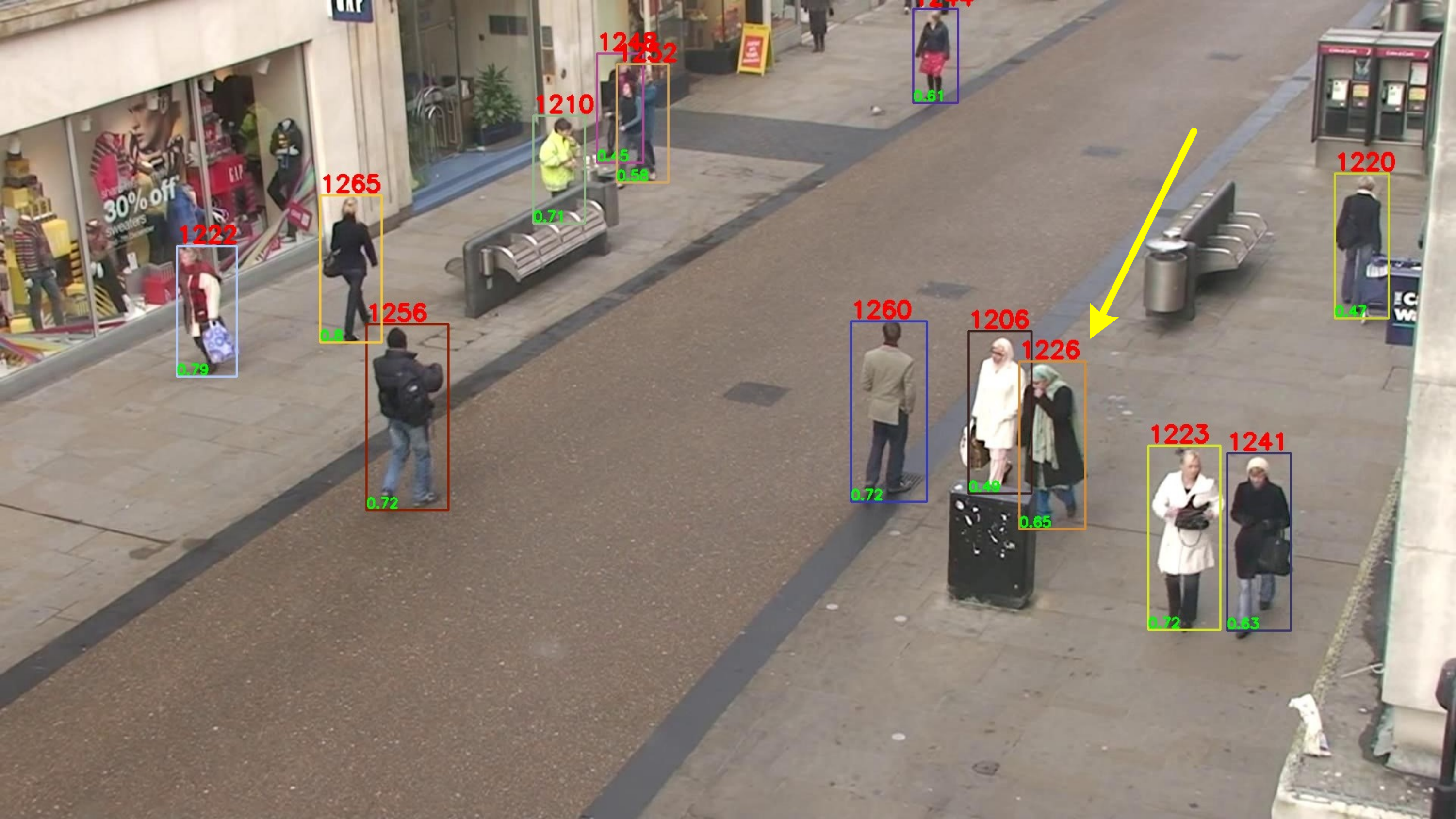}
  \caption{Baseline, MOT15-AVG-000405}
  \label{fig:Baseline, MOT15-AVG-000405}
\end{subfigure}
\hfill
\begin{subfigure}{.32\linewidth}
  \centering
  \includegraphics[width=1.0\linewidth]{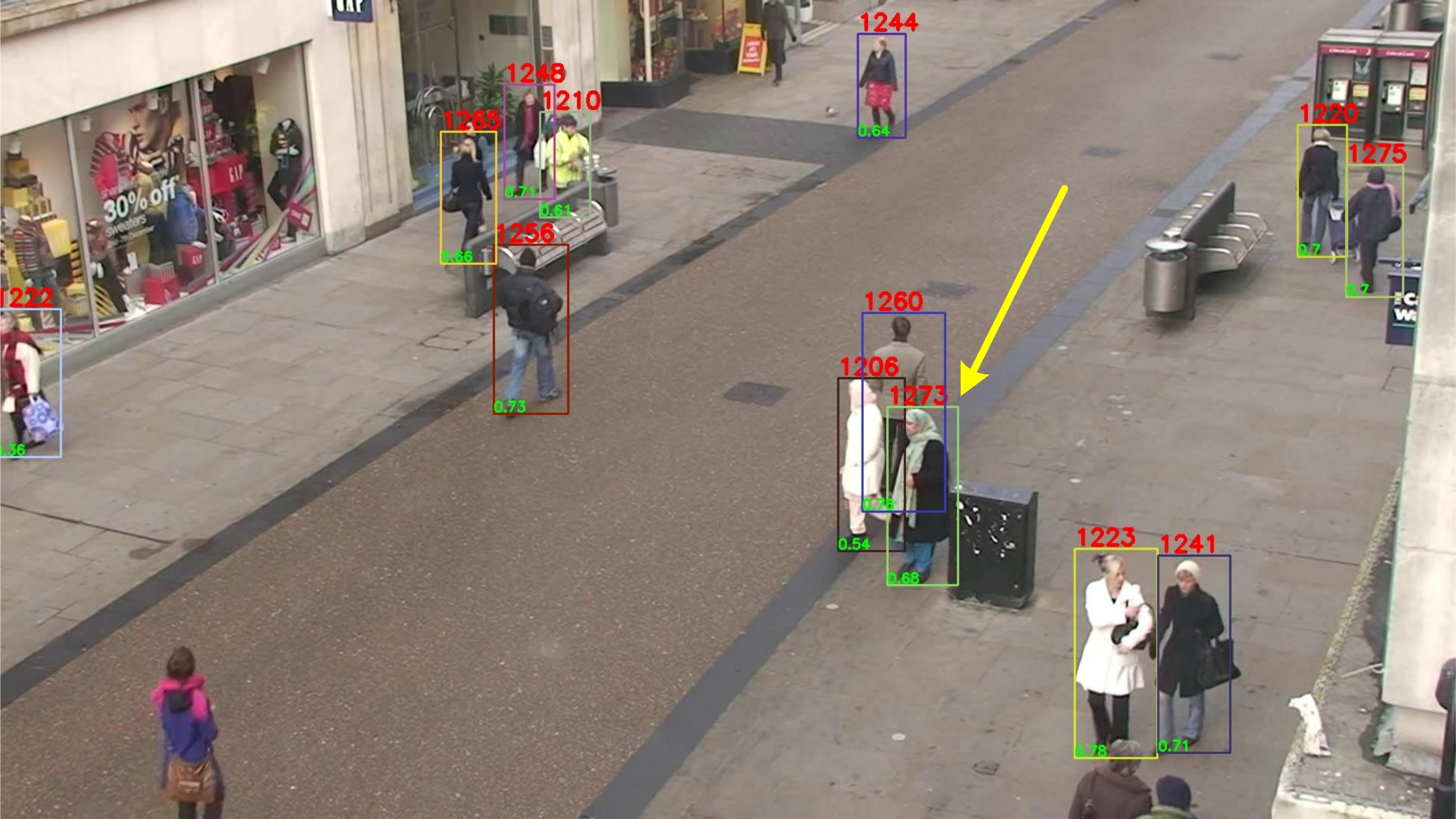}
  \caption{Baseline, MOT15-AVG-000410}
  \label{fig:Baseline, MOT15-AVG-000410}
\end{subfigure}
\hfill
\begin{subfigure}{.32\linewidth}
  \centering
  \includegraphics[width=1.0\linewidth]{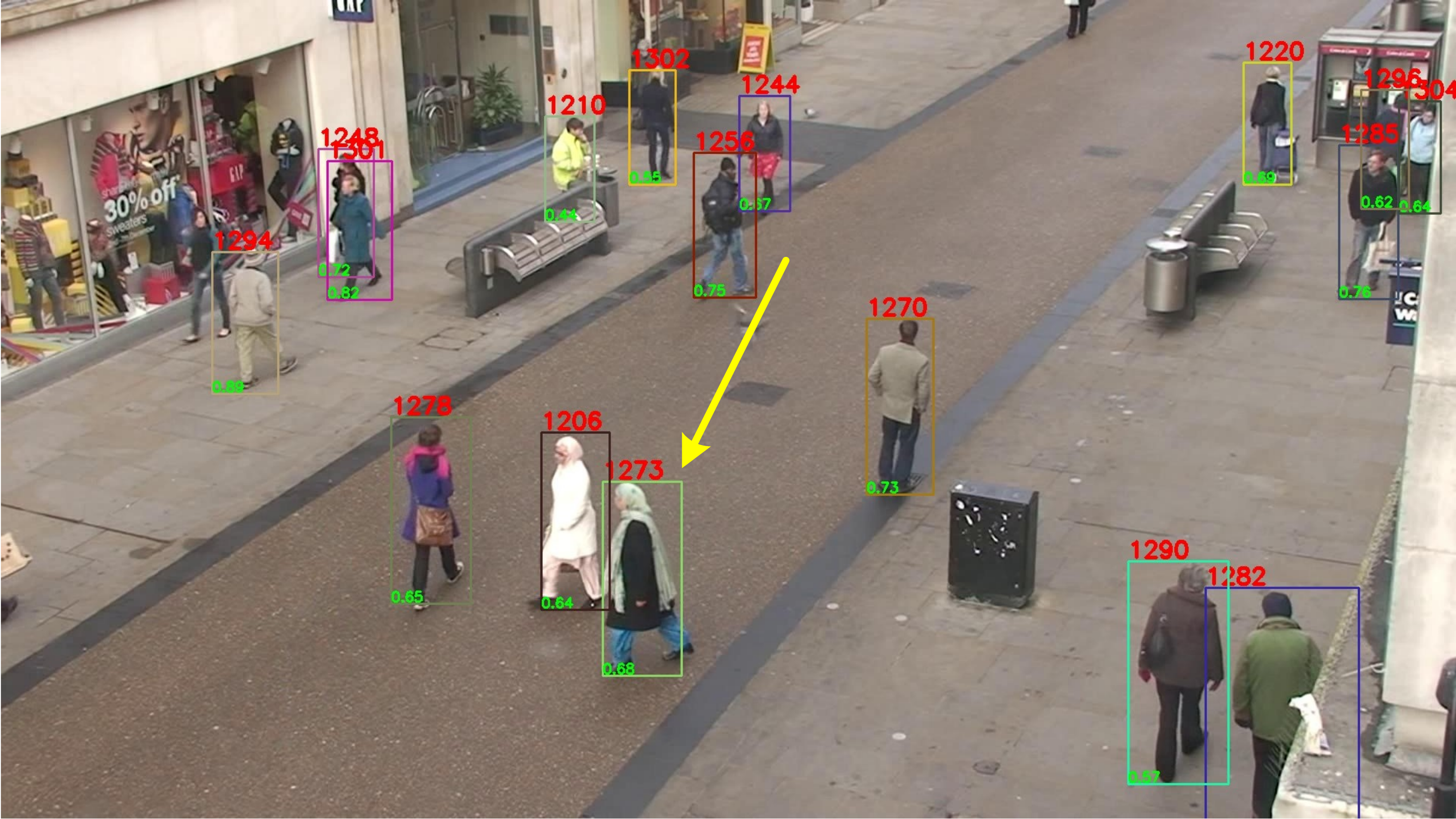}
  \caption{Baseline, MOT15-AVG-000418}
  \label{fig:Baseline, MOT15-AVG-000418}
\end{subfigure}
\newline
\begin{subfigure}{.32\linewidth}
  \centering
  \includegraphics[width=1.0\linewidth]{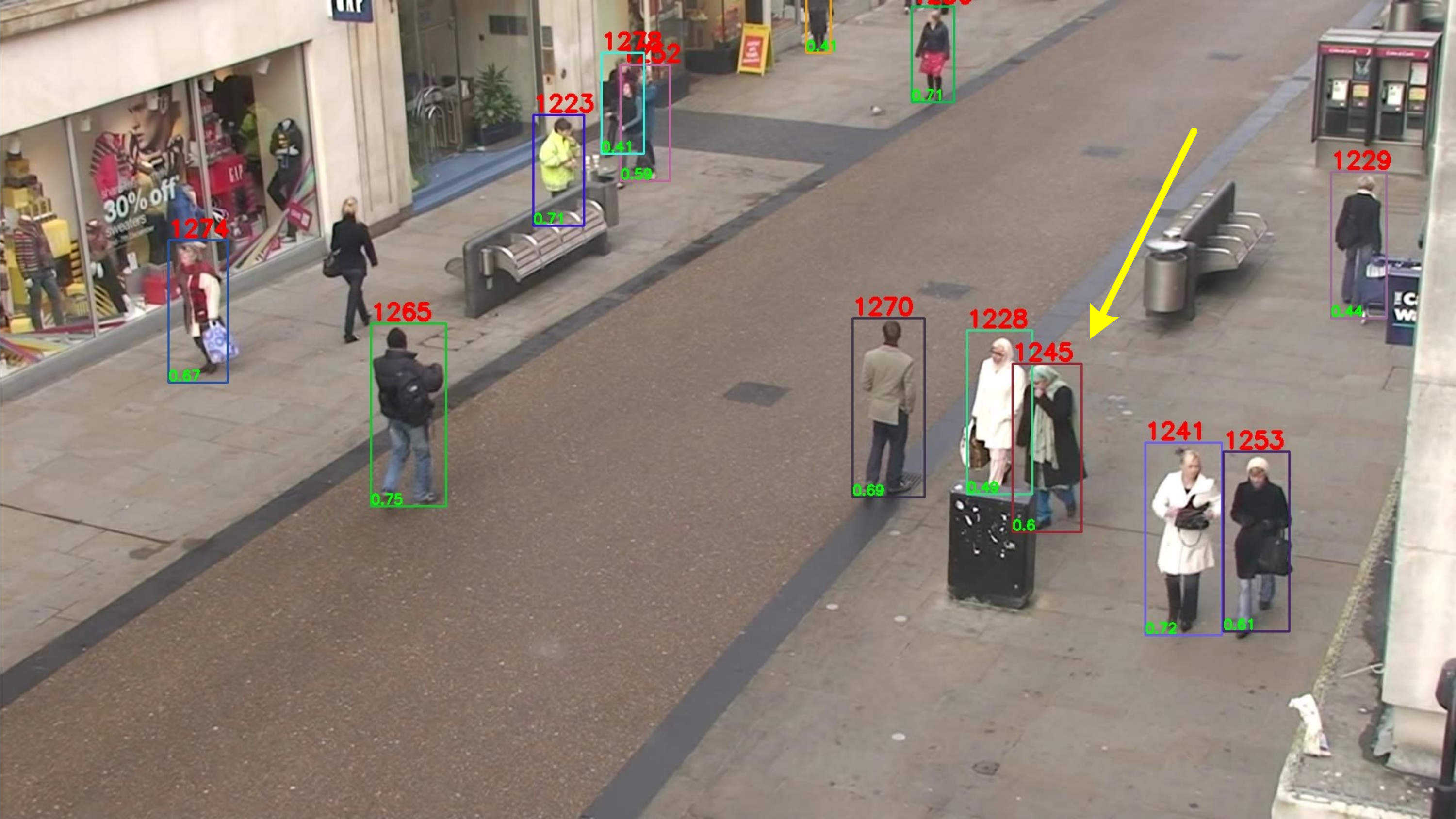}
  \caption{Our method, MOT15-AVG-000405}
  \label{fig:Our method, MOT15-AVG-000405}
\end{subfigure}
\hfill
\begin{subfigure}{.32\linewidth}
  \centering
  \includegraphics[width=1.0\linewidth]{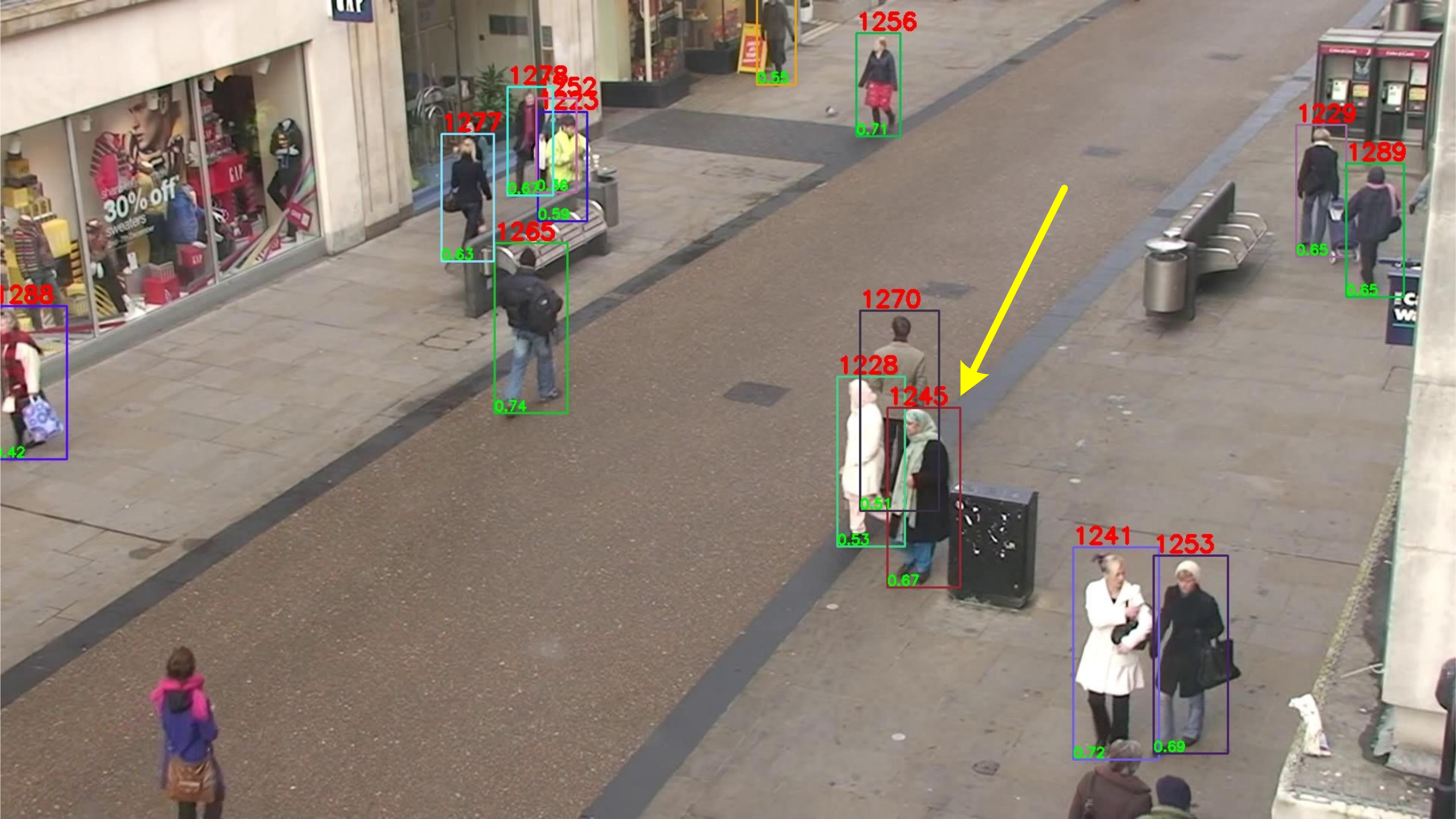}
  \caption{Our method, MOT15-AVG-000410}
  \label{fig:Our method, MOT15-AVG-000410}
\end{subfigure}
\hfill
\begin{subfigure}{.32\linewidth}
  \centering
  \includegraphics[width=1.0\linewidth]{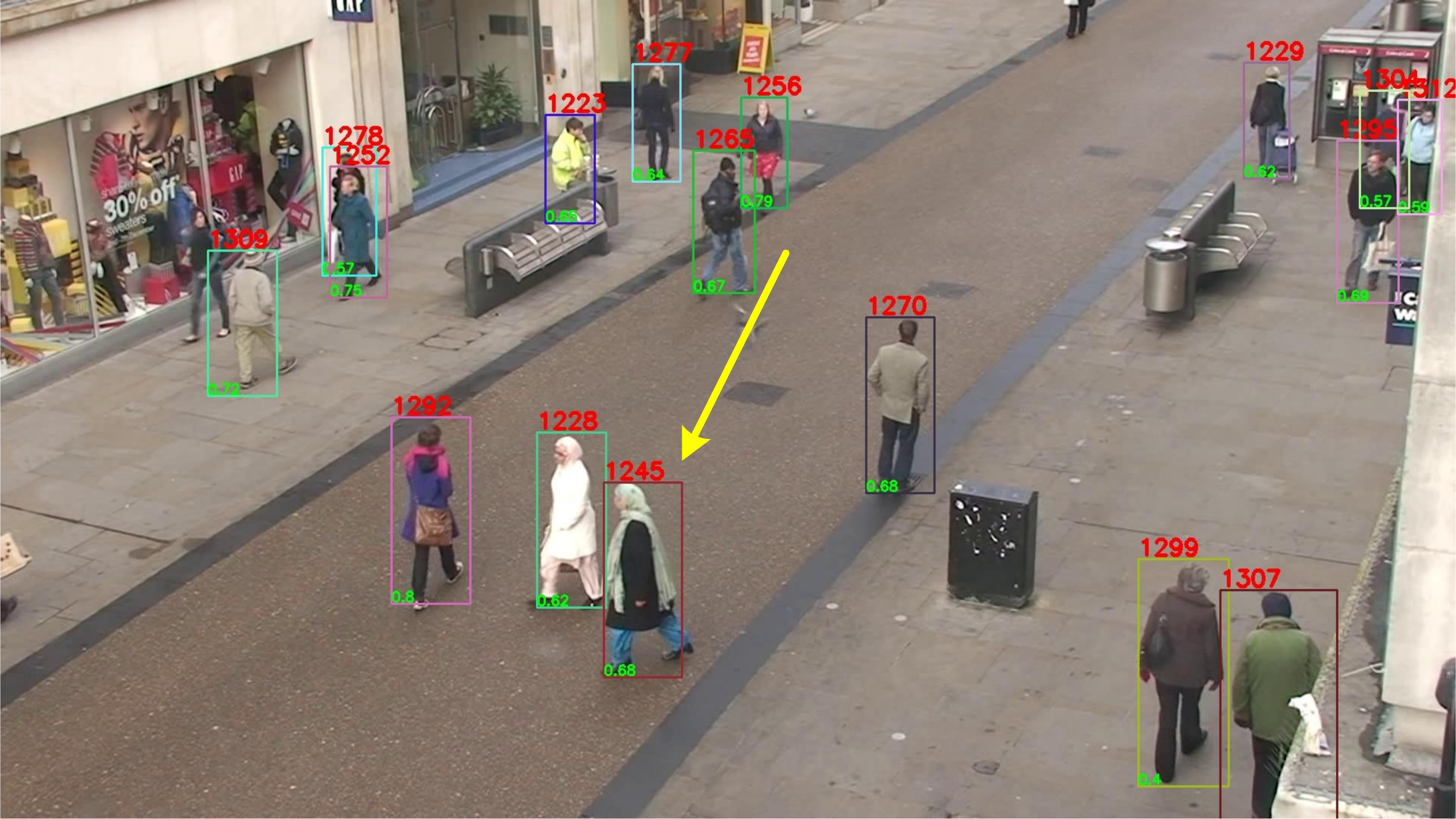}
  \caption{Our method, MOT15-AVG-000418}
  \label{fig:Our method, MOT15-AVG-000418}
\end{subfigure}
\newline
\begin{subfigure}{.32\linewidth}
  \centering
  \includegraphics[width=1.0\linewidth]{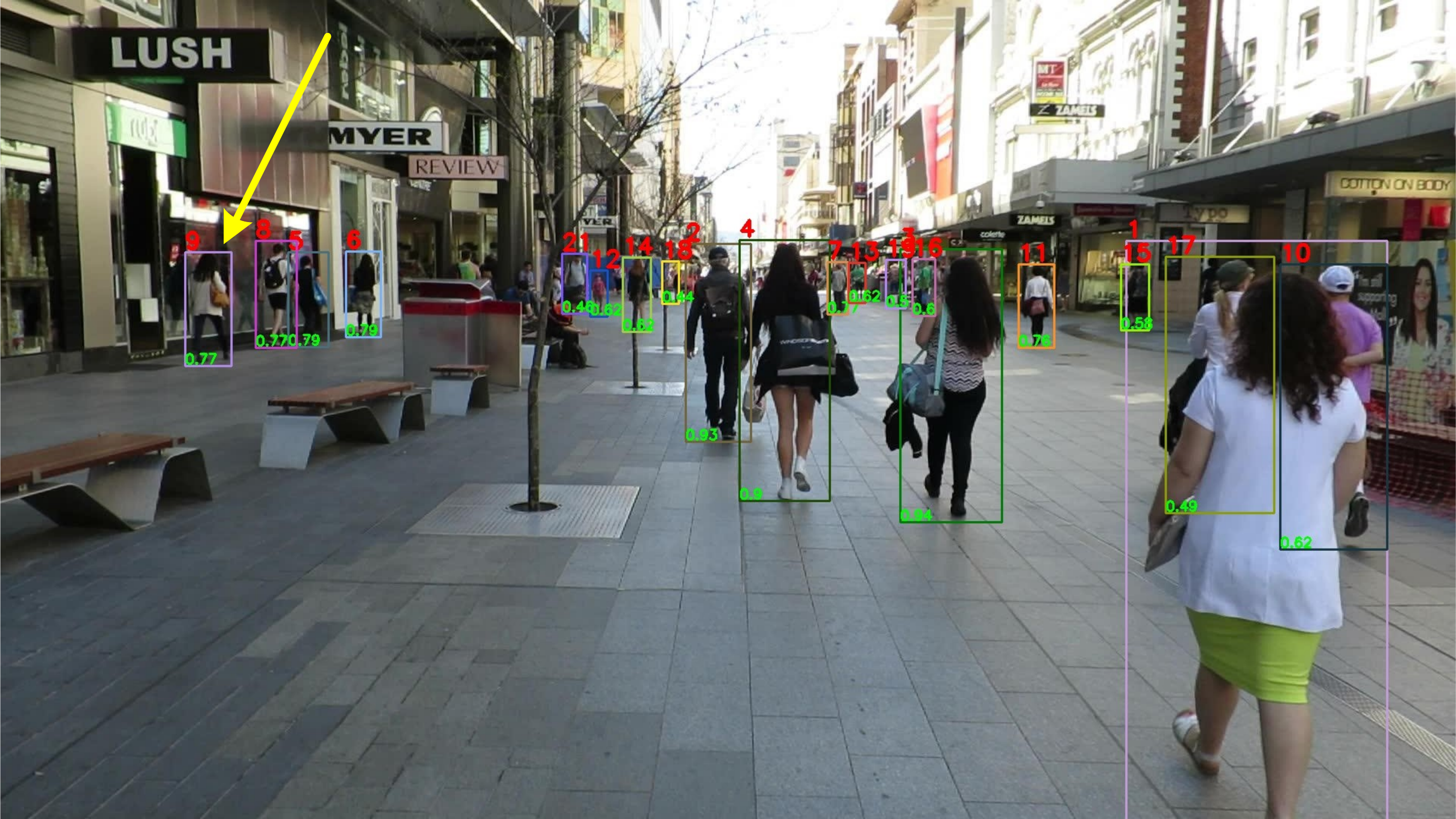}
  \caption{Baseline, MOT16-07-000001}
  \label{fig:Baseline, MOT16-07-000001}
\end{subfigure}
\hfill
\begin{subfigure}{.32\linewidth}
  \centering
  \includegraphics[width=1.0\linewidth]{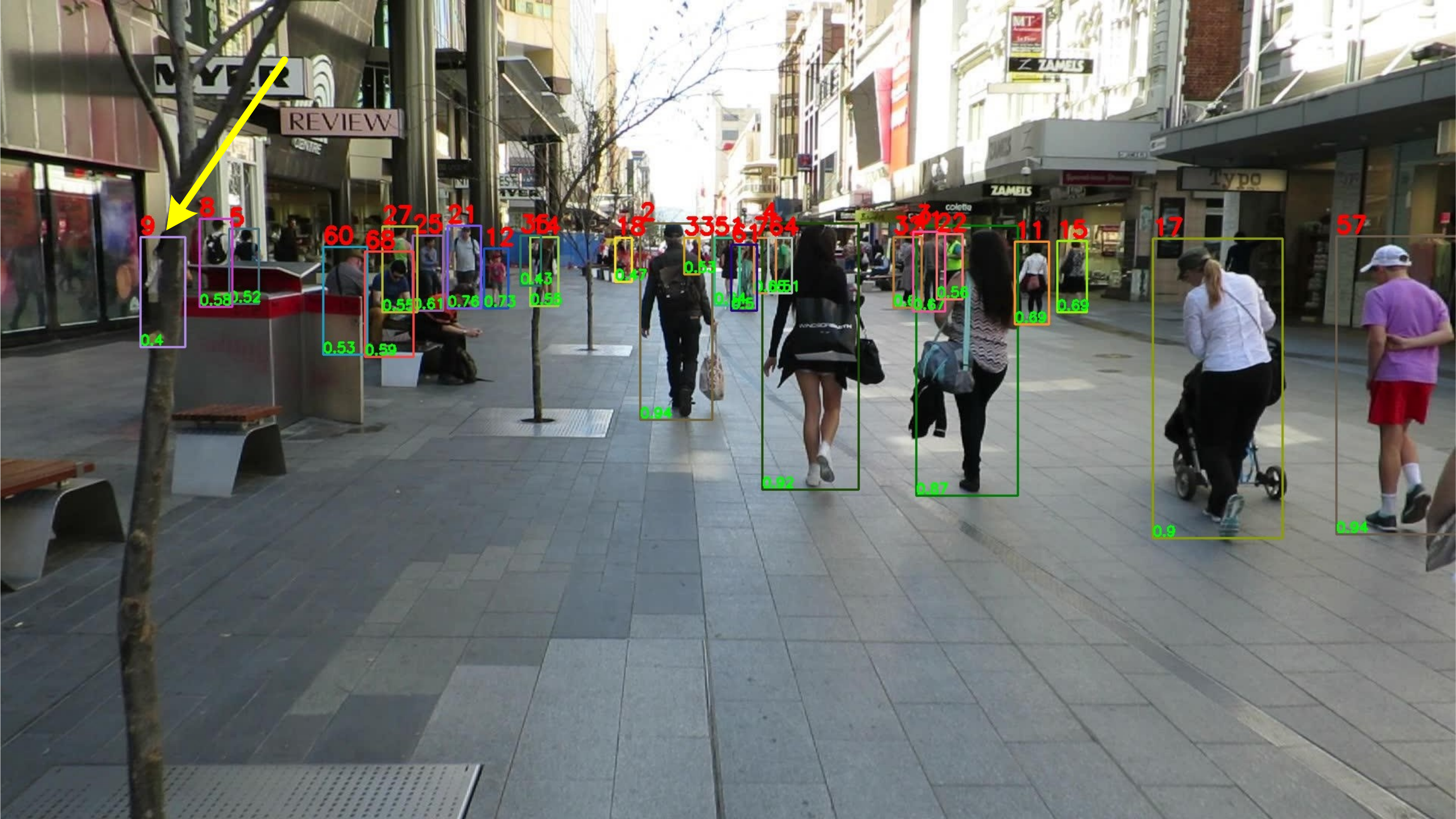}
  \caption{Baseline, MOT16-07-000085}
  \label{fig:Baseline, MOT16-07-000085}
\end{subfigure}
\hfill
\begin{subfigure}{.32\linewidth}
  \centering
  \includegraphics[width=1.0\linewidth]{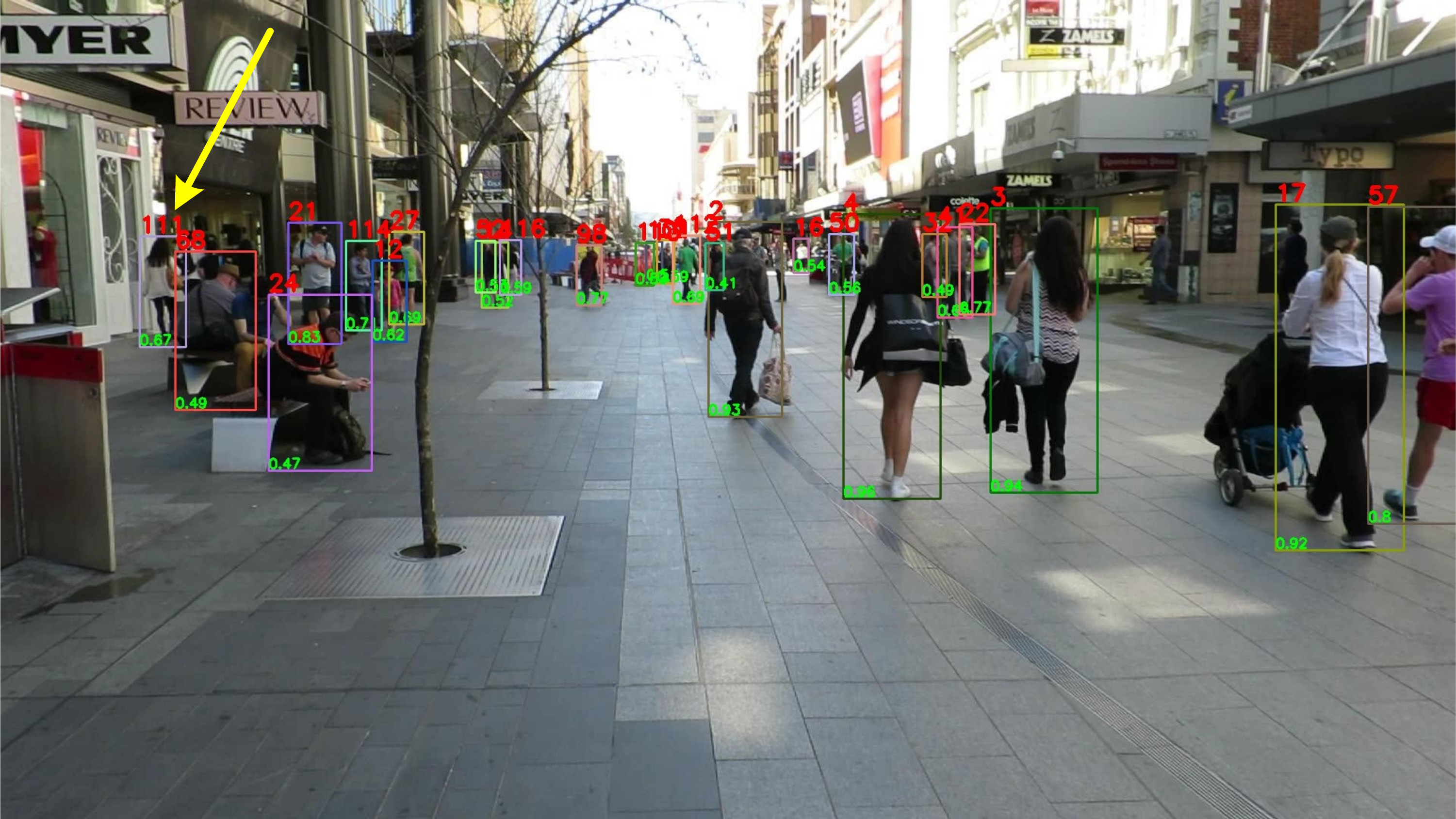}
  \caption{Baseline, MOT16-07-000160}
  \label{fig:Baseline, MOT16-07-000160}
\end{subfigure}
\newline
\begin{subfigure}{.32\linewidth}
  \centering
  \includegraphics[width=1.0\linewidth]{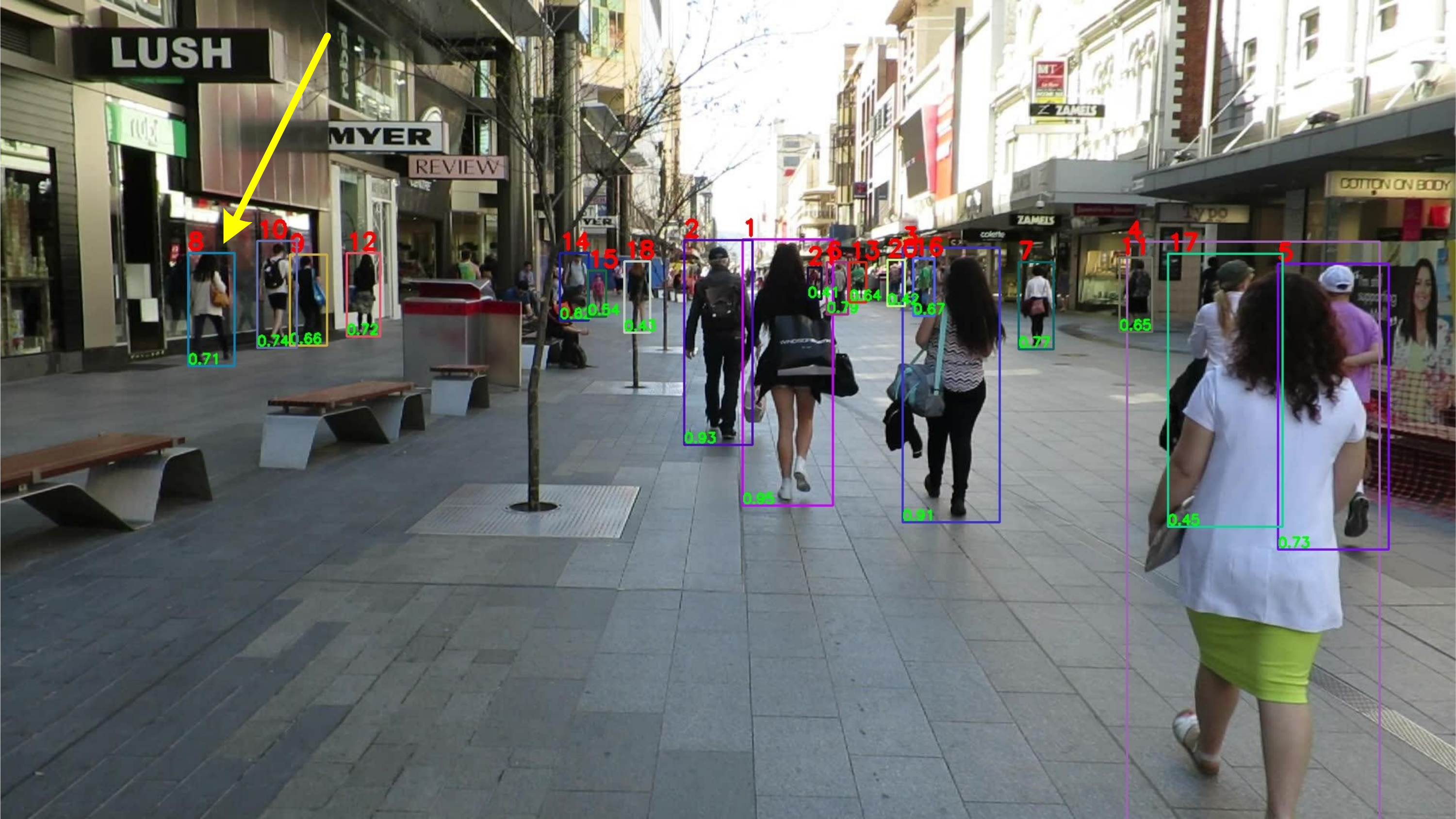}
  \caption{Our method, MOT16-07-000001}
  \label{fig:Our method, MOT16-07-000001}
\end{subfigure}
\hfill
\begin{subfigure}{.32\linewidth}
  \centering
  \includegraphics[width=1.0\linewidth]{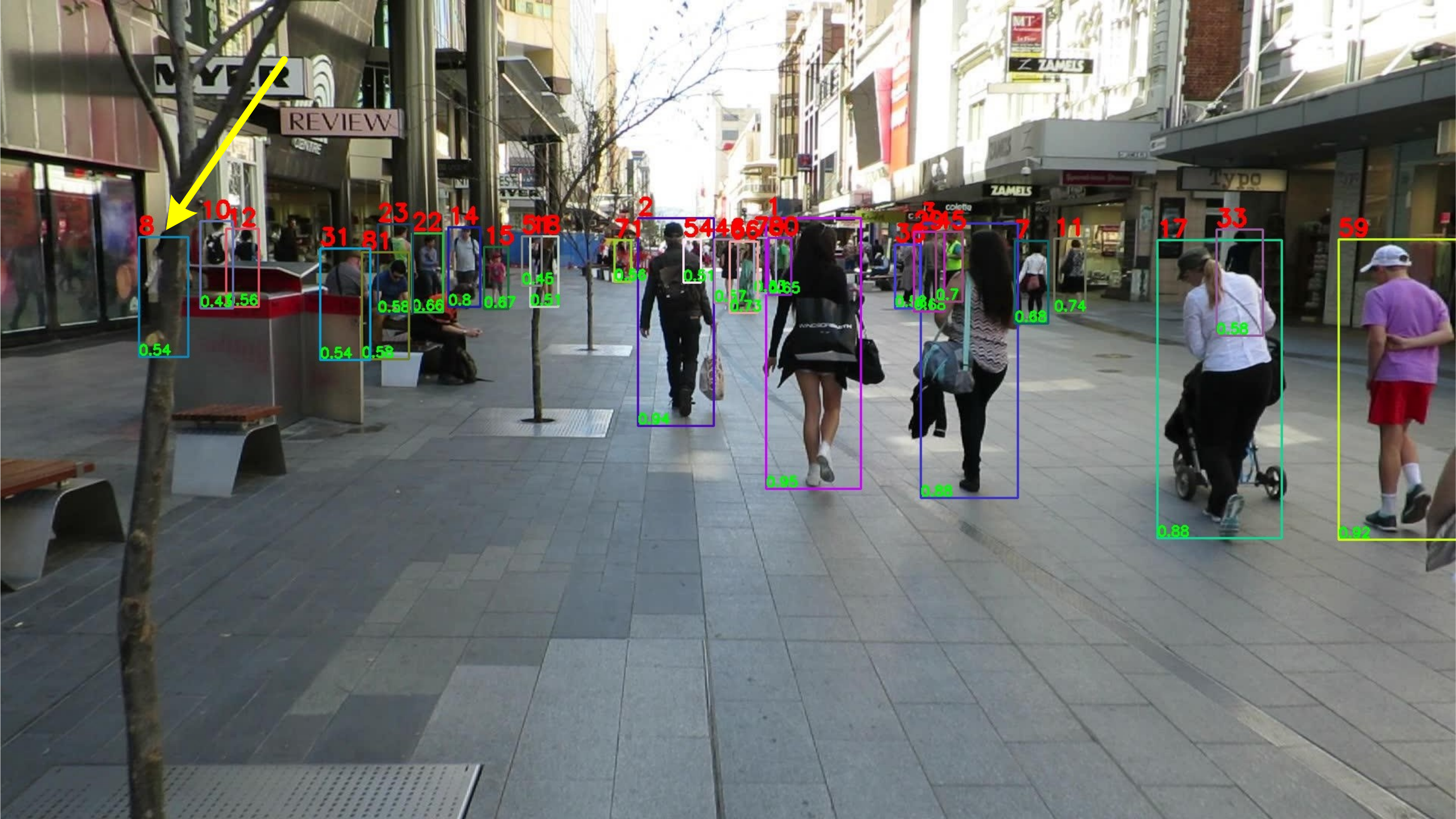}
  \caption{Our method, MOT16-07-000085}
  \label{fig:Our method, MOT16-07-000085}
\end{subfigure}
\hfill
\begin{subfigure}{.32\linewidth}
  \centering
  \includegraphics[width=1.0\linewidth]{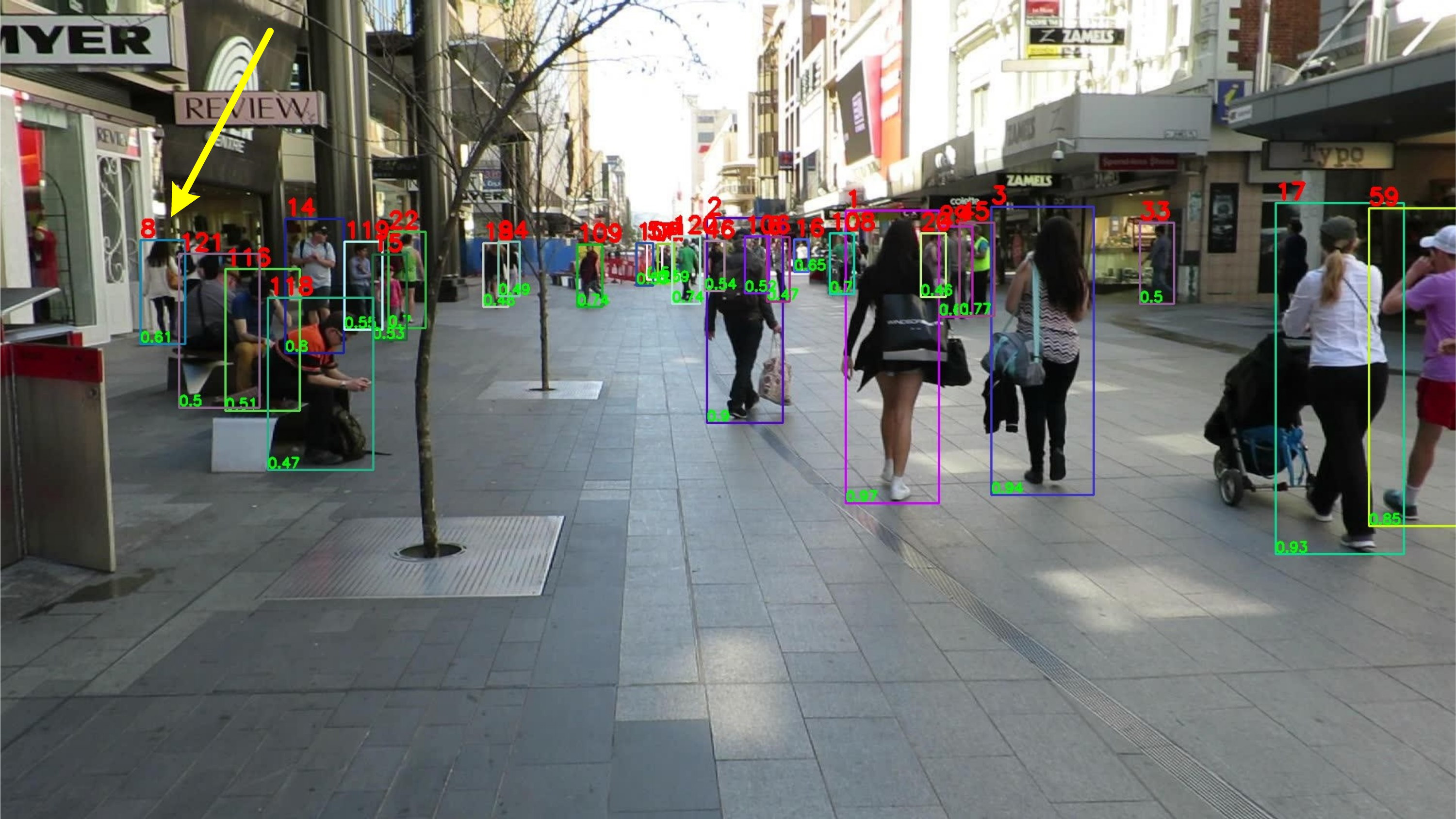}
  \caption{Our method, MOT16-07-000160}
  \label{fig:Our method, MOT16-07-000160}
\end{subfigure}
\newline
\begin{subfigure}{.32\linewidth}
  \centering
  \includegraphics[width=1.0\linewidth]{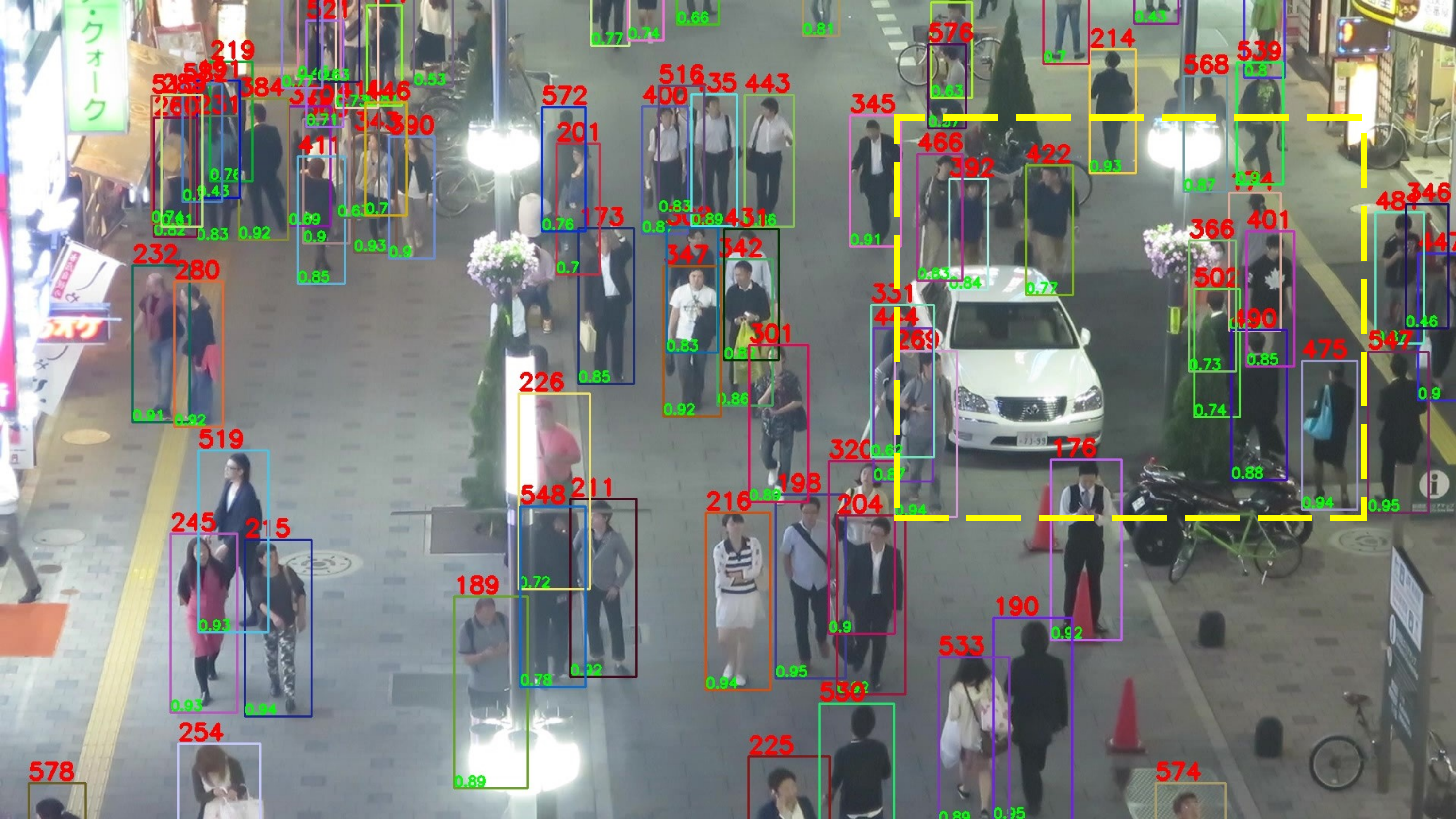}
  \caption{Baseline, MOT17-03-000750}
  \label{fig:Baseline, MOT17-03-000750}
\end{subfigure}
\hfill
\begin{subfigure}{.32\linewidth}
  \centering
  \includegraphics[width=1.0\linewidth]{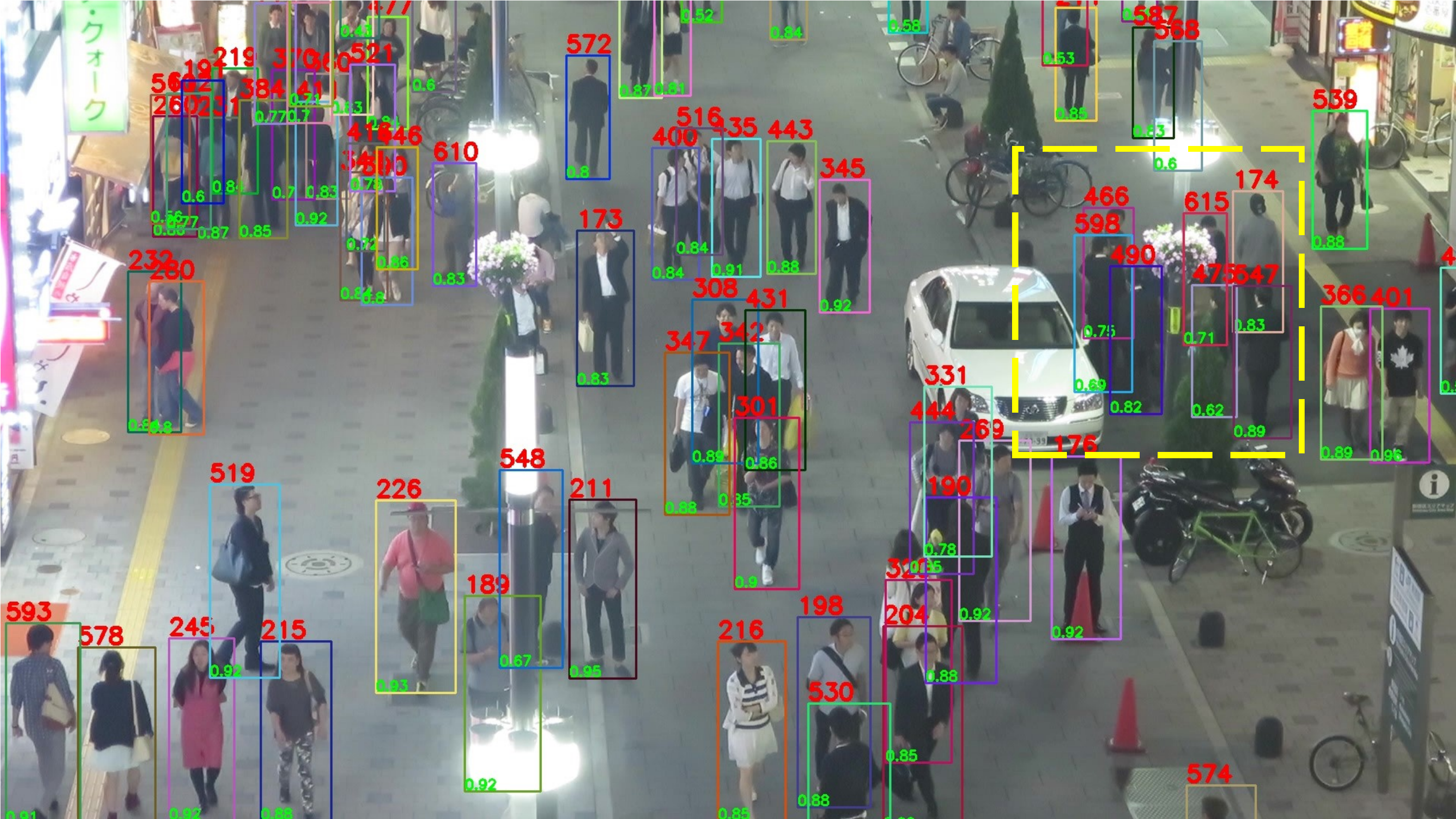}
  \caption{Baseline, MOT17-03-000830}
  \label{fig:Baseline, MOT17-03-000830}
\end{subfigure}
\hfill
\begin{subfigure}{.32\linewidth}
  \centering
  \includegraphics[width=1.0\linewidth]{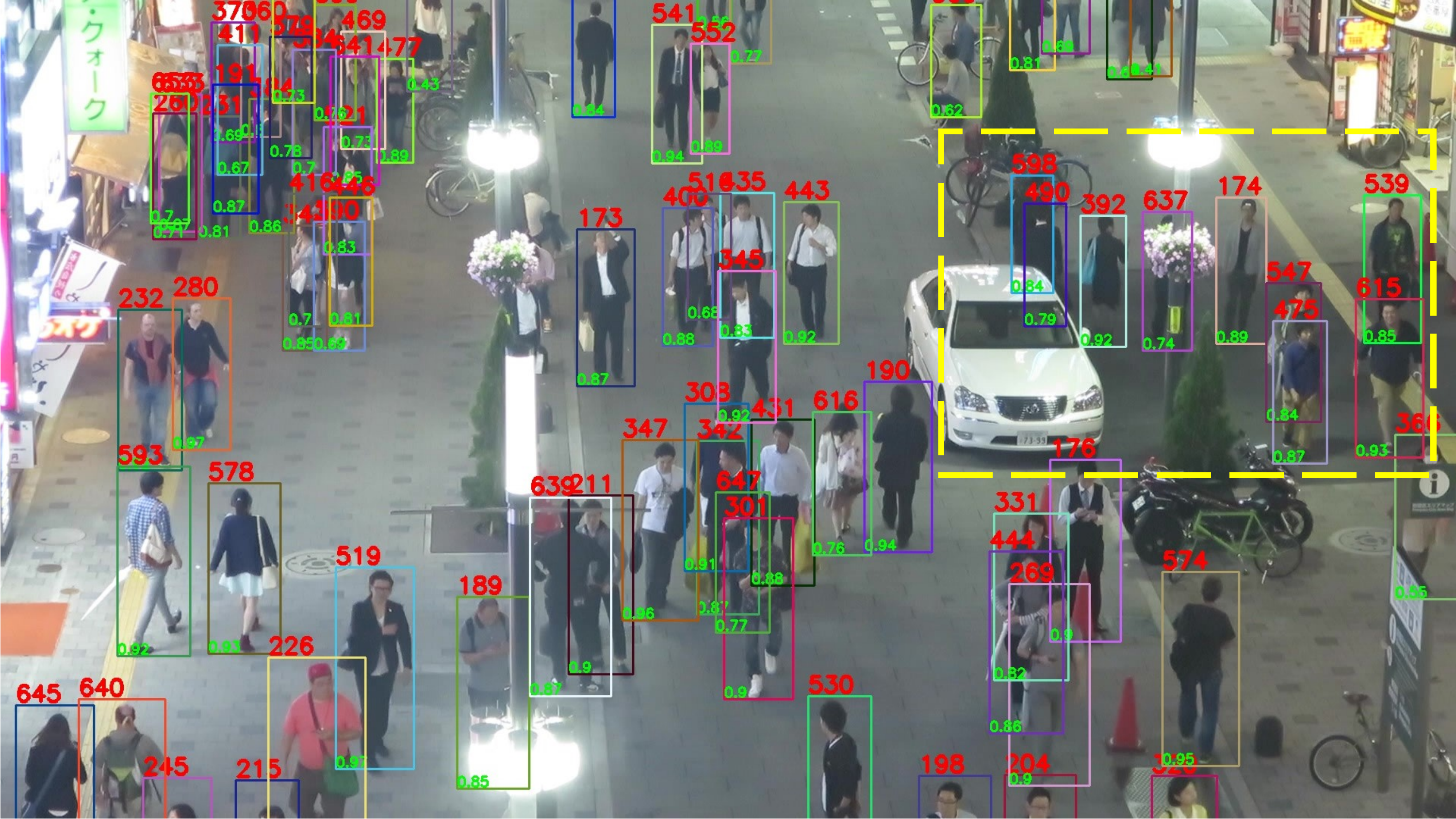}
  \caption{Baseline, MOT17-03-000930}
  \label{fig:Baseline, MOT17-03-000930}
\end{subfigure}
\newline
\begin{subfigure}{.32\linewidth}
  \centering
  \includegraphics[width=1.0\linewidth]{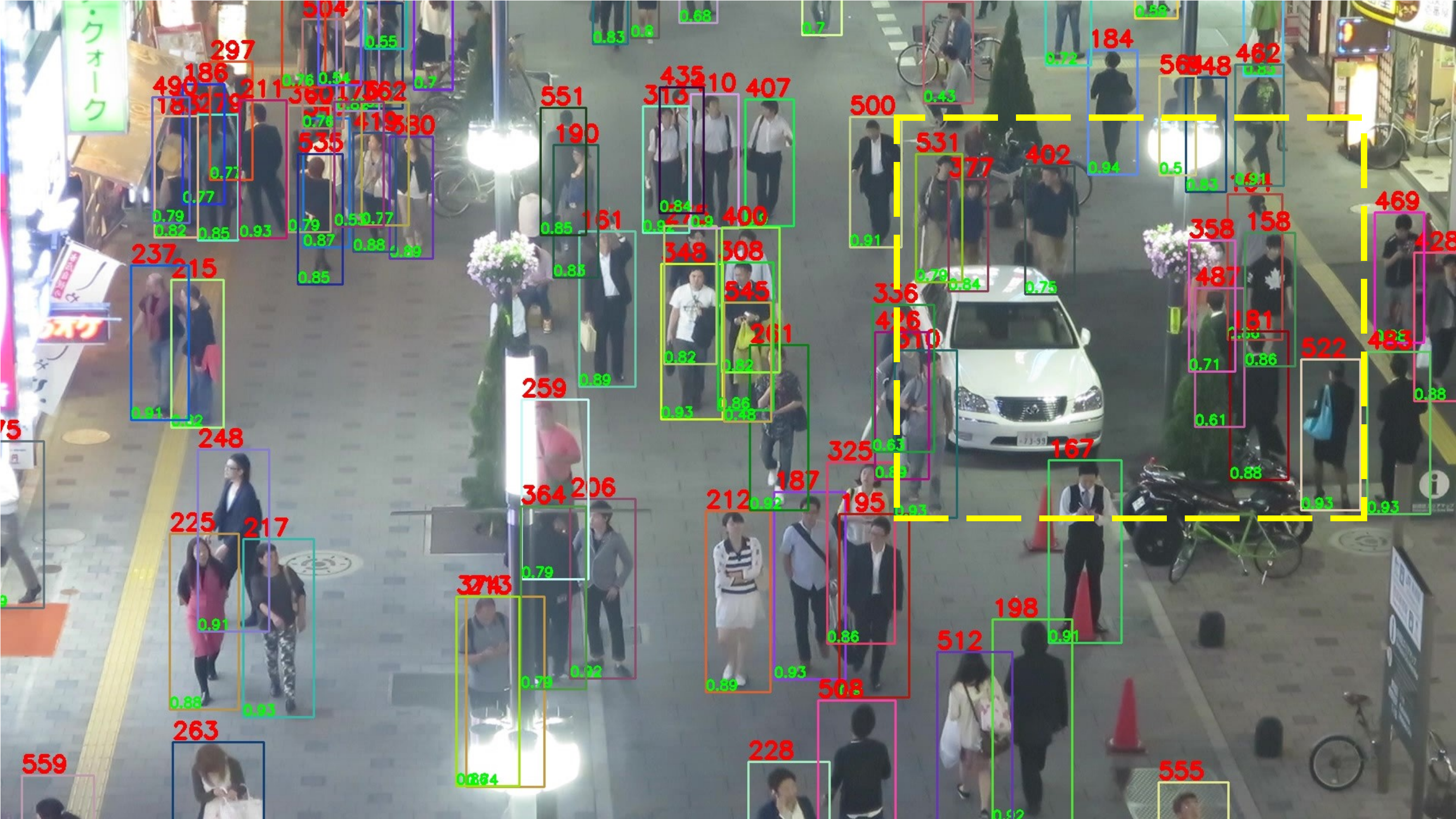}
  \caption{Our method, MOT17-03-000750}
  \label{fig:Our method, MOT17-03-000750}
\end{subfigure}
\hfill
\begin{subfigure}{.32\linewidth}
  \centering
  \includegraphics[width=1.0\linewidth]{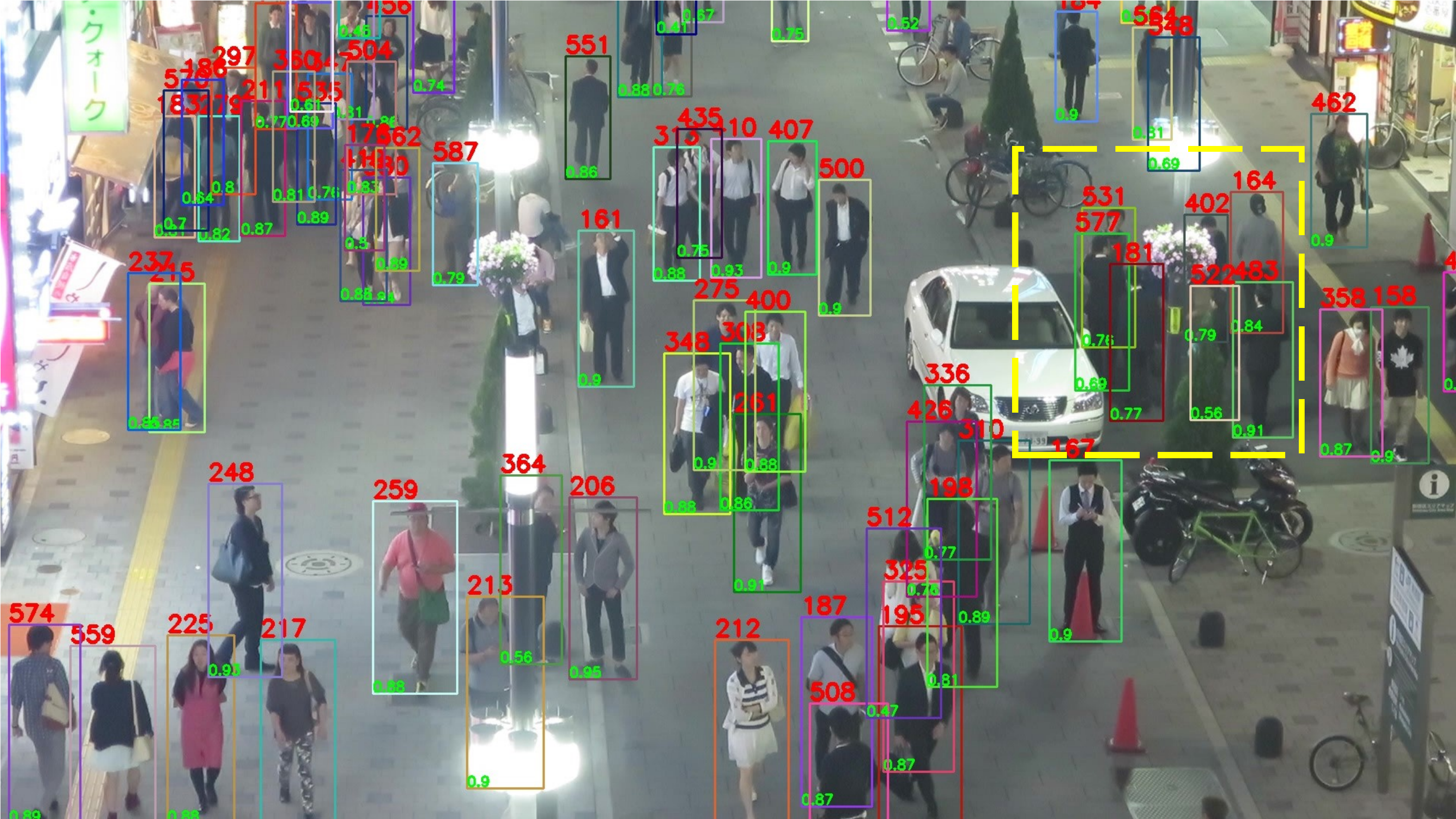}
  \caption{Our method, MOT17-03-000830}
  \label{fig:Our method, MOT17-03-000830}
\end{subfigure}
\hfill
\begin{subfigure}{.32\linewidth}
  \centering
  \includegraphics[width=1.0\linewidth]{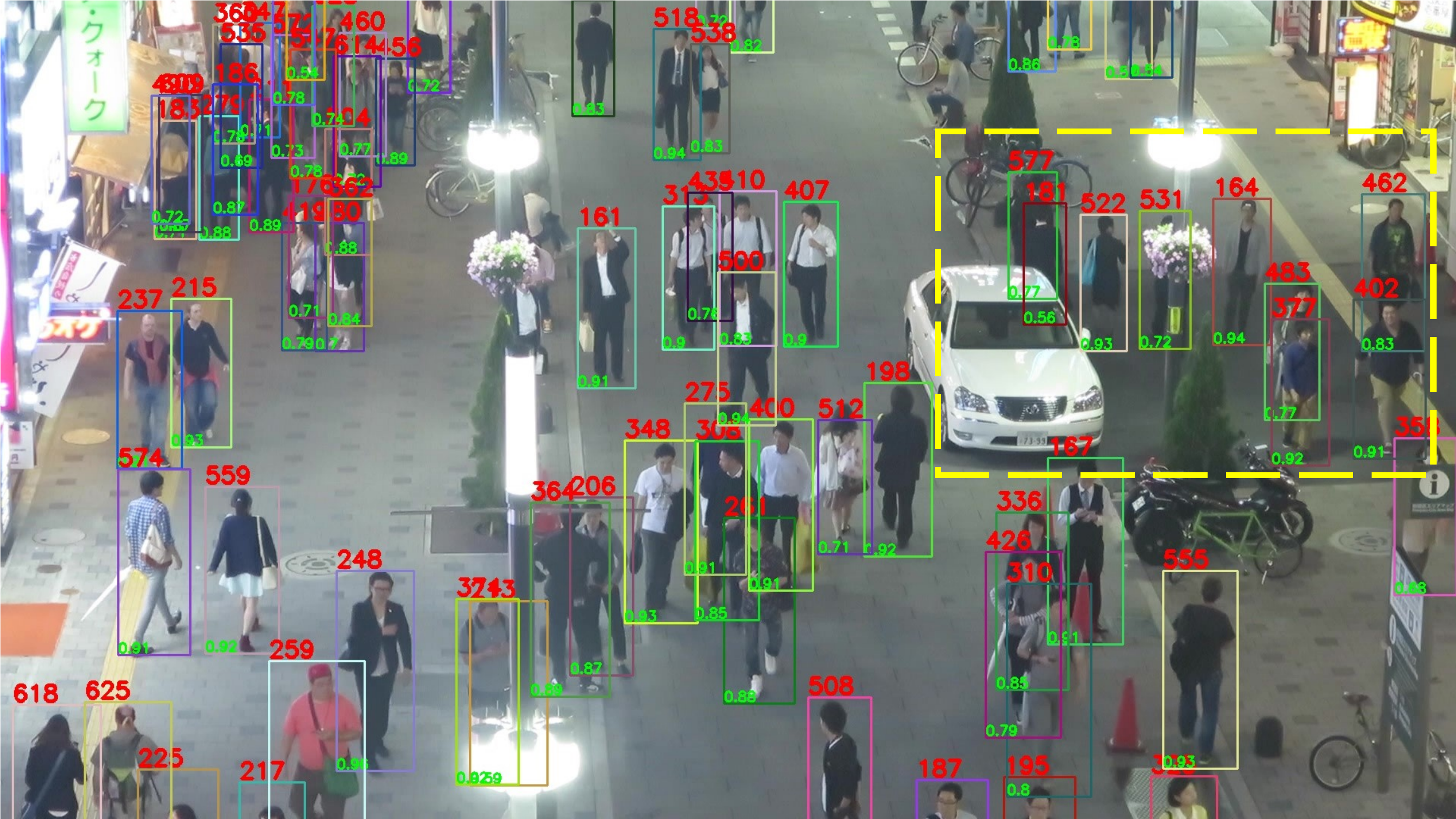}
  \caption{Our method, MOT17-03-000930}
  \label{fig:Our method, MOT17-03-000930}
\end{subfigure}
\caption{Qualitative tracking results of our method. We use the baseline and our method to perform tracking results comparisons on the test set. Bounding boxes and identities are marked in the images. The area of note is marked with a yellow arrow and dashed box.}
\label{fig:Qualitative tracking results of our method.}
\end{figure*}

\quad
In order to demonstrate our method more intuitively, we give examples of SVA and DVA. For SVA, as shown in \cref{fig:Example of backtracking continuation in the Stationary Camera View Data Augmentation (SVA).}, we can observe that the pedestrian in the blue box is backtracked in subsequent frames. As shown in \cref{fig:Example of prediction continuation in the Stationary Camera View Data Augmentation (SVA).}, we can observe that the pedestrian appearing in the blue box of the final frame was predicted in the previous frame. For DVA, as shown in \cref{fig:Comparison of the original image and the image processed by DVA.}, we can observe that the image processed by DVA completely retains the pedestrians, but the background changes unpredictably. We can find that after the image is processed by DVA, the unlabeled negative sample pedestrians are transformed as the background and disappear from the image, that is, the negative sample pedestrians are eliminated. In addition to quantitative tracking results, we also provide qualitative tracking results. We use the FairMOT network as our baseline to compare the qualitative tracking results baseline and our method. As shown in \cref{fig:Qualitative tracking results of our method.}, our method can better reduce pedestrian identity changes caused by occlusions.

\subsection{Questions and Replies}
\label{sec:Questions and Replies}

\noindent \textbf{(1) About the amount of data used in the training phase.}
For data selection, only the SVA-processed image is selected for each selection of stationary camera data, and the DVA-processed image or the original image is selected for each selection of dynamic camera data. This approach ensures that the baseline and our method use the same amount of image data for each epoch of training.

\noindent \textbf{(2) Inference.}
At inference, our method is exactly the same as the baseline. It is worth noting that the fully connected layer of the Re-ID branch is not involved at inference, and the previous layer of the fully connected layer of the Re-ID branch of the network is used as the Re-ID feature.

\noindent \textbf{(3) Motivation of our work.}
MOT is a comprehensive task involving object detection and trajectory association, of which trajectory association is the core issue. For efficient trajectory association, the key is to obtain good pedestrian appearance features. However, the existence of the long-tail distribution of trajectory length makes the network prone to over-focusing on learning pedestrians with longer trajectories and insufficient learning when learning pedestrians with shorter trajectories, which affects the network's acquisition of good pedestrian appearance features.

\noindent \textbf{(4) Why choose the long-tail distribution of trajectory length?} 
Because the long-tail distribution of trajectory length directly affects the network's ability to obtain good pedestrian appearance features, thus affecting trajectory association in the MOT task.

\noindent \textbf{(5) Novelty of SVA and DVA.}
SVA is specially designed for the MOT task, which makes full use of target trajectory characteristics, while the Copy-Paste data augmentation lacks the utilization of target trajectory characteristics and its pasting position is random. DVA aims at generating diverse backgrounds, utilizing SAM to eliminate the foreground and avoid foreground interference with the background, and skillfully combining SAM and Diffusion, as shown in \cref{fig:DVA}.

\noindent \textbf{(6) Benefits of DVA.}
DVA alters the backgrounds of pedestrians so that the network is more focused on learning the appearance characteristics of the pedestrians themselves.

\end{document}